\documentclass[journal]{IEEEtran}

\ifCLASSINFOpdf
  \usepackage[pdftex]{graphicx}
\else
\fi

\usepackage{amsmath}
\usepackage{amssymb}
\usepackage{url}
\usepackage{comment}
\usepackage{subfigure}

\usepackage{amsmath,amsfonts,bm}

\def\eqref#1{equation~\ref{#1}}

\def\1{\bm{1}}

\DeclareMathAlphabet{\mathsfit}{\encodingdefault}{\sfdefault}{m}{sl}
\SetMathAlphabet{\mathsfit}{bold}{\encodingdefault}{\sfdefault}{bx}{n}

\usepackage[breaklinks=true,bookmarks=false]{hyperref}

\begin{document}

\title{Deep Lidar-guided Image Deblurring}

\author{Ziyao~Yi, Diego~Valsesia, \IEEEmembership{Member, IEEE}, Tiziano~Bianchi, \IEEEmembership{Member, IEEE}, and~Enrico~Magli, \IEEEmembership{Fellow, IEEE}
\thanks{The authors are with Politecnico di Torino -- Department of Electronics and Telecommunications, Italy. email: \{name.surname\}@polito.it. This study was carried out within the ``AI-powered LIDAR fusion for next-generation smartphone cameras (LICAM)'' project (CUP: E53D23000790006) – funded by European Union – Next Generation EU  within the PRIN 2022 program (D.D. 104 - 02/02/2022 Ministero dell’Università e della Ricerca). This manuscript reflects only the authors’ views and opinions and the Ministry cannot be considered responsible for them.}
}

\maketitle

\begin{abstract}
The rise of portable Lidar instruments, including their adoption in smartphones, opens the door to novel computational imaging techniques. Being an active sensing instrument, Lidar can provide complementary data to passive optical sensors, particularly in situations like low-light imaging where motion blur can affect photos. In this paper, we study if the depth information provided by mobile Lidar sensors is useful for the task of image deblurring and how to integrate it with a general approach that transforms any state-of-the-art neural deblurring model into a depth-aware one. To achieve this, we developed a universal adapter structure that efficiently preprocesses the depth information to modulate image features with depth features. Additionally, we applied a continual learning strategy to pretrained encoder-decoder models, enabling them to incorporate depth information as an additional input with minimal extra data requirements. We demonstrate that utilizing true depth information can significantly boost the effectiveness of deblurring algorithms, as validated on a dataset with real-world depth data captured by a smartphone Lidar.
\end{abstract}

\section{Introduction}

Image deblurring stands as a crucial tasks in the low-level vision realm, especially in the digital age where cameras have become ubiquitous across a wide array of personal electronic devices such as smartphones and tablets. These devices can easily capture blurred images due to various factors, including hand shaking because of limited space and budget for anti-shake hardware, as well as poor focusing.

Deblurring algorithms have been studied for decades with the aim to recover clear and sharp images from those with indistinct or blurred details. Mathematically, deblurring is an ill-posed inverse problem, which requires strong priors about the nature of the images to be recovered in order to be effectively regularized. Indeed, the recent successes of data-driven methods based on neural networks \cite{Zhang_2018_CVPR,9055201,Kupyn_2018_CVPR,9008540,dong2023multi,Kim2022MSSNet,ren2020neural,chen2021pre,mao2023deep,bai2020single,liu2024motion,min2023out} can be largely attributed to the ability of learning sophisticated image features from training data. However, comparatively fewer works \cite{hyun2013dynamic,yang2023deformable,hyun2014segmentation,10445844} have focused on ways of incorporating side information, mainly in the form of event-cameras, segmentation information or optical flow, as an alternative way to help with the regularization of the deblurring inverse problem. Historically, the guided filter \cite{he2012guided} has been used to modulate a filtering process with a guidance signal for this purpose.

On this note, multimodal imaging platforms, which combine multiple kinds of imaging devices, are currently gaining popularity. In particular, recent mobile devices, such as the Apple iPhone and iPad \cite{applelidar}, are now being equipped with Lidar sensors to provide depth information that can be used for 3D scanning capability. Such Lidars are time-of-flight sensors which send a grid of light pulses and measure the return time to estimate distance at multiple points, thus providing a depth map of the scene. An active sensing instrument is particularly interesting, as it can complement passive optical cameras. In particular, these latter can be prone to generating blurry images in situations requiring, even slightly, longer exposures, integrating received light over a longer periods of time but being sensitive to camera shakes on handheld devices. Alternatively, errors in focusing under challenging conditions may also result in blurred images. This raises the question of whether true depth information from Lidar sensors, particularly smartphone ones, can be effectively used to regularize the deblurring problem and improve image quality.

Several challenges need to be faced in order to answer this question. First, our focus will be on smartphone Lidars and cameras, as this is, possibly, the most widespread multimodal sensing platform at the moment. However, mobile Lidars have significant limitations in spatial resolution due to their size and cost, so it is not obvious if they can provide sufficient information. Moreover, state-of-the-art image restoration models based on neural networks require large datasets to be effectively trained. At the moment, there is no existing dataset of blurry images with associated Lidar depth maps captured by smartphones and assembling one of large size to enable effective training from scratch is indeed a challenging task.

In this work, we positively answer the question of whether smartphone Lidar can boost image deblurring performance. We propose a novel approach that integrates depth maps with blurred RGB images in a way that is able to address the aforementioned challenges. In particular, we design a novel universal adapter neural network to process depth maps and use their features to modulate the features extracted by any state-of-the-art image restoration model. The adapter also deals with the limited resolution of mobile Lidar depth maps by including a super-resolution operation that is capable of preserving their piecewise constant nature when upscaling them to the target resolution. We additionally employ a continual learning strategy to integrate the adapter into existing state-of-the-art deblurring models.

In summary, our main contributions can be regarded as:
\begin{itemize}
    \item we show that true depth information obtained by mobile Lidar sensors can be used to improve image deblurring performance, as experimentally verified with real-world Lidar data;
    \item we propose a novel universal adapter structure that efficiently integrates depth map features with image features for improved deblurring results;
    \item we devise a continual learning approach to pretrained encoder-decoder models to minimize the extra data requirements of exploiting depth information. 
\end{itemize}

\section{Background and Related Work}

\subsection{Image deblurring}

Deblurring is a classic ill-posed inverse problem, that has a long history of methods attempting the reconstruction of a sharp image from the blurred observations. Mathematically, the observed image $\mathbf{y}$ is modeled as the convolution between a blur kernel $\mathbf{k}$ and the original image $\mathbf{x}$:
\begin{align}
    \mathbf{y} = \mathbf{k}  \circledast  \mathbf{x}.
\end{align}
The knowledge of the blur kernel discriminates between non-blind and blind deblurring problems.

Traditional approaches cast reconstruction as the solution of a regularized least squares optimization problem, where the regularizer incorporates prior knowledge about the original images. Extensive literature has focused on devising sophisticated image priors \cite{kundur1996blind,starck2003wavelets,fergus2006removing,krishnan2009fast,danielyan2011bm3d,cai2011framelet}.

Deep learning approaches have enjoyed great success thanks to their ability to learn the prior directly from training data. A common approach is supervised learning, where pairs of blurred and unblurred pictures are available for training \cite{Zhang_2018_CVPR,Kupyn_2018_CVPR,9008540,dong2023multi,Kim2022MSSNet,chen2021pre,wang2022uformer,zamir2022restormer,chen2022simple}. Zhang et al. \cite{Zhang_2018_CVPR} proposed a network composed of three CNNs and a RNN to be used as a deconvolution operator. DeblurGAN \cite{Kupyn_2018_CVPR} and DeblurGAN-v2 \cite{9008540} introduced adversarial learning for deblurring tasks. MRLPFNet \cite{dong2023multi} and MSSNet \cite{Kim2022MSSNet} used multi-scale architectures to jointly explore image details and main structures for deblurring. Given the success of Transformers in both language processing \cite{vaswani2017attention} and vision \cite{dosovitskiy2020image}, they have also gained attention in the deblurring literature. For example, IPT \cite{chen2021pre} first applied standard transformer blocks and trained on the large-scale datasets. More recently. Uformer \cite{wang2022uformer} designed a general U-shaped Transformer-based structure, which proved to be efficient and effective for image restoration. Stripformer\cite{Tsai2022Stripformer} constructed intra- and inter-strip tokens to reweight image features.  Restormer \cite{zamir2022restormer} proposed a Transformer-based architecture that is able to learn long-range dependencies while remaining computationally efficient. On the other hand, NAFNet \cite{chen2022simple} proved that simple neural network designs using gating mechanisms can obtain excellent performance without the need for the expensive self-attention operation of Transformers. 

Some works have studied the use of information from other domains such as segmentation maps \cite{hyun2013dynamic,10445844}, optical flow \cite{hyun2014segmentation}, and event-camera data \cite{yang2023deformable} to better regularize the reconstruction process and improve image quality.
A small number of works have also explored using depth information to improve deblurring performance. \cite{6215220} proposed a hierarchical depth estimation based on region trees to progressively generate credible blur kernel estimates. Li et al. \cite{9043904} first extract the depth map and adopt a depth refinement network to restore the edges and structure in the depth map. Notably, all these works estimate depth information from the blurry image itself. This is in contrast with our work, which uses external information in the form of Lidar data, posing the issue of ensuring its quality and proper fusion, but, theoretically, providing truly independent measurements.

Deep learning approaches to image deblurring require careful dataset selection, in order to ensure the availability of sufficient data for training that are representative of the problem of interest. Some datasets for the deblurring tasks are generated by averaging frames from high-frame-rate videos in order to simulate blurs due to long exposures. For example, the GoPro \cite{nah2017deep} dataset took 240 fps videos with the GoPro cameras and then averaged a variable number (7 - 13) of successive frames to produce blurs of different strengths. Similarly, the REDS \cite{9025509} and DVDs \cite{su2017deep} datasets have been constructed for video deblurring. Besides averaging video frames, Levin \cite{5206815} collected real blurred images by capturing images on the wall while shaking the cameras and providing the blur kernels. The works in \cite{ren2020neural,zhong2020efficient}, and \cite{rim2022realistic} also provide images affected by real blur.

In this work, we are interested in studying the performance of deblurring when in presence of depth data acquired by mobile Lidars. No such dataset is currently readily available for this task. However, the ArkitScenes \cite{dehghan2021arkitscenes} dataset presents a large number of indoor scenes acquired with Apple mobile devices, iPhones and iPads, together with registered Lidar scans. In particular, the depth information is provided both by Lidar scans from the mobile sensors and scans from a high-end professional Lidar (Faro Focus S70). While the RGB images are not affected by blur, it is possible to simulate it using standardized blur kernels \cite{5206815} used for benchmarking belurring methods. This allows to study the effectiveness of the depth information in regularizing the deblurring procedure. The availability Lidar data with both low and high spatial resolution also allows to study its impact on the restoration process.

\begin{figure*}[ht]
  \centering
    \includegraphics[width=0.8\linewidth]{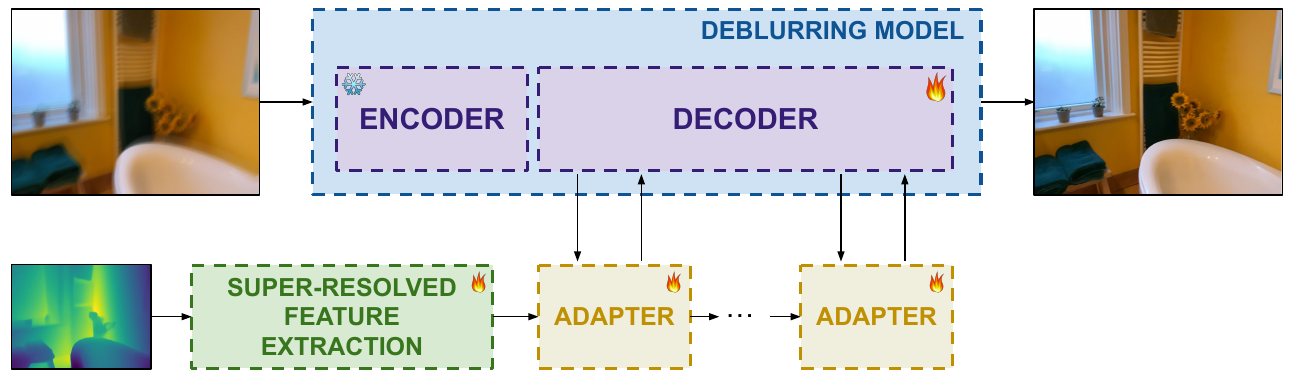}
    \caption{Pipeline of a generic depth-guided image deblurring model. The encoder part of the deblurring pretrained model is frozen, while the decoder part is trainable. Depth feature extraction is combined with super-resolution, if needed. Adapters modulate image features in the decoder model according to depth features.}
    \label{fig:pipeline}
\end{figure*}

\subsection{Continual Learning}
The effectiveness of deep learning models increases significantly, as they are scaled in size and amount of training data. Many recent models can be very large and expensive to train, thus when one wants to leverage them for a task that is not exactly the training one, it is desirable to incorporate the new requirements in the least expensive way possible. This has brought a recent interest in novel ways to continue training such as low-rank adapters (LoRA) \cite{hu2021lora} and HyperNetworks \cite{ha2016hypernetworks}. This field is generally known as ``continual learning'', or incremental learning, or lifelong learning, and has developed multiple strategies to accomplish its goals, e.g., regularization, model expansion, or rehearsal \cite{10444954}. In the case of this paper, which involves extending existing models to support a new modality, the scheme we develop falls in the general class of model expansion, which adds parameters to learn the new tasks, classes, or incorporate new modalities. 

An example of continual learning methods in computer vision is represented by VPT \cite{jia2022visual}, which introduces less than $1\%$ of new trainable parameters in the input space while keeping the model backbone frozen to solve various downstream image recognition tasks. Gao et al. propose CPrompt \cite{gao2024consistent}, for more aligned training and testing, which surpasses its prompt-based counterparts and achieves state-of-the-art performance on multiple continual learning benchmarks. Besides pure vision prompt learning, multi-modal prompt learning has also been proposed. MaPLe \cite{Khattak_2023_CVPR} improves alignment between the vision and language representations. Zhu et al. \cite{zhu2023visual} developed Visual Prompt multi-modal Tracking (ViPT) which outperforms the full fine-tuning paradigm on multiple downstream tracking tasks including RGB+Depth, RGB+Thermal, and RGB+Event tracking. 

In the image restoration literature, Potlapall et al. \cite{potlapalli2023promptir} proposed PromptIR, which uses prompts to encode degradation-specific information and then guides the restoration network dynamically. ProRes \cite{ma2023prores} and PromptGIP \cite{liu2023unifying} add additional images as its prompts. MiOIR \cite{kong2024towards} adopts sequential prompt learning strategies to adapt to multiple restoration tasks. However, these methods are focused on the same image domain, or focus on the solution of new restoration tasks, neglecting to explore the possibility of using the information from other domains. In our work, we need to carefully devise continual learning strategies to modulate existing image features at a pixel level in order to incorporate information from a different modality with different properties such as radiometry and spatial resolution.

\section{Proposed Method}

This section presents the proposed approach towards studying the effectiveness of mobile Lidar depth maps, as well as the novel solutions to effectively incorporate this information into state-of-the-art deblurring models.

\subsection{Intuition}
Image blur in photos is usually the result of improper focusing or motion due to long exposure times, e.g. because of hand shaking, and results in object boundaries or edges that appear indistinct and smeared. Lidar is an active instrument that is capable of measuring the distance of surfaces in the scene from the camera. Being active, it does not suffer from the same limitations as RGB cameras. For instance, in a low-light setting, the need to integrate light for a sufficiently long time results in blur due to even modest shaking. However, the Lidar measurement is unaffected by this and can discern sharp object boundaries, provided a sufficiently high-resolution capture, in the form of a depth map. Since results in literature \cite{9043904} observed that even estimating depth from blurry images alone can be useful, the conjecture we seek to verify is whether a real depth map, even of modest resolution, provided by an independent instrument can effectively guide the restoration process towards sharper results, particularly around objects, leading to a more accurate and visually pleasing deblurred image.

Formally, let us call $\mathbf{y}$ the blurry observed RGB image, and $\mathbf{d}$ the depth map acquired by the Lidar instrument, possibly at a lower spatial resolution. We seek to develop a joint multimodal model $f_\theta(\mathbf{y},\mathbf{d})$ that can estimate the clean image $\mathbf{x}$. In developing this model, we need to consider two main factors: i) joint training data with blurry images and associated Lidar depth maps are scarce and will be, in general, difficult to acquire on large scales; ii) future advancements in the state of the art will generally focus on the unimodal image deblurring.
This leads us to formulate the joint model as a factorized one, i.e.,
\begin{align}
    f_\theta(\mathbf{y},\mathbf{d}) = g_{\theta_g}(\mathbf{y}) \circ h_{\theta_h}(\mathbf{d}),
\end{align}
where a unimodal deblurring model from the state of the art $g_{\theta_g}(\mathbf{y}) $ can be combined with a depth processor $h_{\theta_h}(\mathbf{d})$ which takes care of extracting suitable features from the depth data and combining them with those from the deblurring model, possibly in a manner that is as universal as possible. 
A continual learning strategy is desirable to train such joint model and exploit the rich datasets used to train unimodal deblurring models. With this scheme, the limited multimodal training data can be used to train only the $\theta_h$ parameters, and apply a small update to the pretrained $\theta_g$ parameters. 

With this in mind, the following section will focus on three main aspects that are critical for the overall performance of the multimodal model, and its wide applicability, namely: i) extracting high-quality depth features, also addressing the limitations in spatial resolution of the Lidar capture; ii) ensuring the interaction operator $\circ$ is well-constructed; and iii) constructing a suitable continual learning strategy.

\subsection{Lidar-guided Deblurring}

\begin{figure*}[t]
  \centering
    \includegraphics[width=0.85\linewidth]{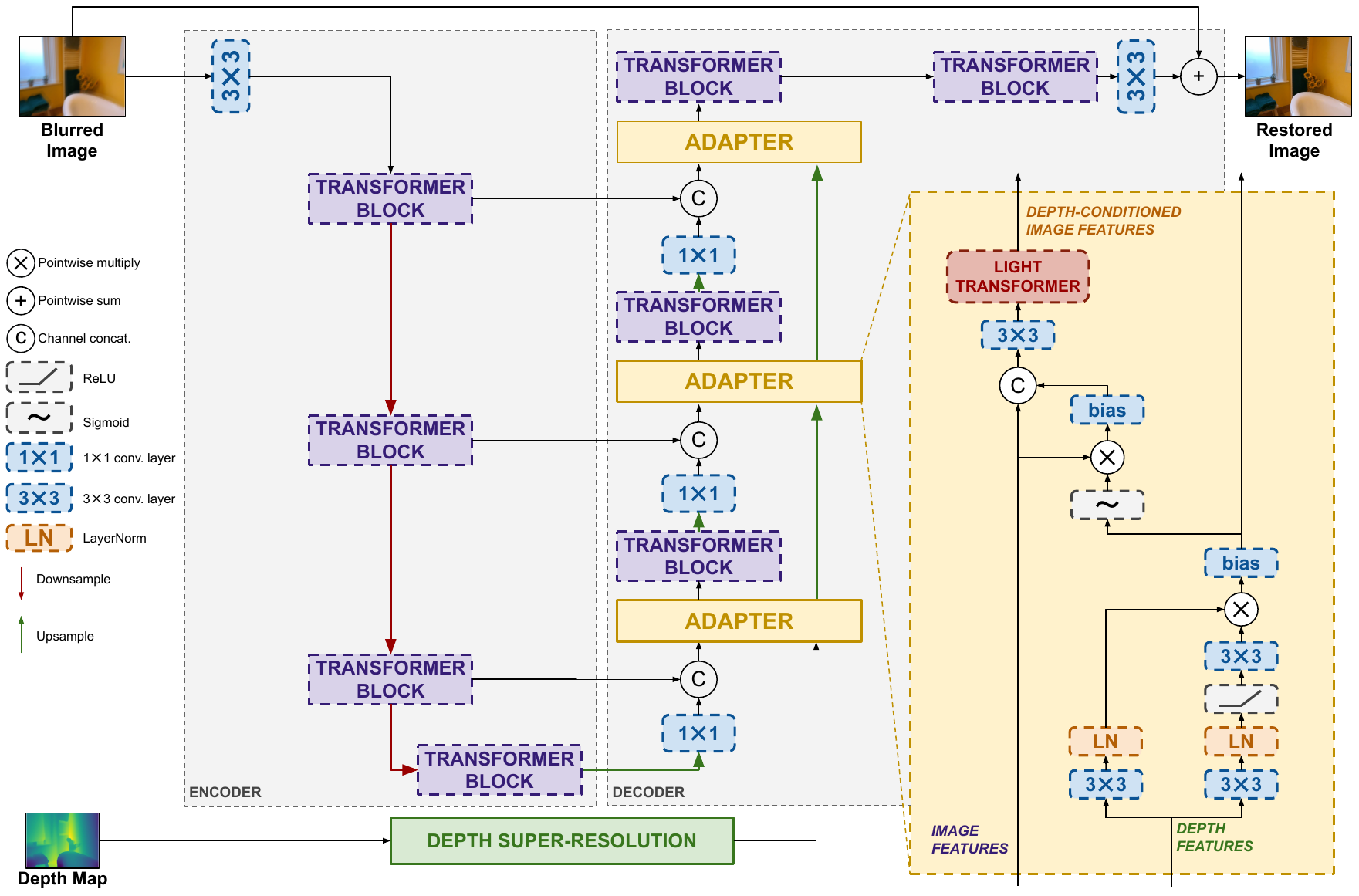}
    \caption{The depth-restormer architecture example. The adapter is added on each level of the decoder part, which is in front of the transformer block. There are also 3 depth blocks corresponding to each adapter. The architecture of the depth block and the adapter is also shown in the figure.}
    \label{fig:depth-restormer}
\end{figure*}

\begin{figure}[t]
  \centering
    \includegraphics[width=0.99\linewidth]{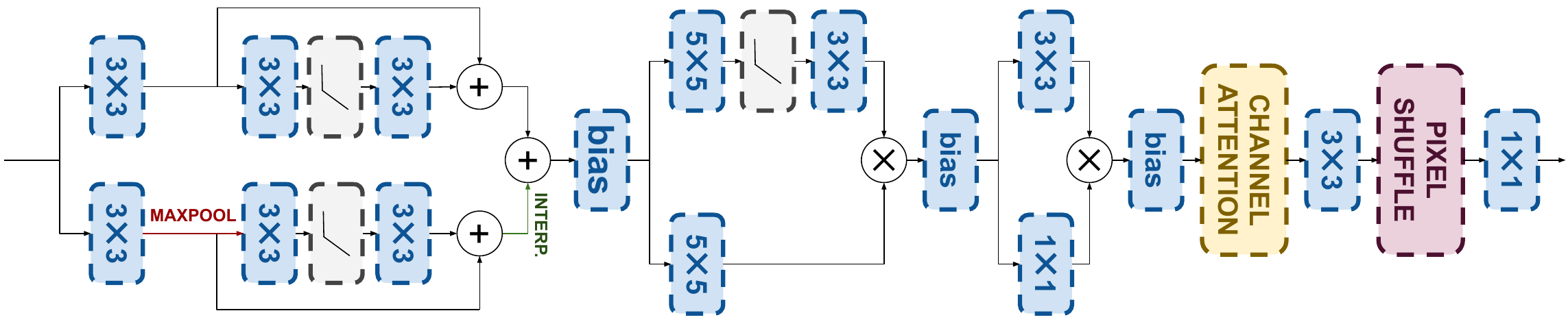}
    \caption{Depth super-resolution architecture.}
    \label{fig:superres-net}
\end{figure}

This section presents the previously mentioned key ingredients towards designing general Lidar-guided image deblurring models. A high-level overview of them is presented in Fig. \ref{fig:pipeline} and a specific adaptation to the Restormer architecture \cite{zamir2022restormer} is shown in Fig. \ref{fig:depth-restormer}.

\subsubsection{Quality of depth features} \label{sec:quality_depth}
Depth maps are approximately piecewise smooth images, which, ideally, should show sharp transitions between objects as they appear at different relative distances or are separated by background. However, the quality of real depth maps, as captured by Lidar instruments, can be variable depending on a number of factors. In particular, they might present missing values due to lost light return when a surface scatters the transmitted pulse away from the camera. This can be coarsely fixed by inpaiting based on neighboring measurements or with some RGB guidance. Additionally, the spatial resolution of the depth map can be limited due to manufacturing size and cost constraints, particularly on mobile devices. For example, the Apple iPad Pro used in the ARKitScenes dataset \cite{dehghan2021arkitscenes} captures depth at a resolution of $256 \times 192$ pixels, which is substantially smaller than the resolution of the RGB camera. 

It is thus clear that feature extraction must be combined with a super-resolution model to extract depth features that are as useful as possible to guide the deblurring model. Indeed, Sec. \ref{sec:exp_superres} will show that naive upsampling of the depth map provides significantly diminished guidance, leading to more modest deblurring improvements. Depth-map super-resolution is known to require ad-hoc models \cite{zhong2023guided,wang2022depth} to preserve the piecewise smooth nature of depth images and their sharp edges. In this work, we propose a lightweight design of a depth super-resolution network whose features serve as the starting point of the depth processing model $h_{\theta_h}$. The architecture we use is detailed in Fig. \ref{fig:superres-net}. A supervised pretraining process is used by exploiting the ARKitScenes data which provide paired captures from the low-resolution mobile Lidar of the iPad Pro and the high-resolution Faro Focus S70 Lidar. Notice that the input for the super-resolution process is just the depth map and the RGB image is not used to prevent contamination with blurry data.
After this pretraining, the last layer which projects features back to image space is removed to directly provide deep features to the adapter modules described in the next section. Indeed, this design choice is motivated by recent results showing that the conventional approach towards integrating depth maps \cite{Rombach_2022_CVPR} in other models via concatenation or cross-attention with the depth map itself can be suboptimal \cite{ariav2023fully}. This is because the depth map can be regarded as a shallow feature and its combination with deeper features may result in information misalignment. Since we seek to integrate depth features with deep features at the decoder stage of encoder-decoder restoration models, it is thus important to leverage deep depth features.

\subsubsection{Depth Adapters}
Fusion of multimodal information in deep neural networks can be performed in several different ways, depending on design constraints and the type of information. For example, popular basic methods use early fusion by concatenating the raw inputs or shallow features of the different modalities. When a pretrained unimodal model is available, this is a suboptimal solution, as the new joint model will not fully exploit the new modality.

In our problem, the side information provided by the depth map can be regarded as a guidance about the edges to be used to regularize the main features extracted from the image. The image filtering literature before deep learning has shown that the guided filter \cite{he2012guided} is a compelling solution to this problem. The guided filter can be regarded as a spatially-variant filter whose coefficients are derived from second order statistics of the guidance signal (see Eq. (11) in \cite{he2012guided}).

We propose a generalization of the guided filtering scheme in the form of lightweight neural network adapters. These adapters process a multiresolution sequence of depth and image features, and, similarly to the guided filter, modulate the image features using the depth ones. More specifically, the proposed adapter architecture is shown in Fig. \ref{fig:depth-restormer} for an encoder-decoder restoration model such as Restormer. The input depth features are first processed by a convolutional attention operation, which mimics the second order statistics of the original guided filter, then passed through a sigmoid to stabilize the operations and then multiplied by the image features. The resulting depth-conditioned image features are concatenated to original image features for further processing by a convolutional layer and a lightweight transformer before being reintroduced to the deblurring model. The lightweight transformer is composed by a sequence of a multi-Dconv head transposed attention (MDTA) and a Gated-Dconv feed-forward network (GDFN), as proposed in \cite{zamir2022restormer}, with a single attention head to constrain complexity. Depth features processed by the attention operation are also propagated forward to the next stage, and possibly upsampled.

\subsubsection{Continual Learning}
Most existing continual learning approaches for multimodal problems focus on high-level vision tasks, such as image segmentation \cite{sariyildiz2020learning}, visual question answering \cite{cossu2024continual}, and tracking \cite{zhu2023visual}, all of which emphasize the semantic information of images. Most of these methods freeze the entire pretrained model and modify the training strategy, leaving only the adapter or a simple decoder (generally, a fully-connected layer) trainable. This approach typically leads to unsatisfactory results in low-level vision tasks \cite{potlapalli2023promptir} such as deblurring, unless generative models are used.  Instead of freezing all the pre-trained model, we only freeze the encoder part of the deblurring model, while keeping the entire decoder part trainable. Additionally, in order to fully utilize the depth map, and match its features with the deep image features of the model decoder, the depth adapter blocks are sequentially connected at multiple points of the decoder, typically at different spatial resolutions. Adapters are fully trained on the paired depth-image data, and the pretrained depth super-resolution network is also finetuned.

\subsubsection{Example: Depth-Restormer}
The proposed approach based on depth super-resolution, adapters and continual learning is well-suited to introduce the Lidar depth information into any state-of-the-art image deblurring model. As an example, we report the full architecture of the well-known Restormer model \cite{zamir2022restormer} with the proposed additions to deal with Lidar depth, resulting in the Depth-Restormer model shown in Fig. \ref{fig:depth-restormer}. In the Restormer architecture, the encoder progressively reduces the image resolution after every transformer block with skip connections to the corresponding layers in the decoder half. Notice how we introduce the adapters in the decoder stage, after every upsampling operation. Similar considerations can be made for any state-of-the-art deblurring method with an encoder-decoder structure or with an inherent symmetric design between first and second half of the model, which is most of the existing approaches. Indeed, Sec. \ref{sec:experiments} reports experiments on depth-enhanced versions of multiple state-of-the-art models, including Restormer \cite{zamir2022restormer}, NAFNet \cite{chen2022simple}, Stripformer \cite{Tsai2022Stripformer}, DeblurDiNATL \cite{liu2024deblurdinat}.

\section{Experimental Results} \label{sec:experiments}
This section reports experimental results to validate several points of interest. First and foremost, we seek to answer the question whether image quality is improved by providing mobile Lidar depth maps. This is done by presenting results on several state-of-the-art deblurring architectures adapted following the proposed approach. Next, we validate the design of the proposed approach, particularly regarding the need for depth super-resolution, the adapter design and the continual learning strategy. 

\subsection{Experimental Setting}

\begin{table*}[t]
  \caption{Effect of Lidar depth maps on state-of-the-art deblurring methods}
\centering
    \begin{tabular}{l ccccccc}
                \textbf{Model}                &\textbf{PSNR (dB) $\uparrow$}     &\textbf{ $\Delta$PSNR}   &\textbf{SSIM$\uparrow$} &\textbf{$\Delta$SSIM} & \textbf{LPIPS$\downarrow$} & \textbf{$\Delta$LPIPS}  & \textbf{Parameters (M)}\\
    \hline\hline
                Restormer \cite{zhong2020efficient}          & 34.52             &       -         &0.9318         &-            &0.1369  &  - &26.1   \\
                \textbf{Depth-Restormer}    & \textbf{36.62}             &  2.10        & \textbf{0.9446}       & 0.0128    & \textbf{0.1093} &  0.0276 &30.0\\
    \hline            
                Stripformer \cite{Tsai2022Stripformer}  & 35.17  & - &0.9337 & - &0.1171 & - &19.7 \\
                \textbf{Depth-Stripformer}          & \textbf{36.34}  & 1.17  &\textbf{0.9412} & 0.0075 &\textbf{0.1118} & 0.0053 &22.2\\
    \hline
                DeblurDiNATL\cite{liu2024deblurdinat}       &36.53              & -              &0.9436                    &- &\textbf{0.1098} & 0.0012 &10.6 \\
                \textbf{Depth-DeblurDiNATL} & \textbf{36.72}              &0.19               &\textbf{0.9448}                     & 0.0012 & 0.1110 & - &12.1 \\
    \hline
                    NAFNet \cite{chen2022simple}             & 37.24             &     -           &0.9430         &  -           &0.1160   & - & 17.1
\\
                \textbf{Depth-NAFNet}       & \textbf{37.28}             & 0.04         & \textbf{0.9434}         & 0.0004            & \textbf{0.1147} & 0.0013  &23.7
\\
    \hline
    \end{tabular}
\label{table:main}
\end{table*}

\begin{figure*}
  \centering
    \subfigure[Depth Map]{\includegraphics[width=0.24\linewidth]{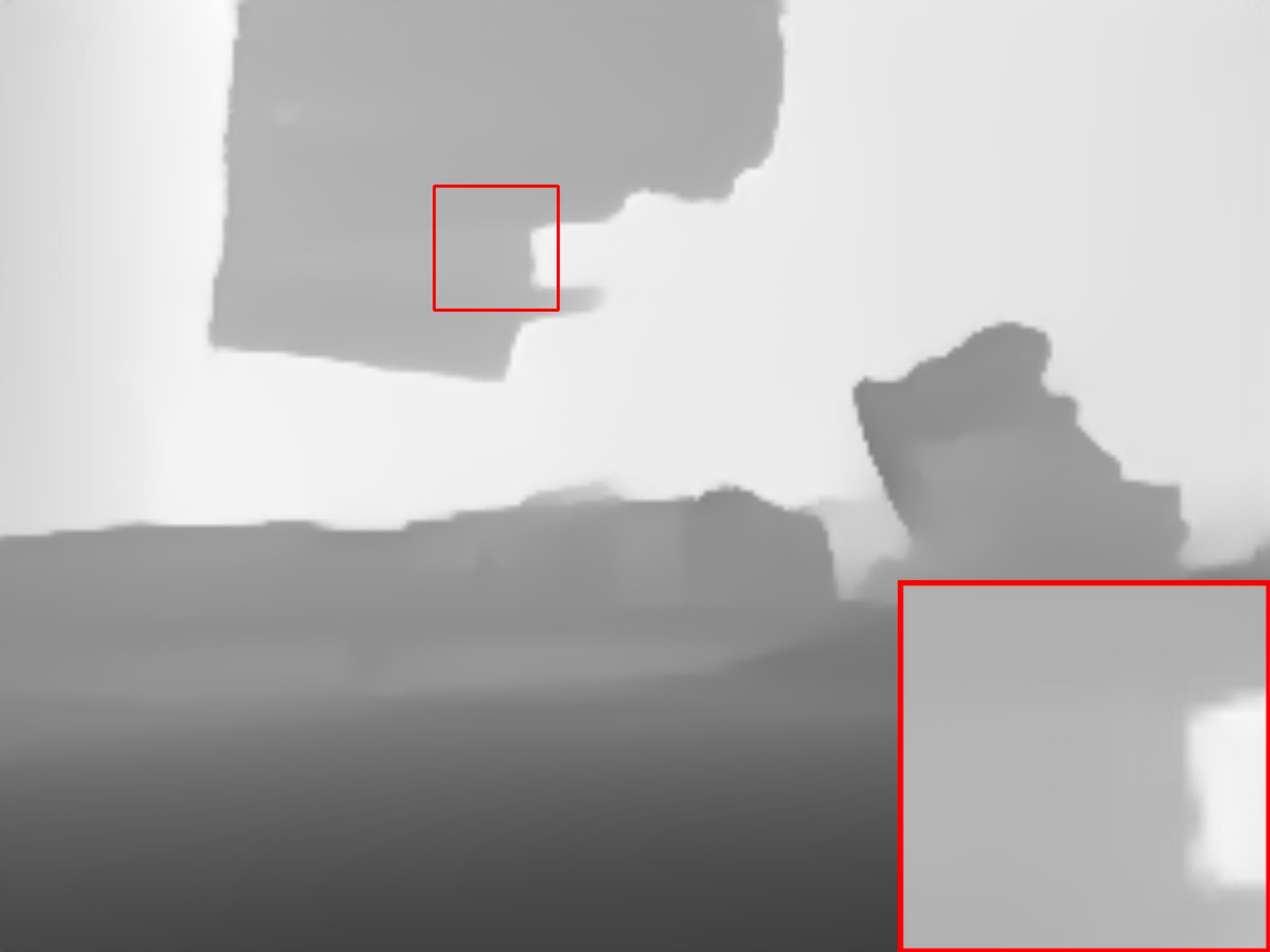}}
    \subfigure[Restormer]{\includegraphics[width=0.24\linewidth]{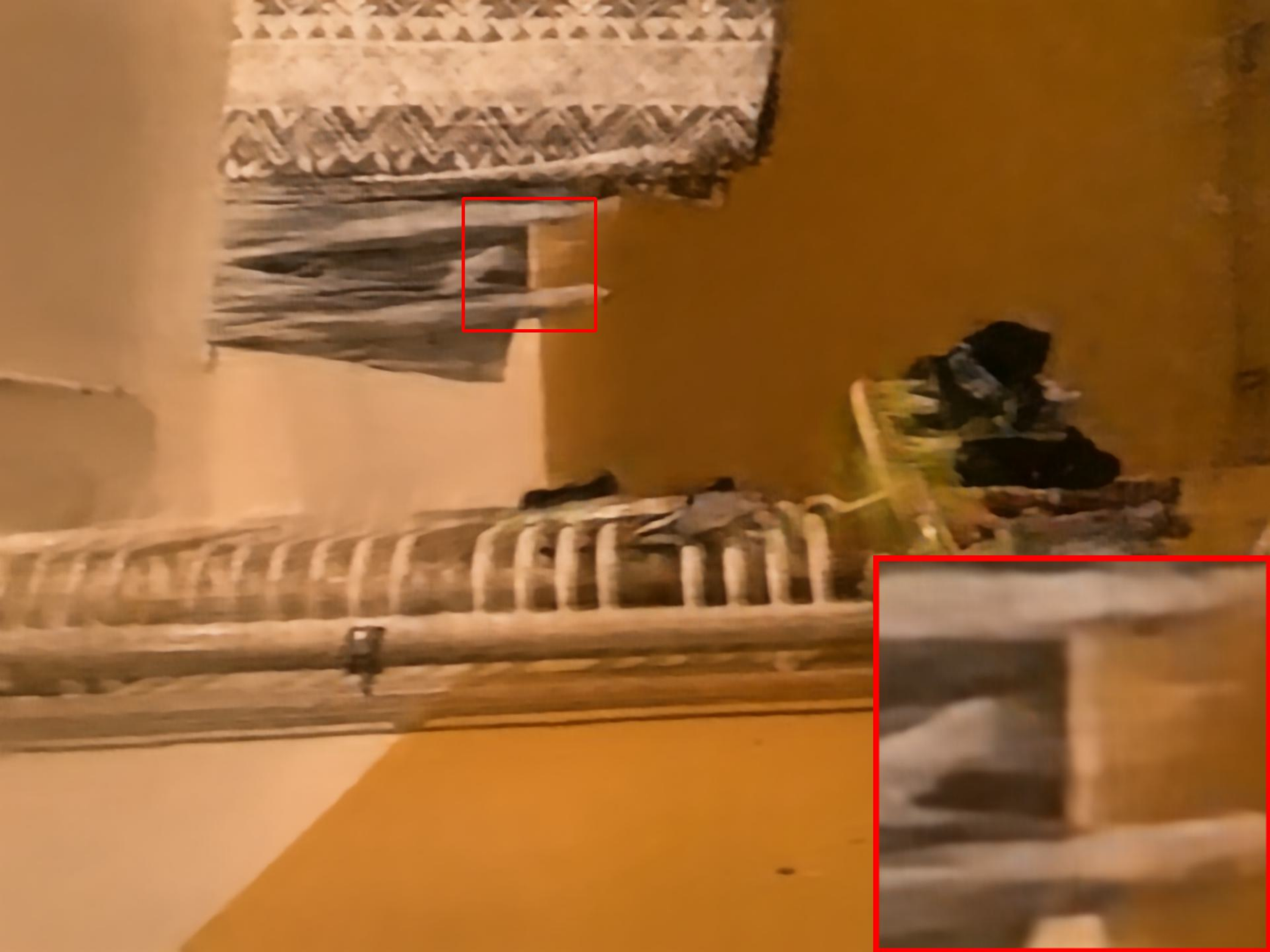}}
    \subfigure[Depth-Restormer]{\includegraphics[width=0.24\linewidth]{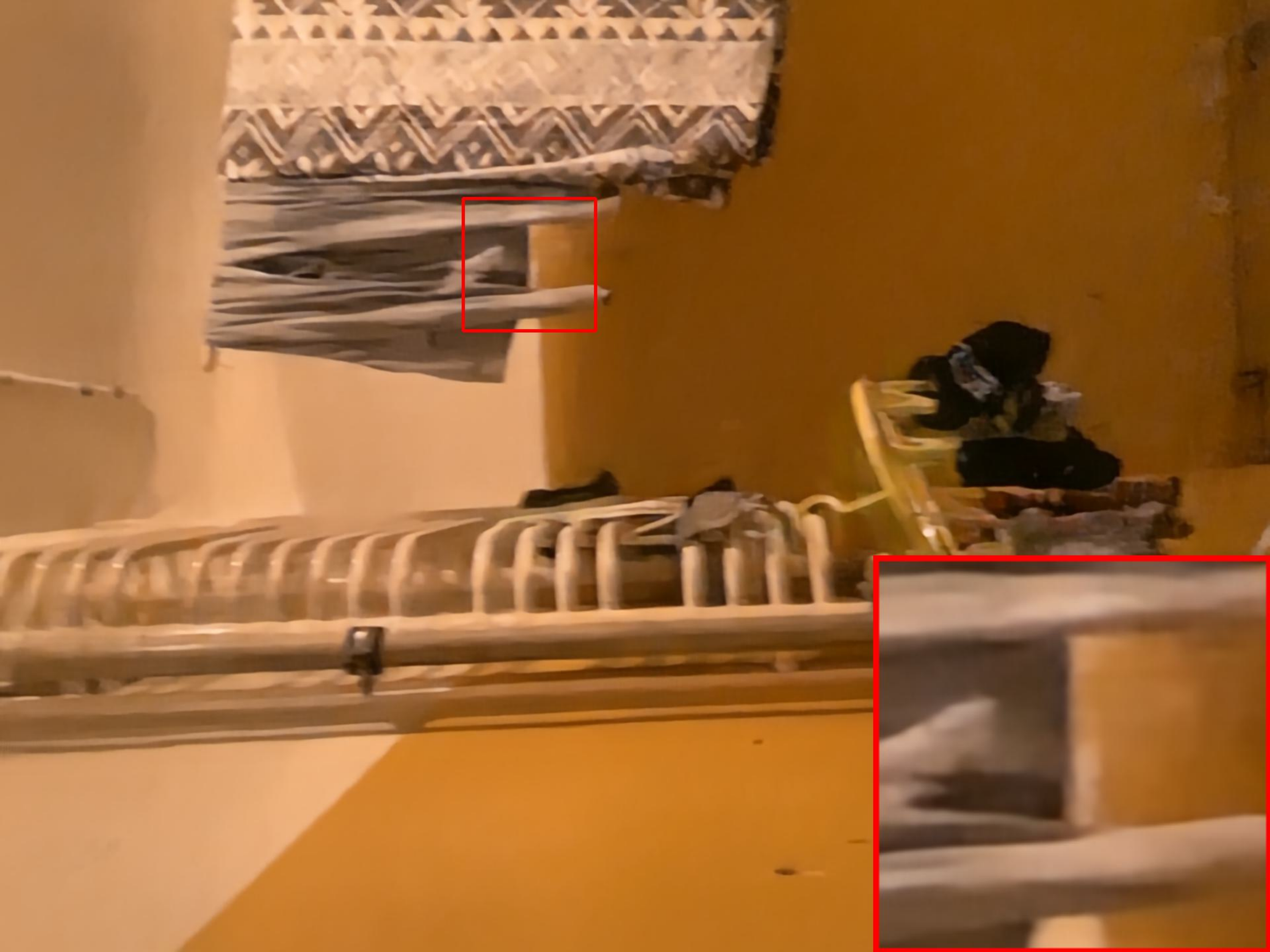}}
    \subfigure[GT]{\includegraphics[width=0.24\linewidth]{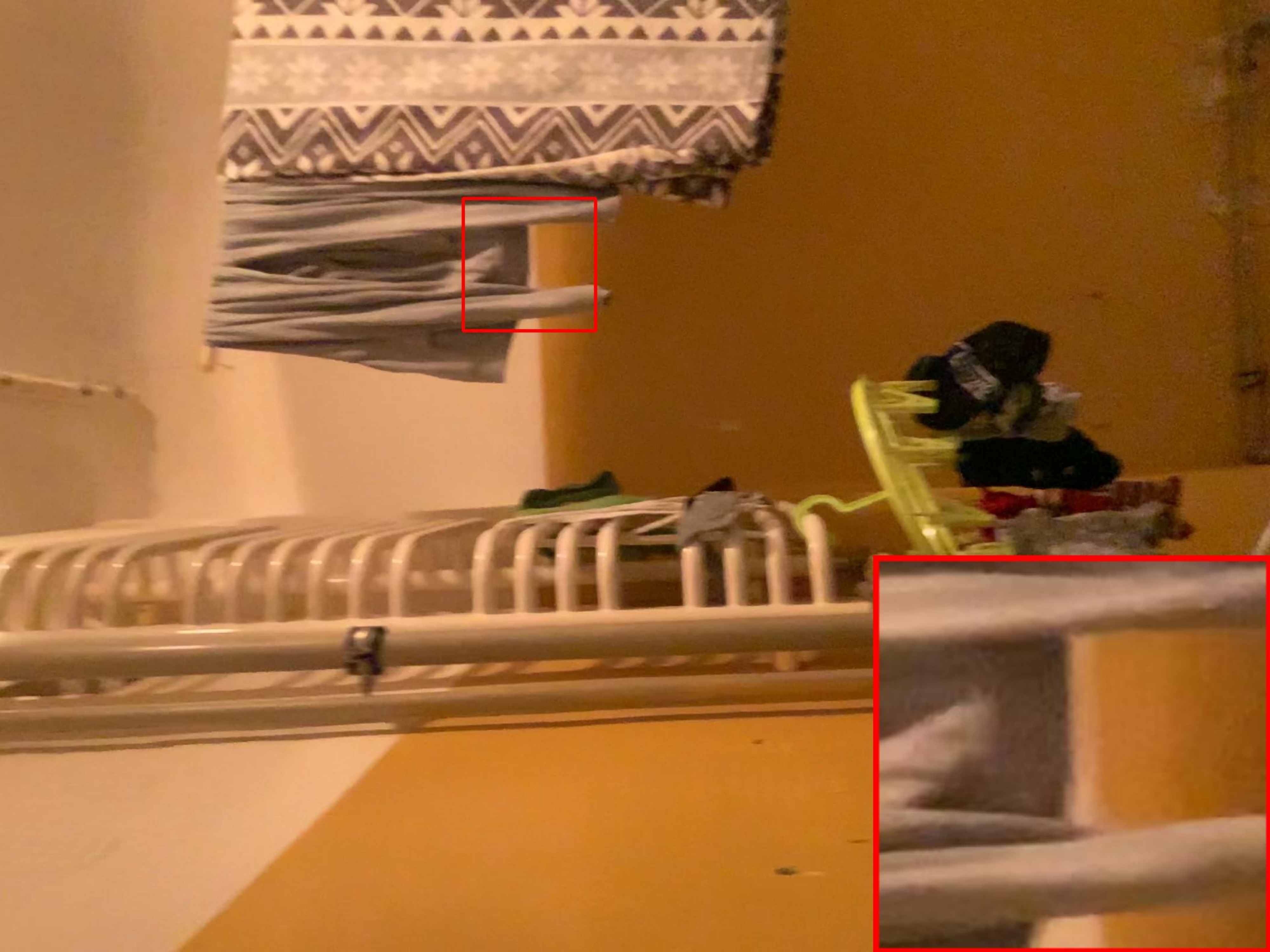}}
  \caption{Left to right: Mobile Lidar depth map, Restormer deblurred image, Depth-Restormer, ground truth. As shown in the detail, Depth-Restormer has sharper object edges.}
\label{fig:result_with_depth}
\end{figure*}

In our experiments, we use a subset of the ArkitScenes dataset \cite{dehghan2021arkitscenes}, specifically the portion used for RGB-D guided upsampling, which contains 29,264 image-depth pairs in the training set. For validation, we randomly sample 500 pairs from the original validation set. Image blur is simulated by randomly choosing a blur kernel from a set of standard benchmark kernels, following the approach from \cite{5206815}. 
The depth super-resolution network is pretrained by using the low-resolution depth maps from the iPad Lidar as input and the high-resolution depth maps from the Faro Focus S70 Lidar as ground truth, discarding any pixels with invalid measurements. The L1 loss is used for this training, which is run for about 100 epochs.

We selected four main state-of-the-art deblurring models to be tested with and without the Lidar augmentation: Restormer \cite{zhong2020efficient}, NAFNet \cite{chen2022simple}, Stripformer \cite{Tsai2022Stripformer}, and DeblurDiNATL \cite{liu2024deblurdinat}. The versions with the proposed Lidar depth improvements is denoted with the prefix ``Depth-*''. Training generally follows the protocols outlined in the original papers, with $128\times128$ patches for the Restormer and Stripformer and $256\times 256$ for NAFNet and DeblurDiNATL. The initial learning rate is $5\times10^{-5}$, and the Adam optimizer with $\beta_1=0.9$ and $\beta_2=0.999$ and cosine annealing decay policy are utilized. The published version of the method and the depth-enhanced one are trained on the same data, and with the same protocol to ensure a fair comparison.

Deblurring results are evaluated in terms of Peak Signal-to-Noise Ratio (PSNR), Structural Similarity Index Measure (SSIM) on RGB images. Also, the LPIPS metric \cite{zhang2018unreasonable} is evaluated for a more perception-oriented metric.

\subsection{Main Results}
\begin{figure*}
  \centering
   \includegraphics[width=0.18\linewidth]{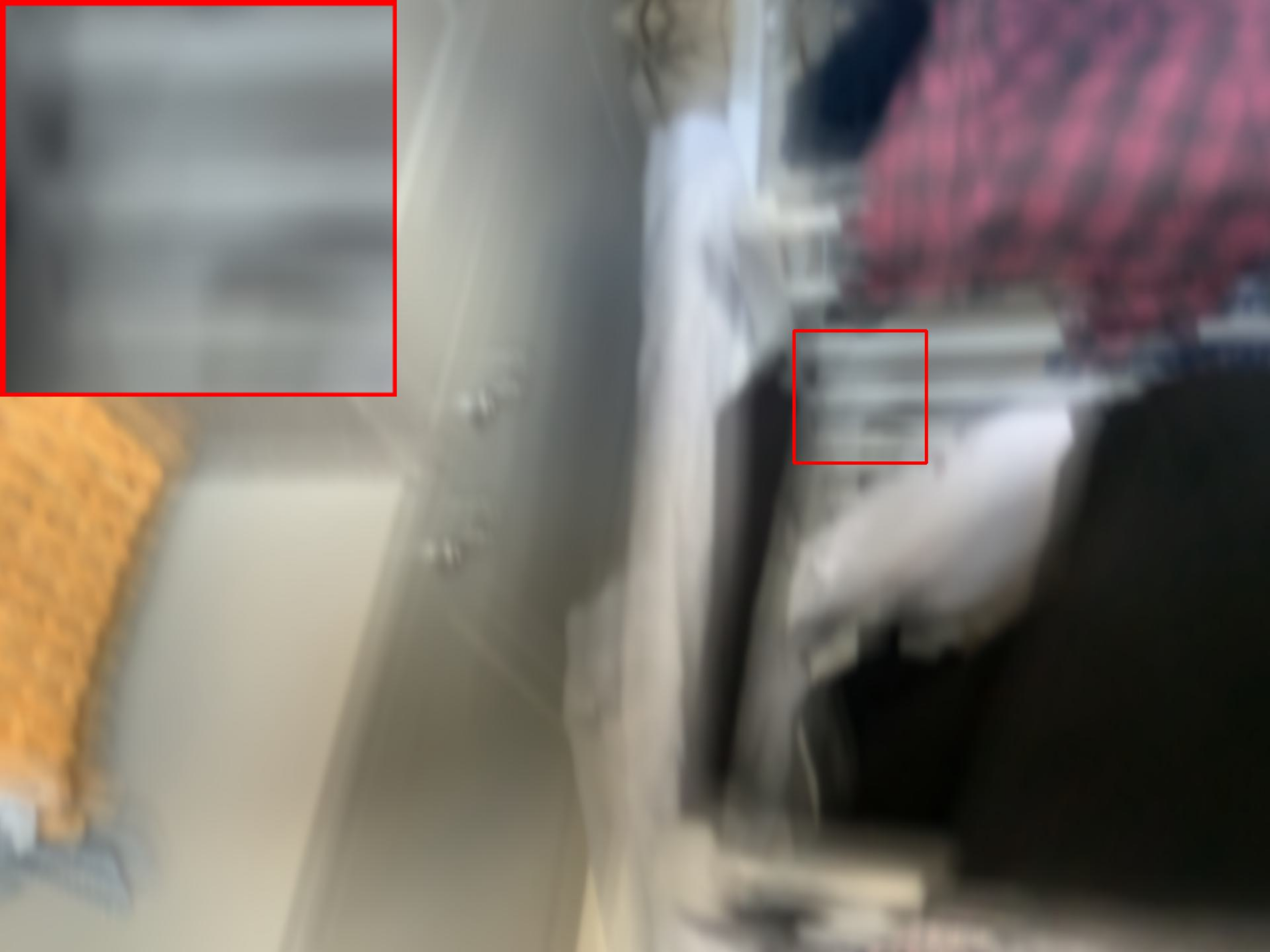}
   \includegraphics[width=0.18\linewidth]{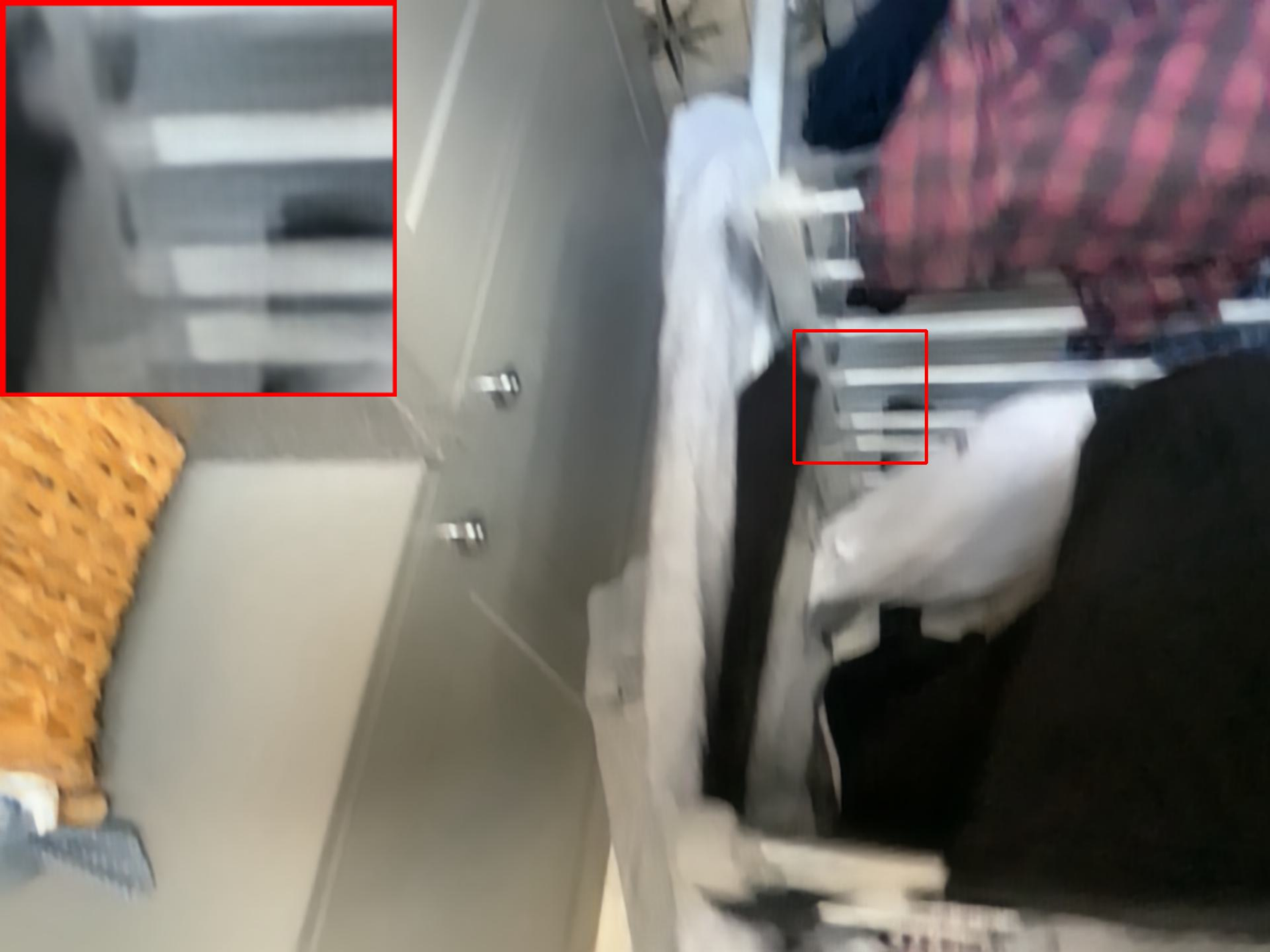}
   \includegraphics[width=0.18\linewidth]{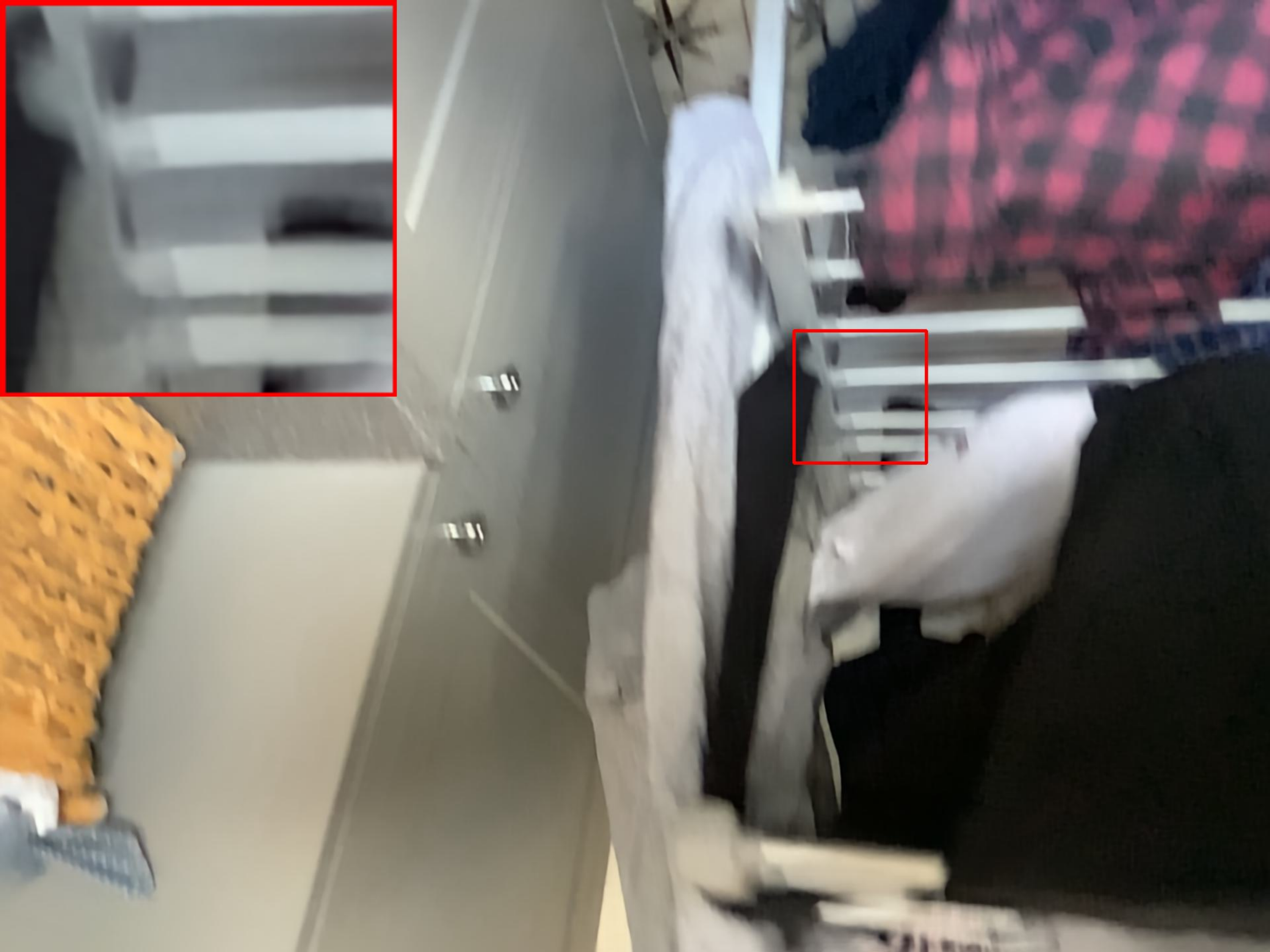}
   \includegraphics[width=0.18\linewidth]{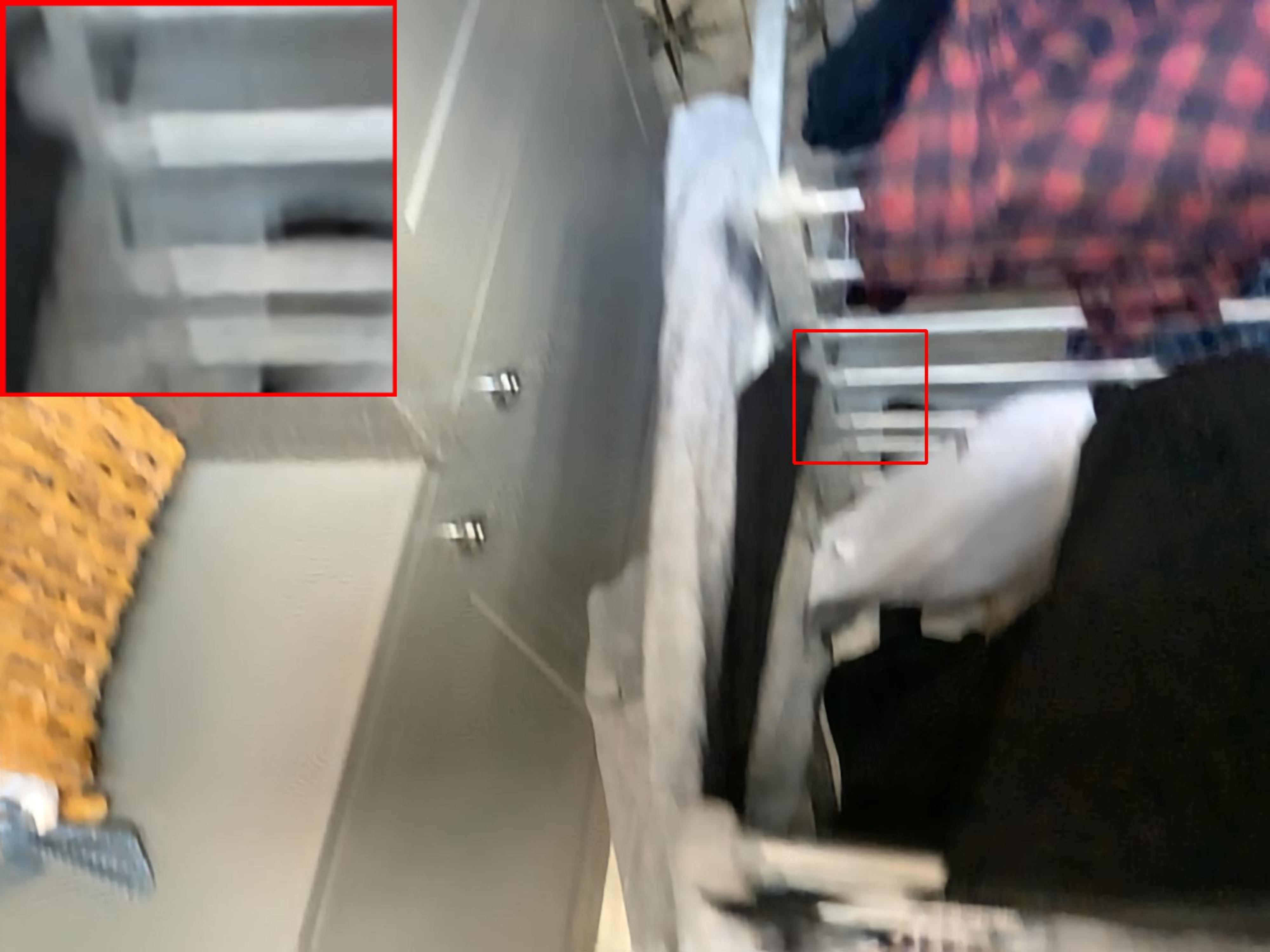}
   \includegraphics[width=0.18\linewidth]{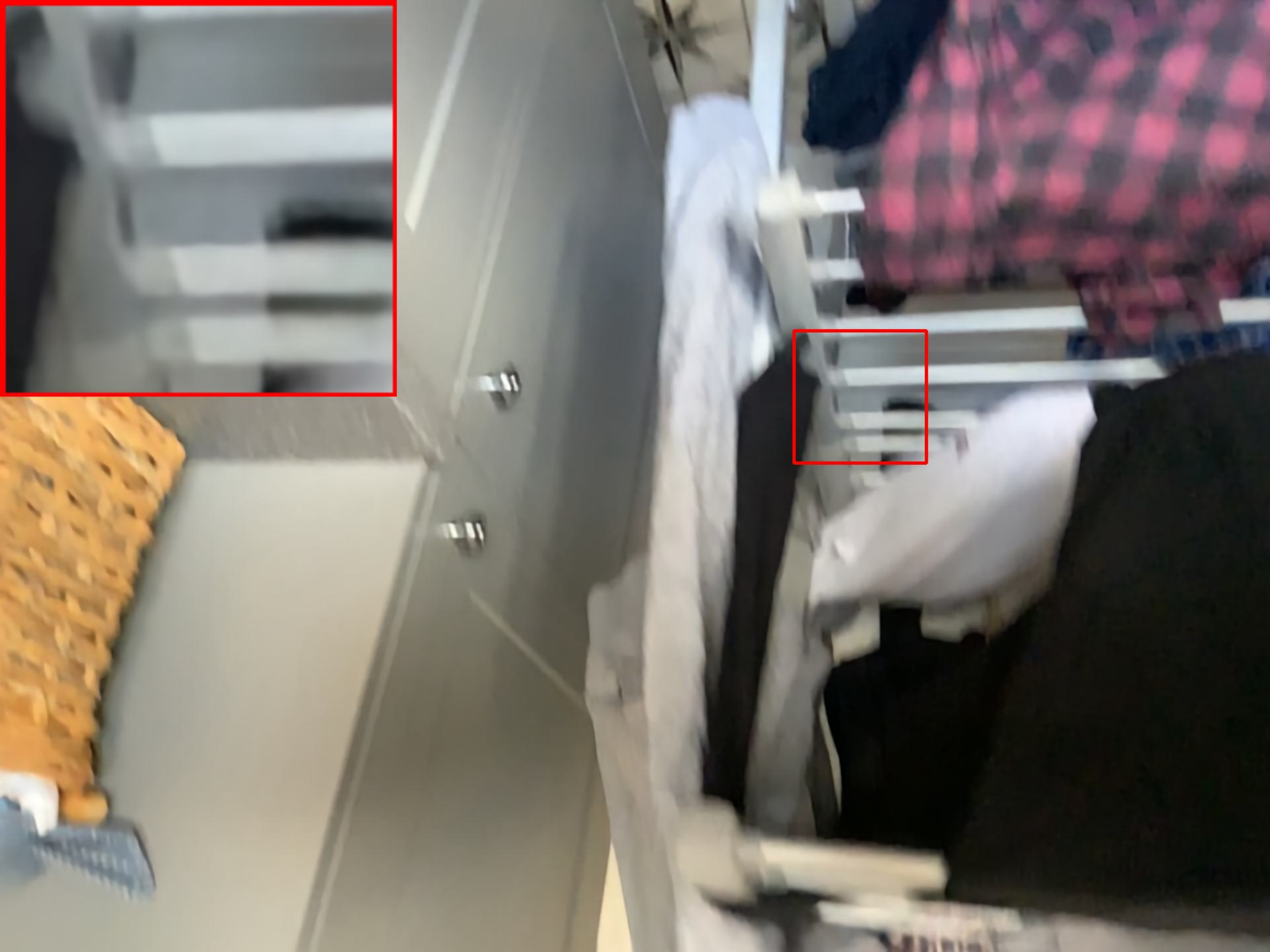} \\[1pt]
   \includegraphics[width=0.18\linewidth]{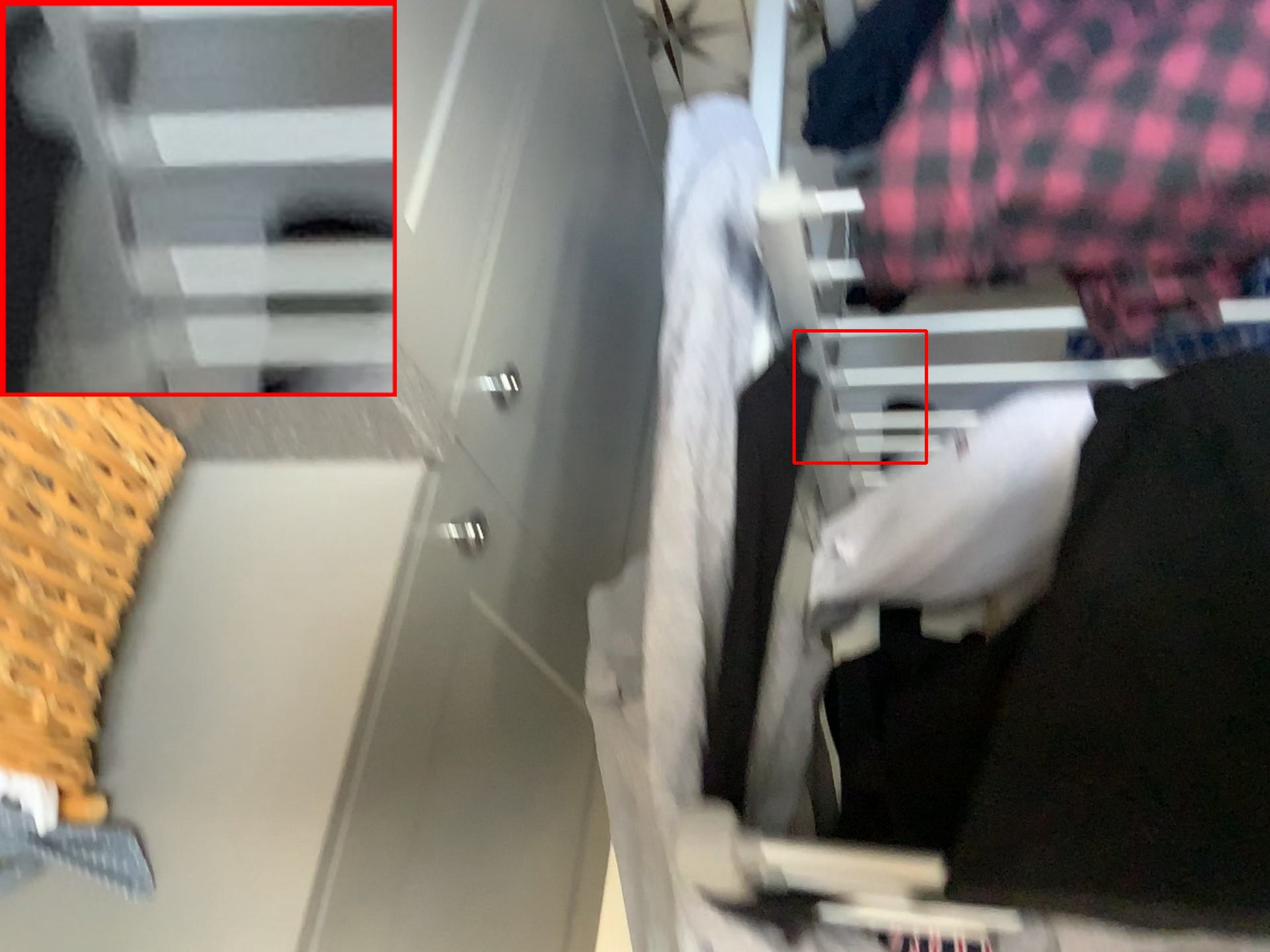}
   \includegraphics[width=0.18\linewidth]{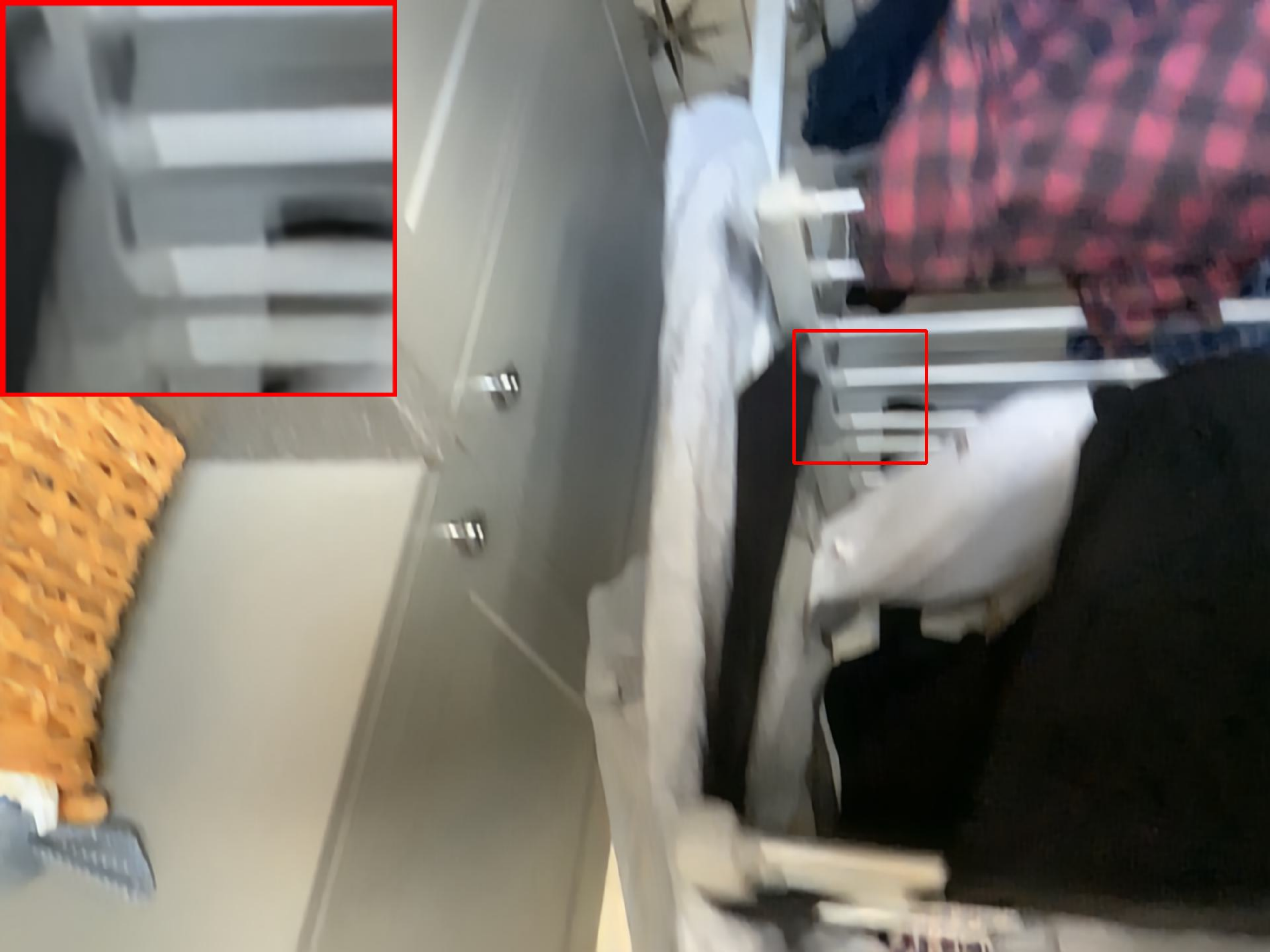}
   \includegraphics[width=0.18\linewidth]{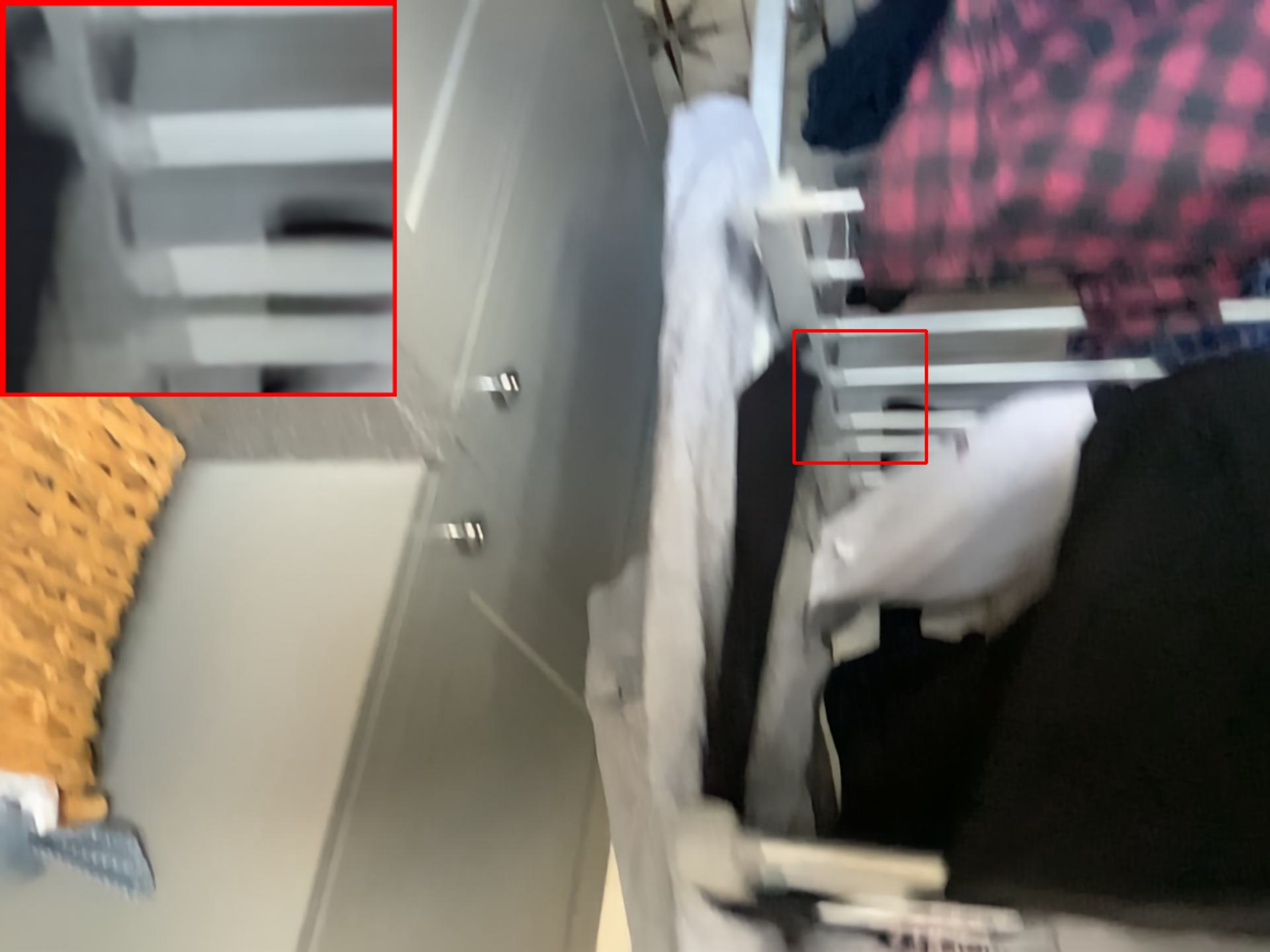}
   \includegraphics[width=0.18\linewidth]{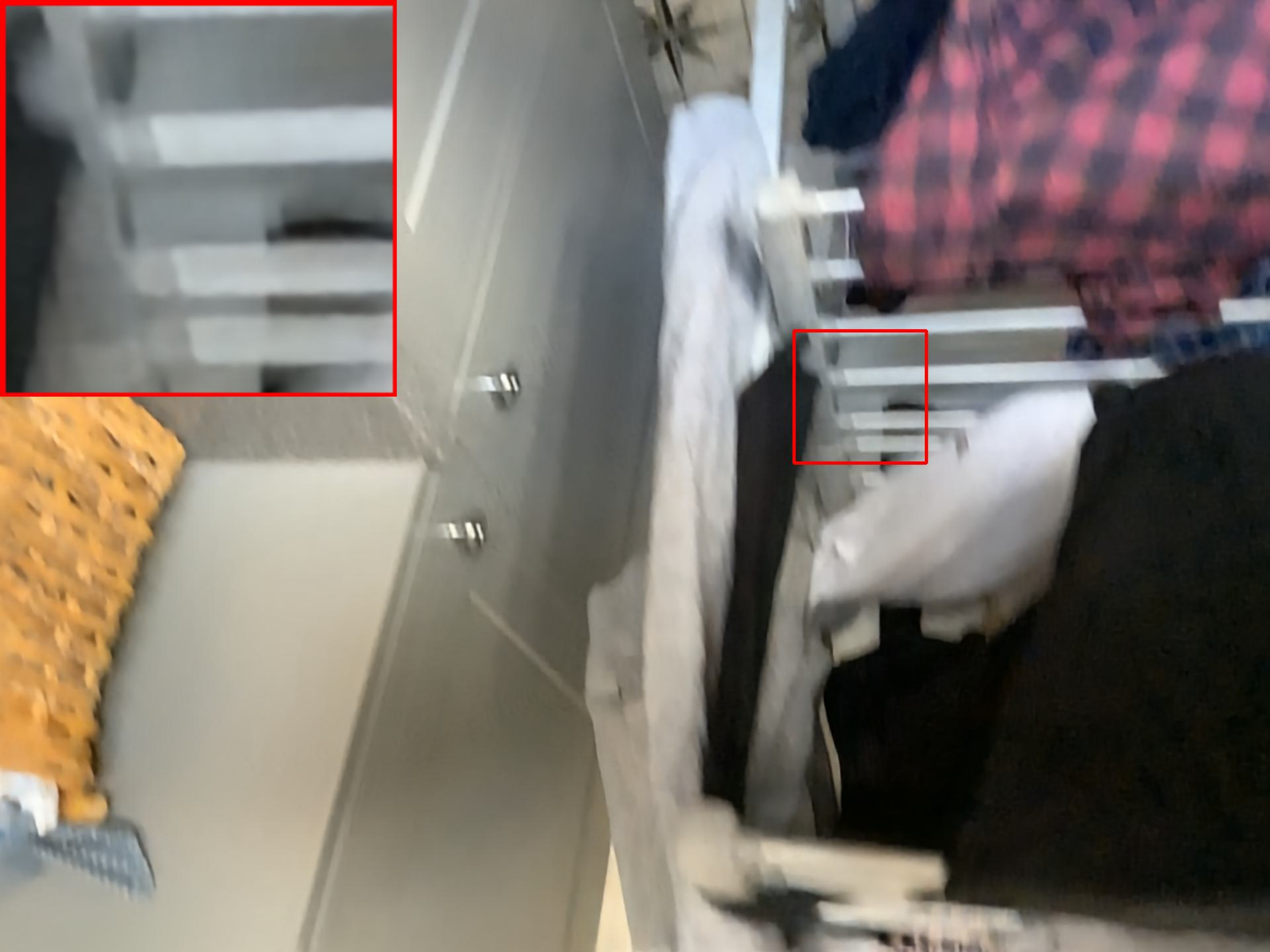}
   \includegraphics[width=0.18\linewidth]{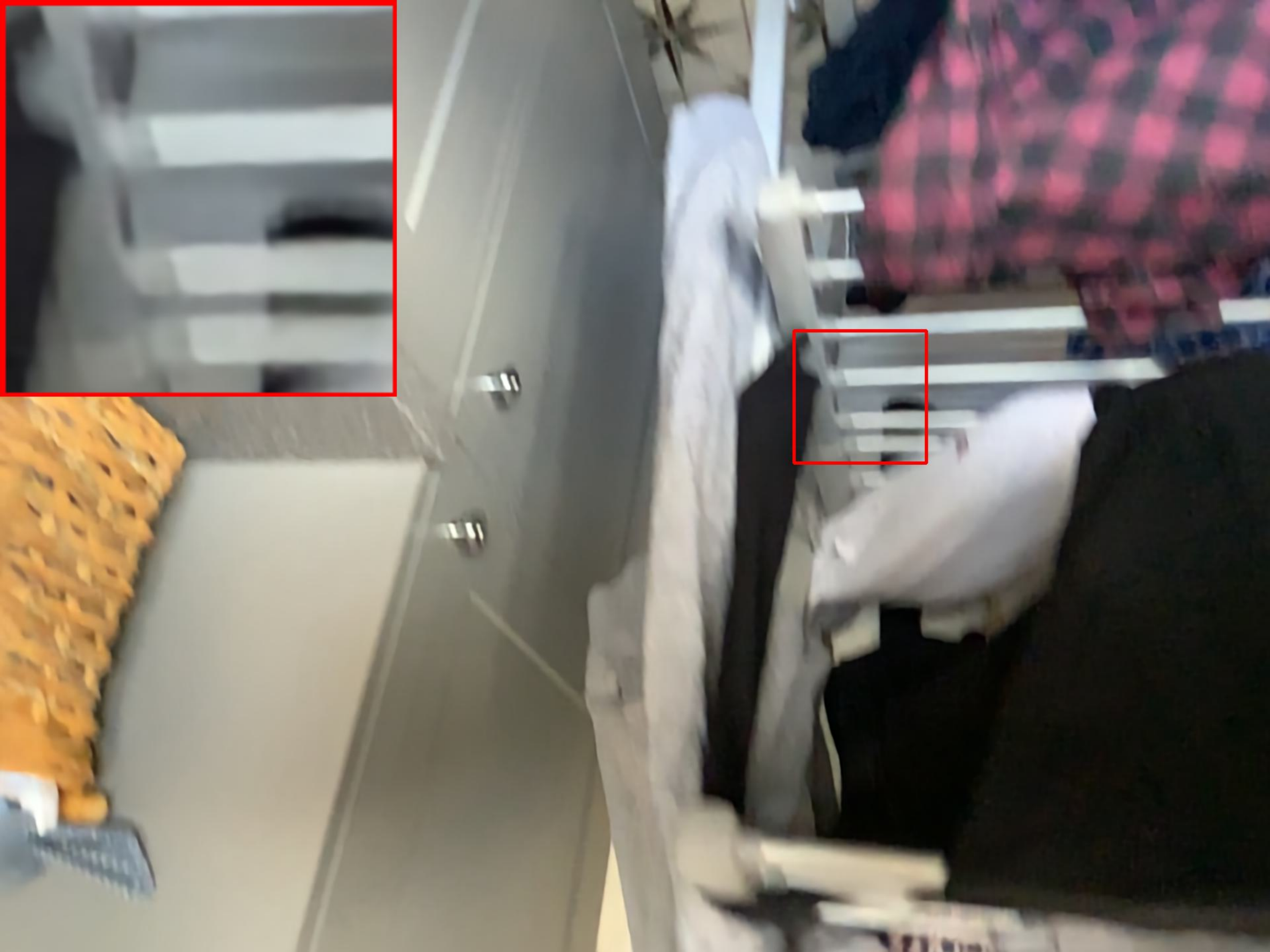} \\[6pt]    
   \includegraphics[width=0.18\linewidth]{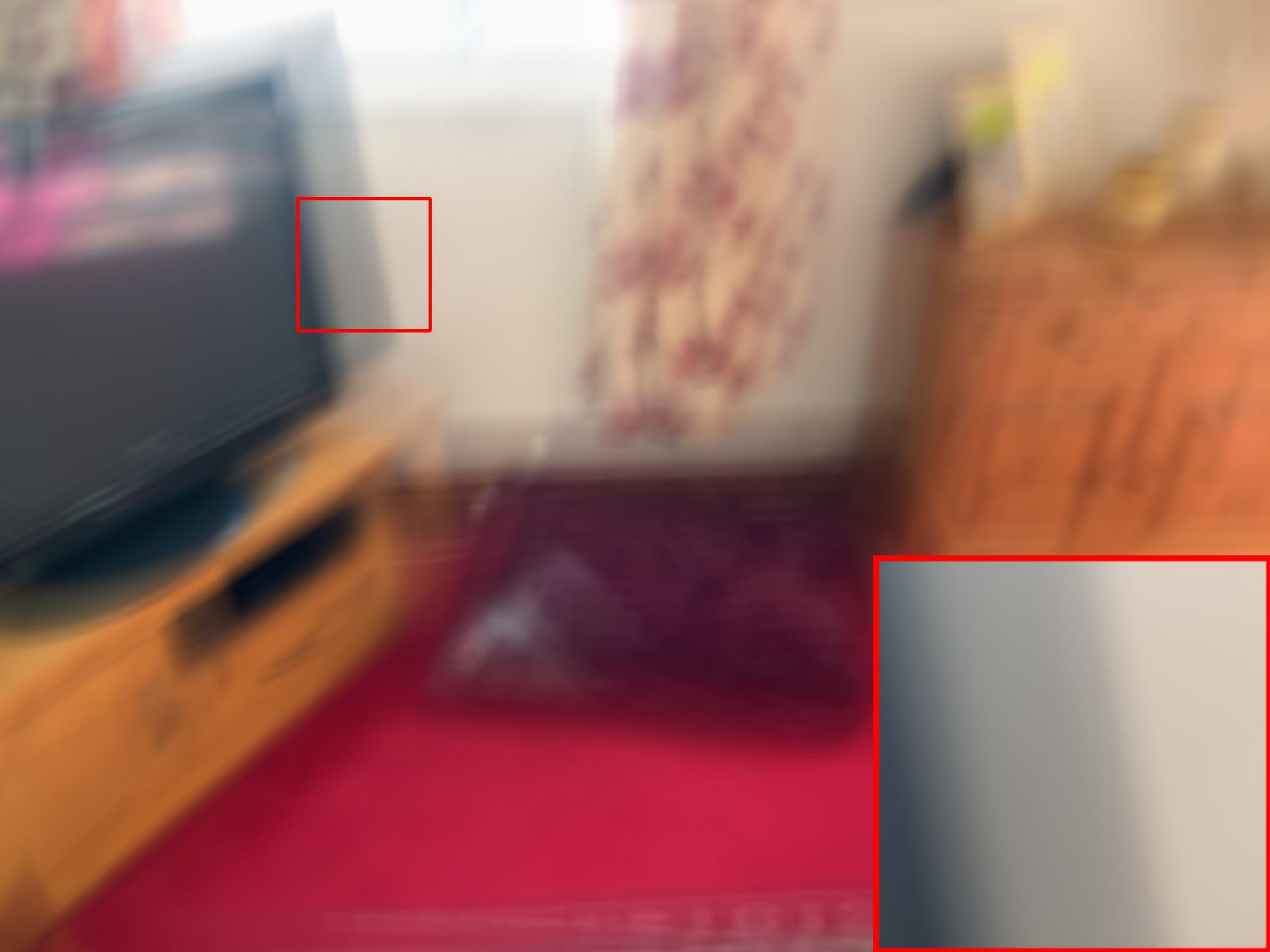}
   \includegraphics[width=0.18\linewidth]{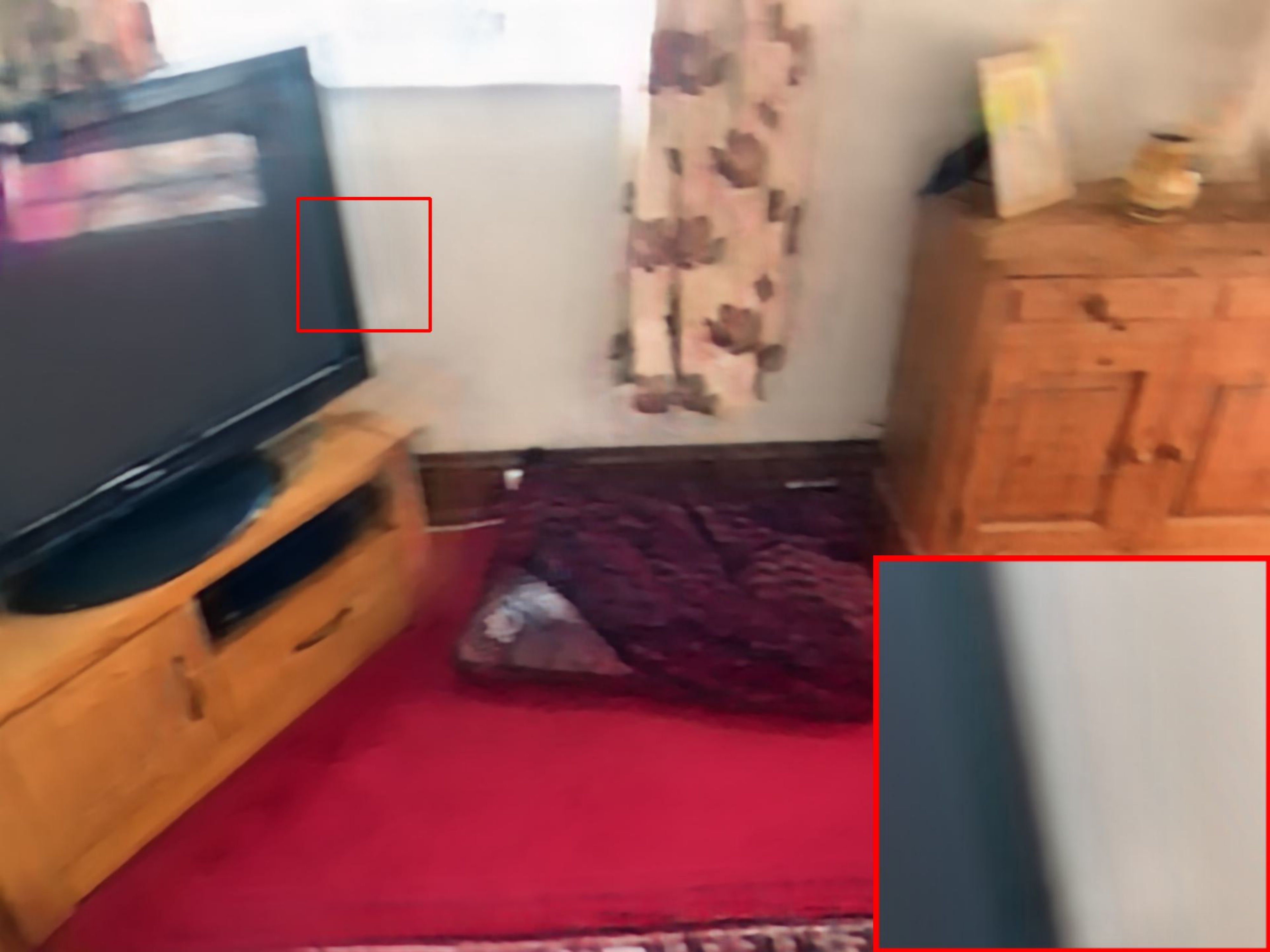}
   \includegraphics[width=0.18\linewidth]{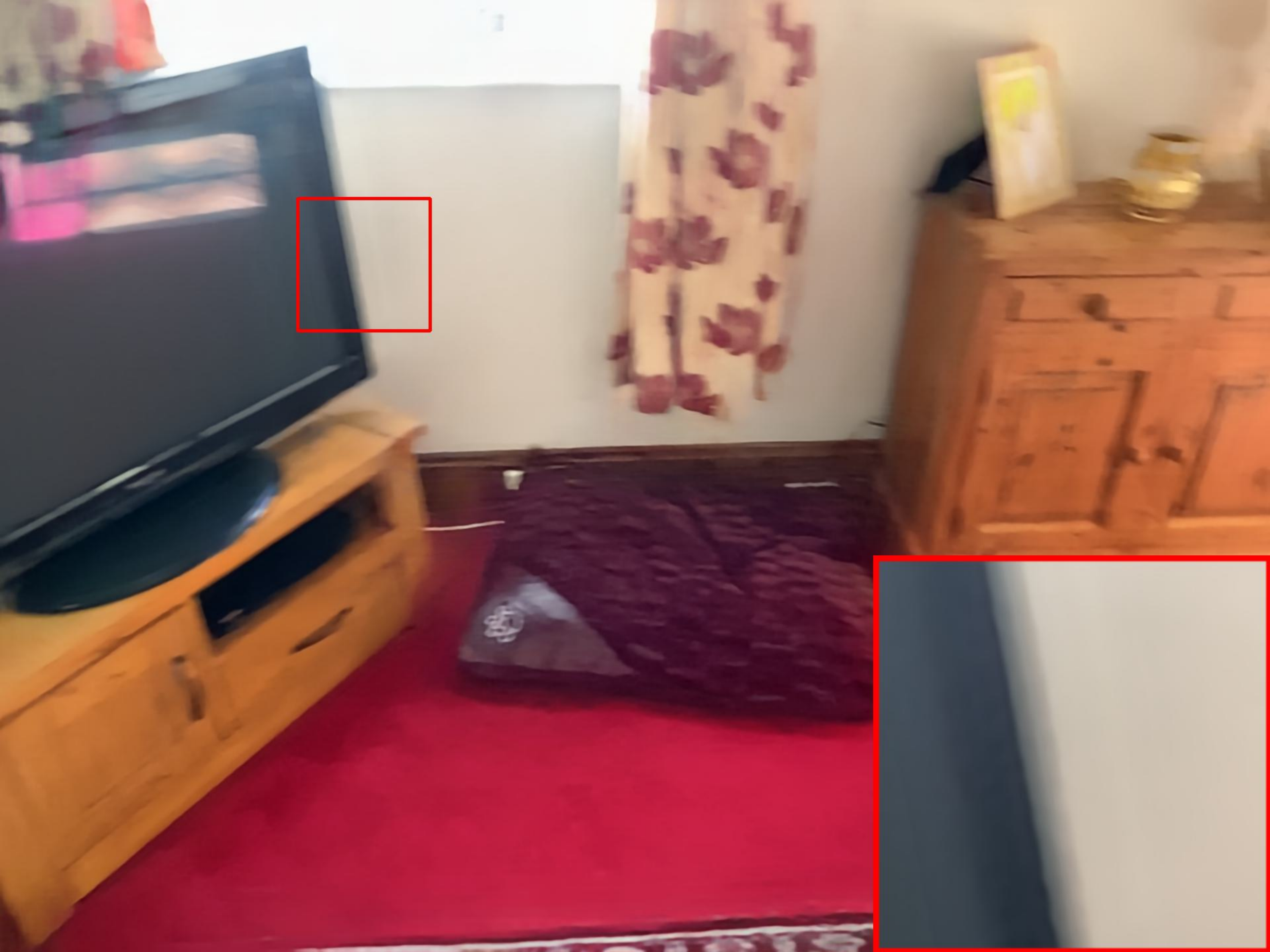}
   \includegraphics[width=0.18\linewidth]{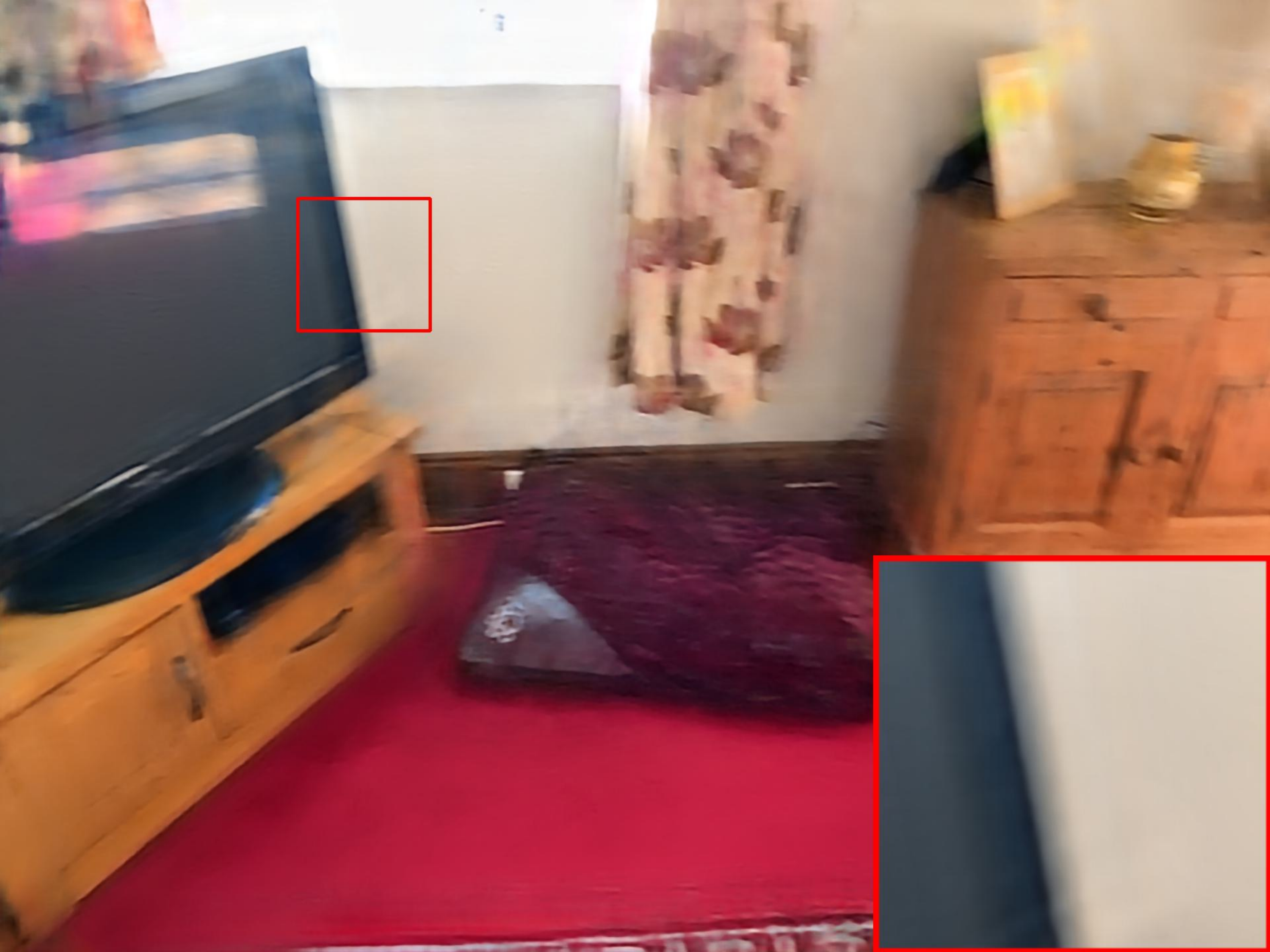}
   \includegraphics[width=0.18\linewidth]{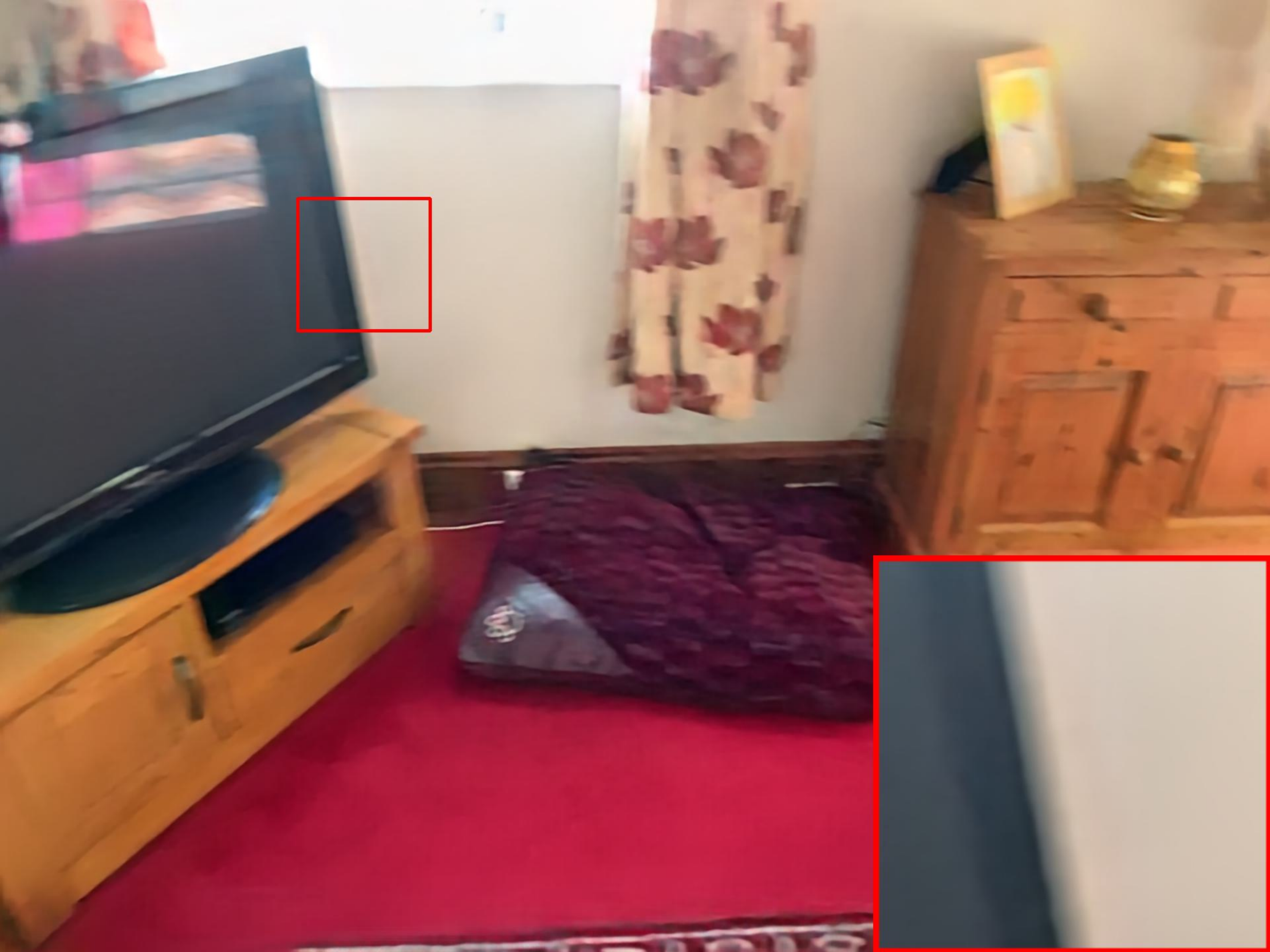} \\[1pt]
   \includegraphics[width=0.18\linewidth]{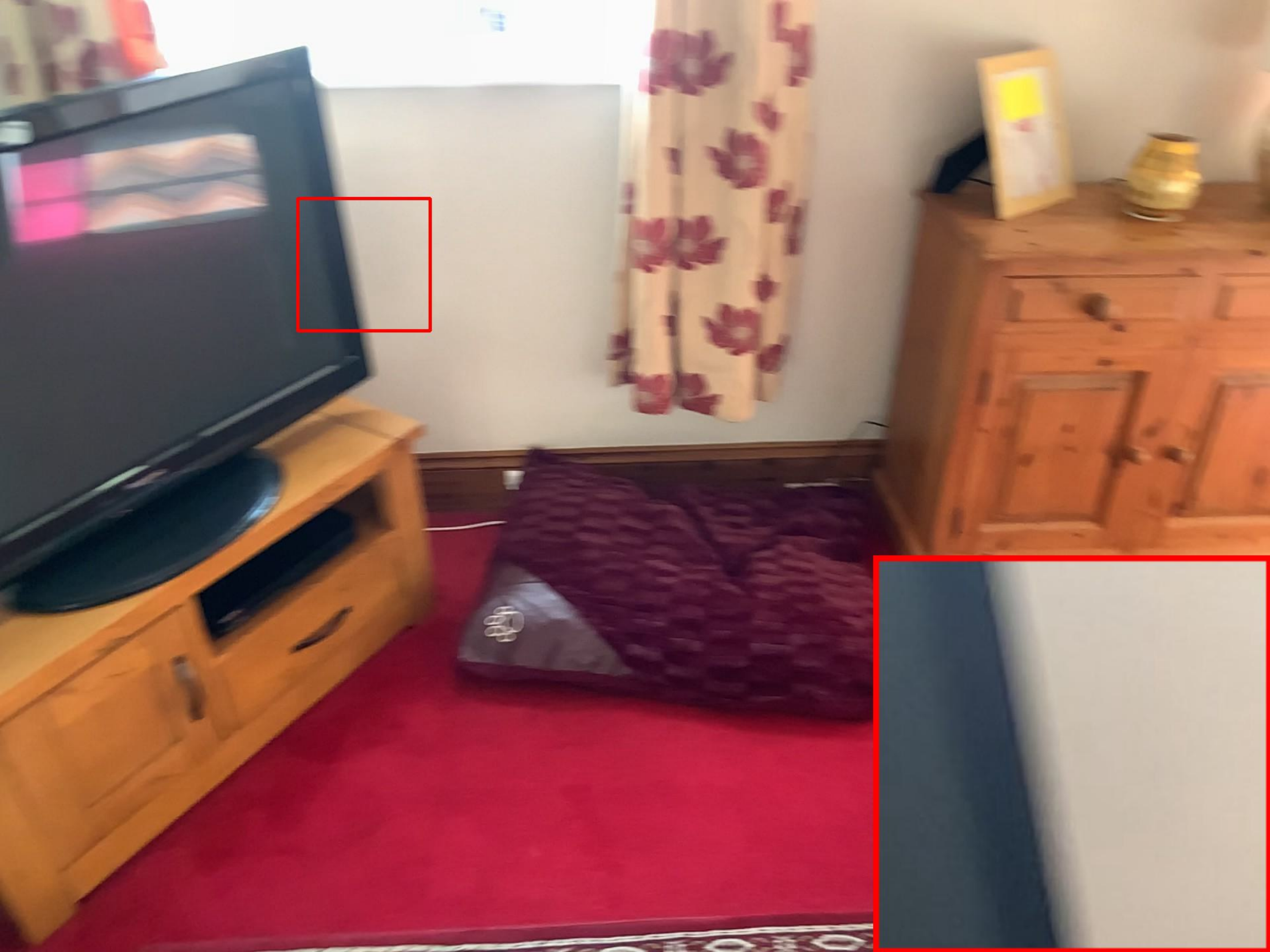}
   \includegraphics[width=0.18\linewidth]{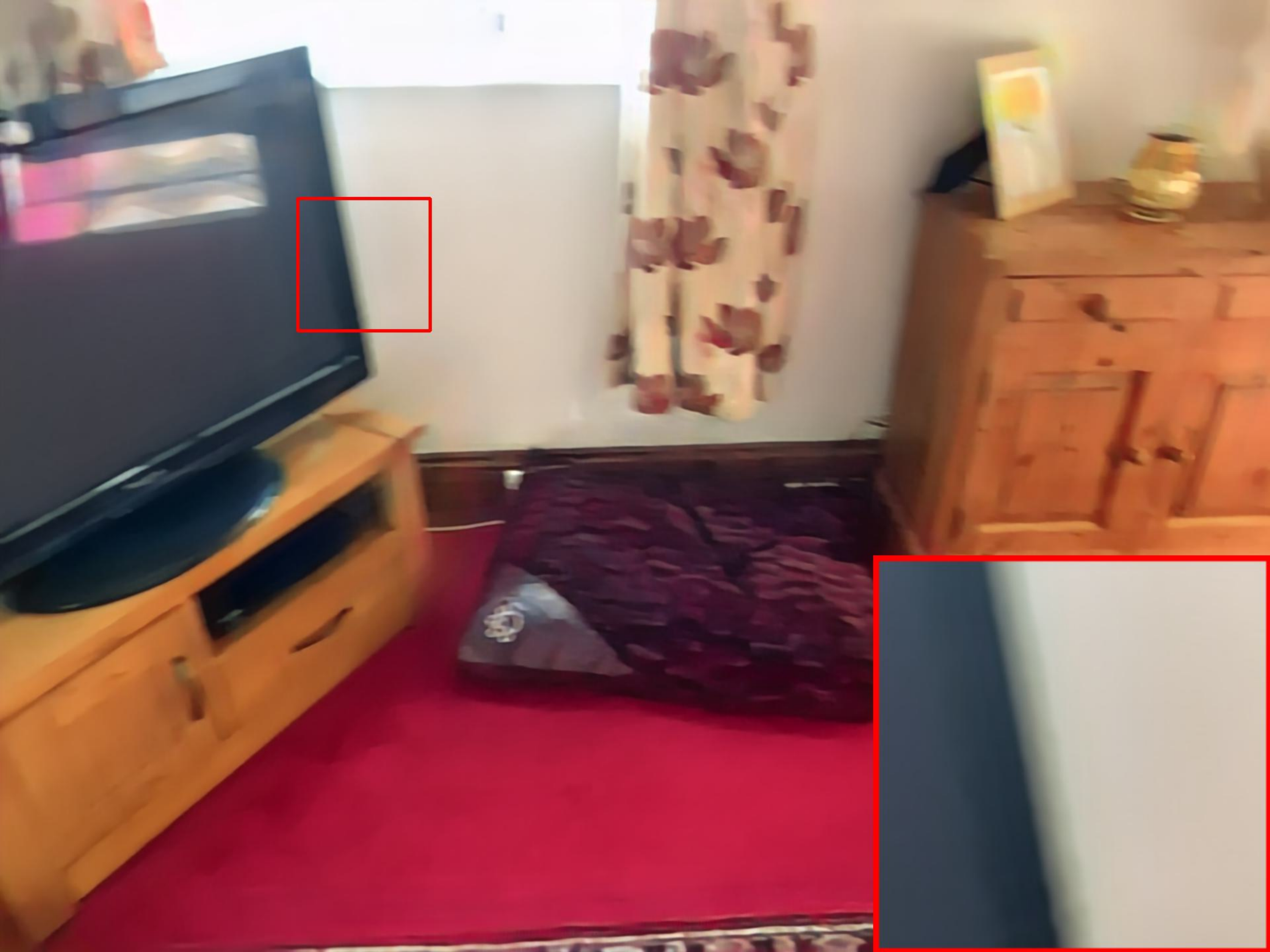}
   \includegraphics[width=0.18\linewidth]{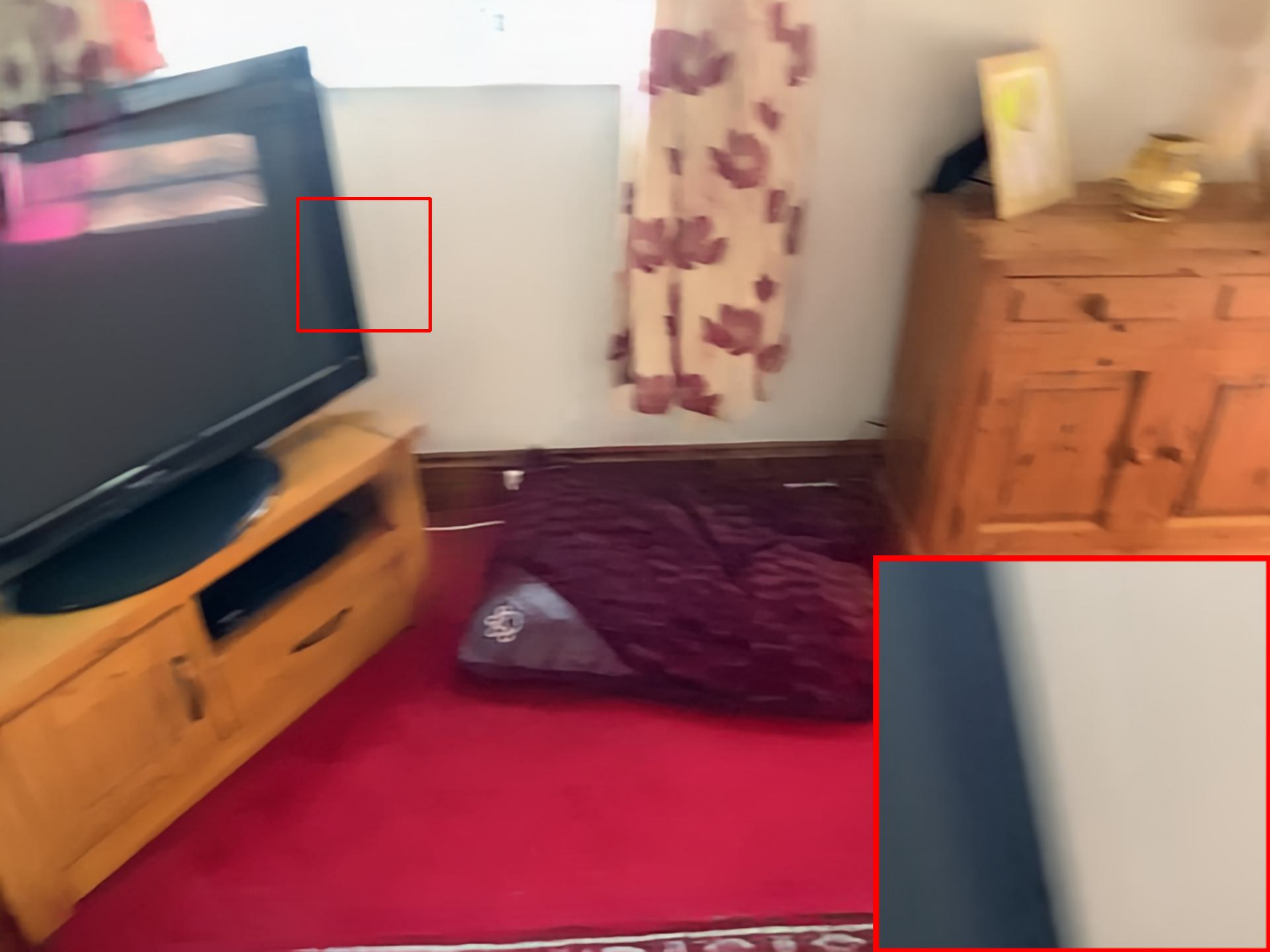}
   \includegraphics[width=0.18\linewidth]{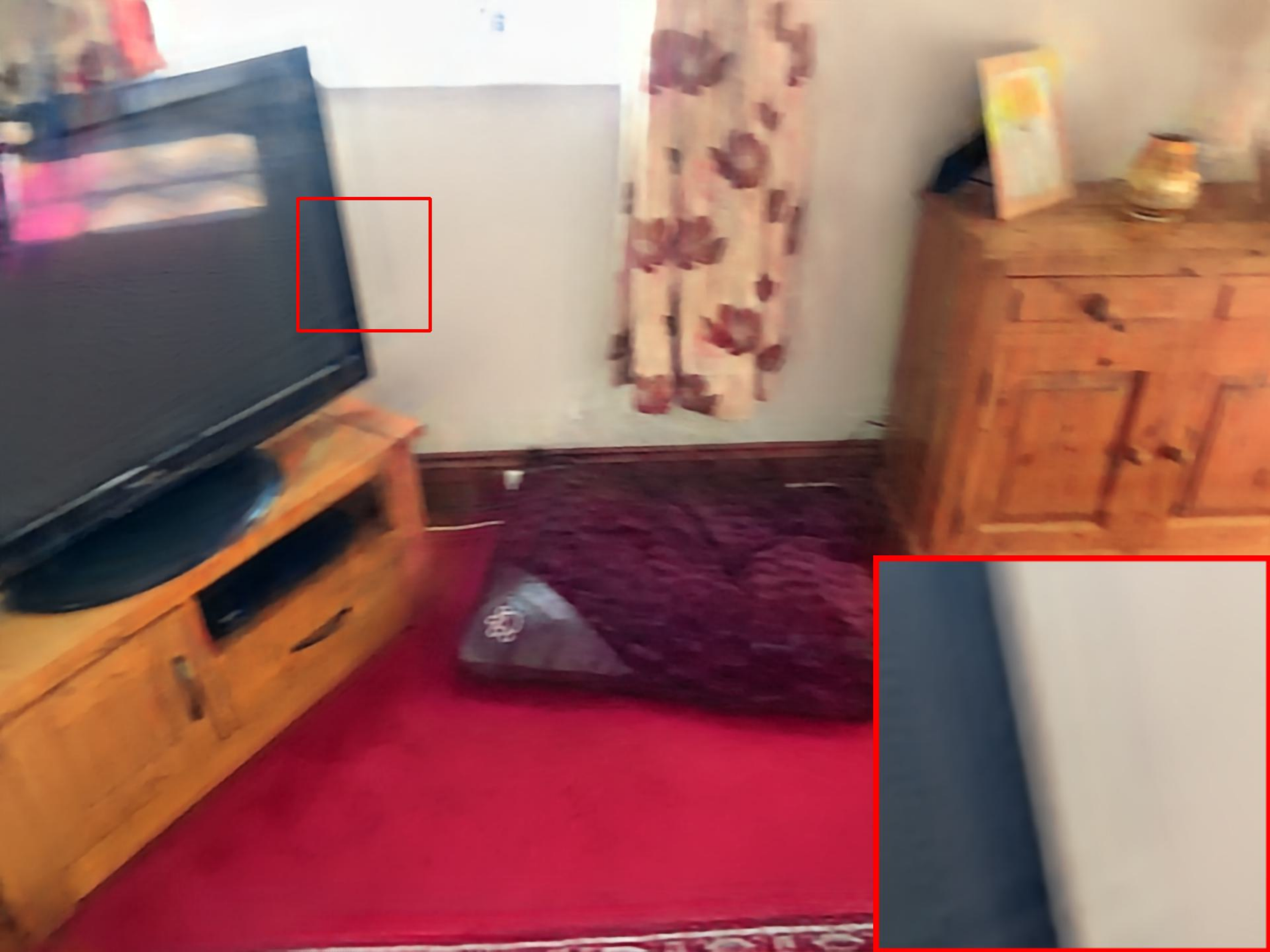}
   \includegraphics[width=0.18\linewidth]{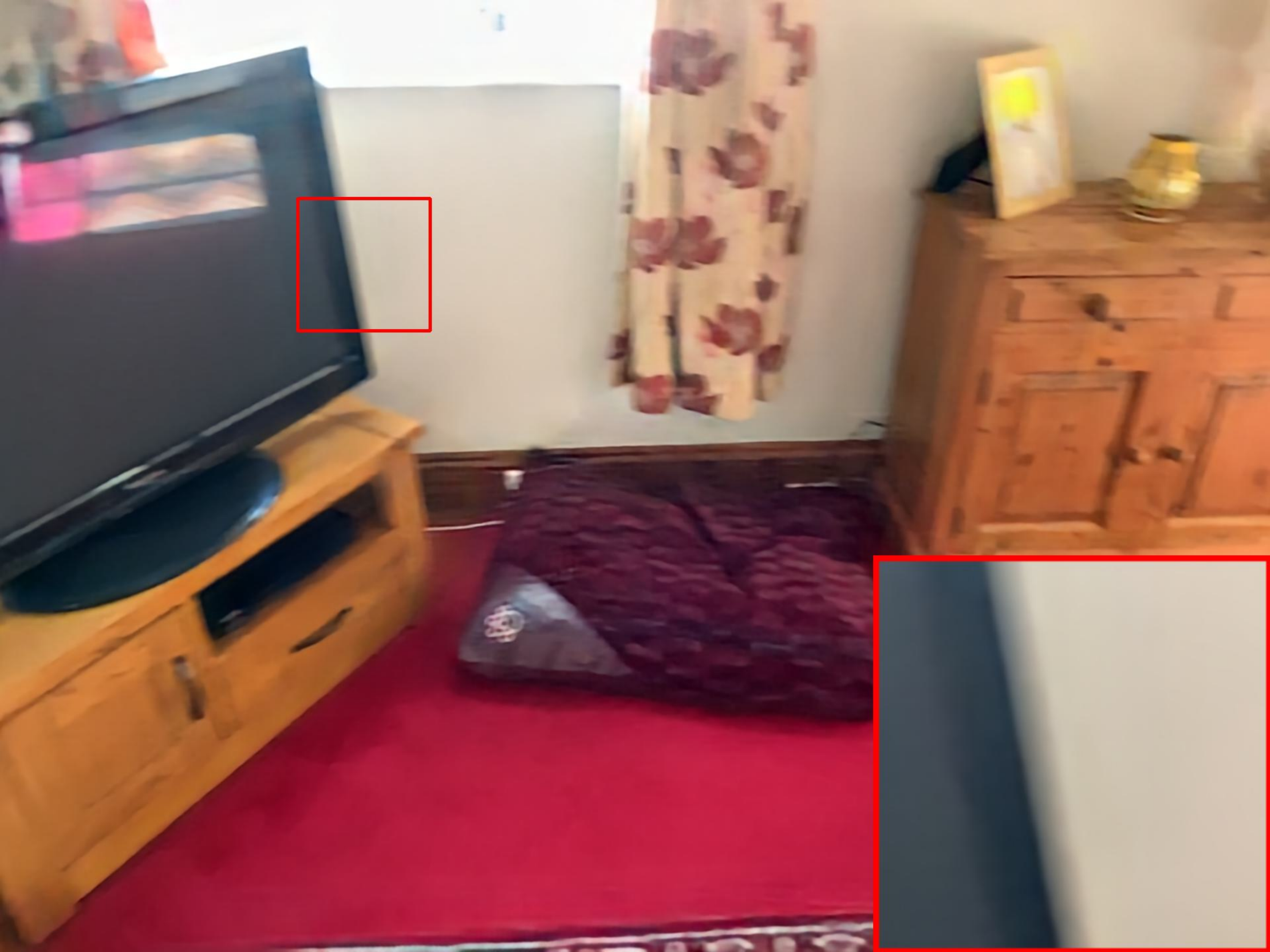} \\[6pt]     
   \includegraphics[width=0.18\linewidth]{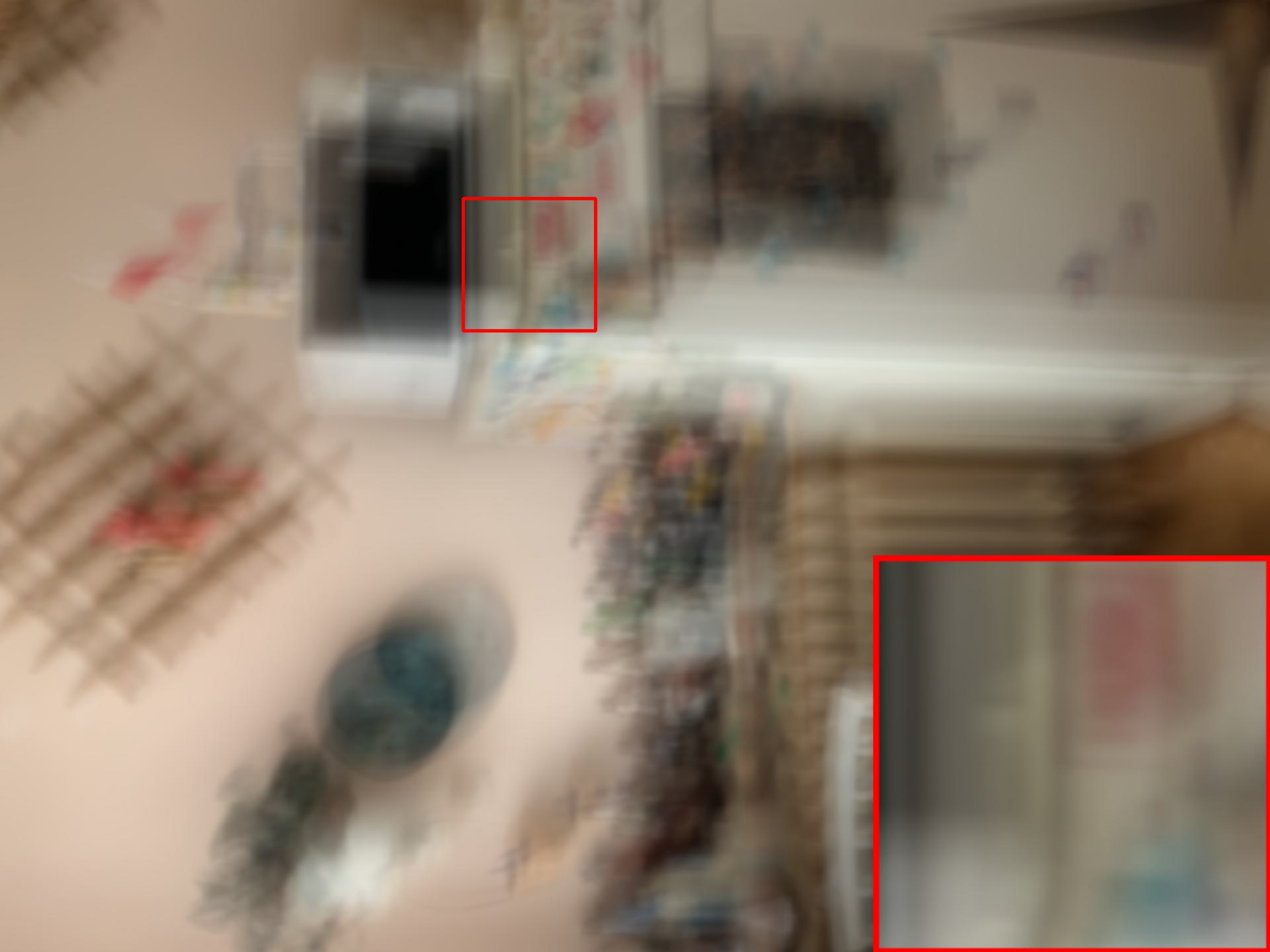}
   \includegraphics[width=0.18\linewidth]{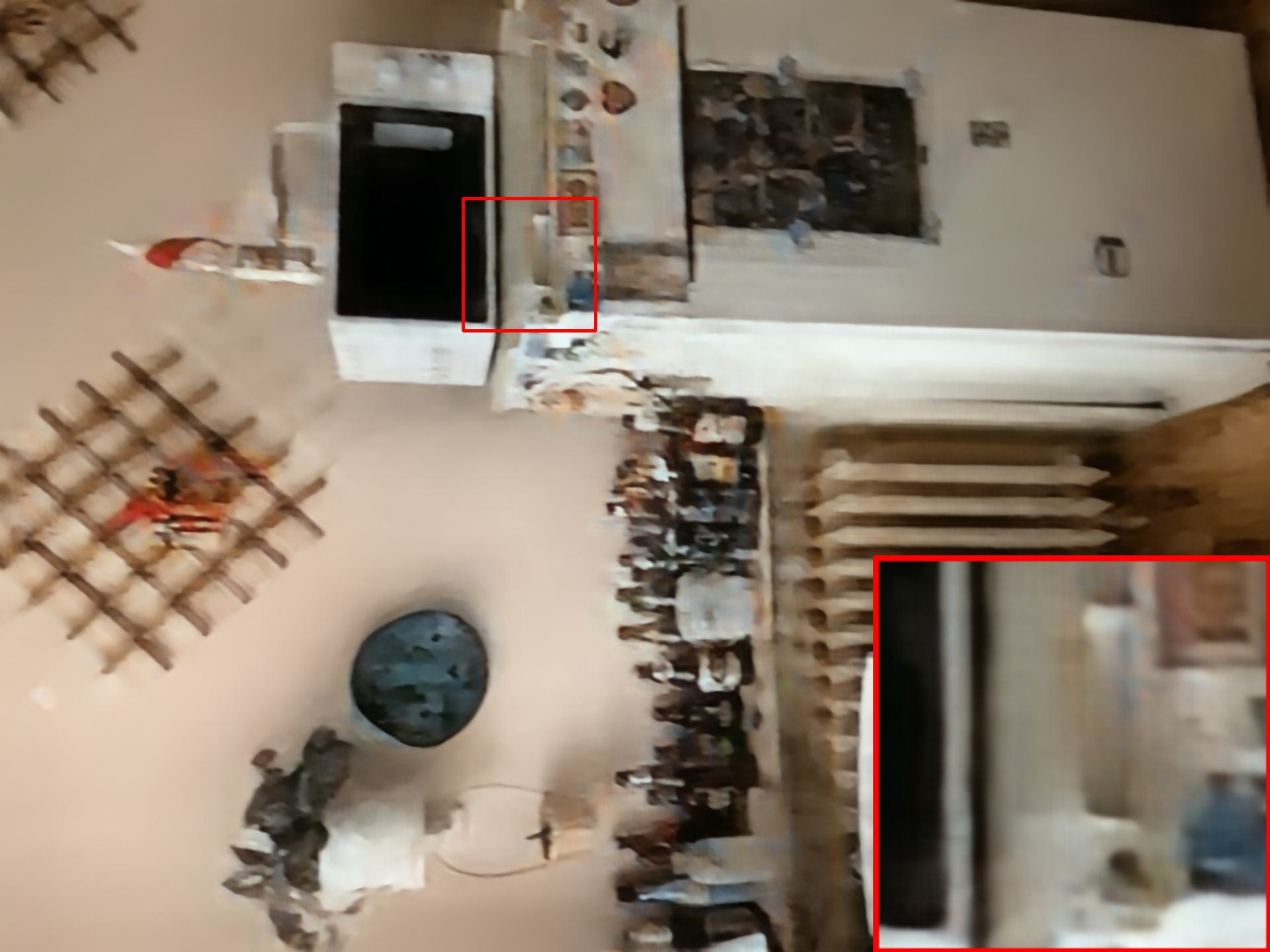}
   \includegraphics[width=0.18\linewidth]{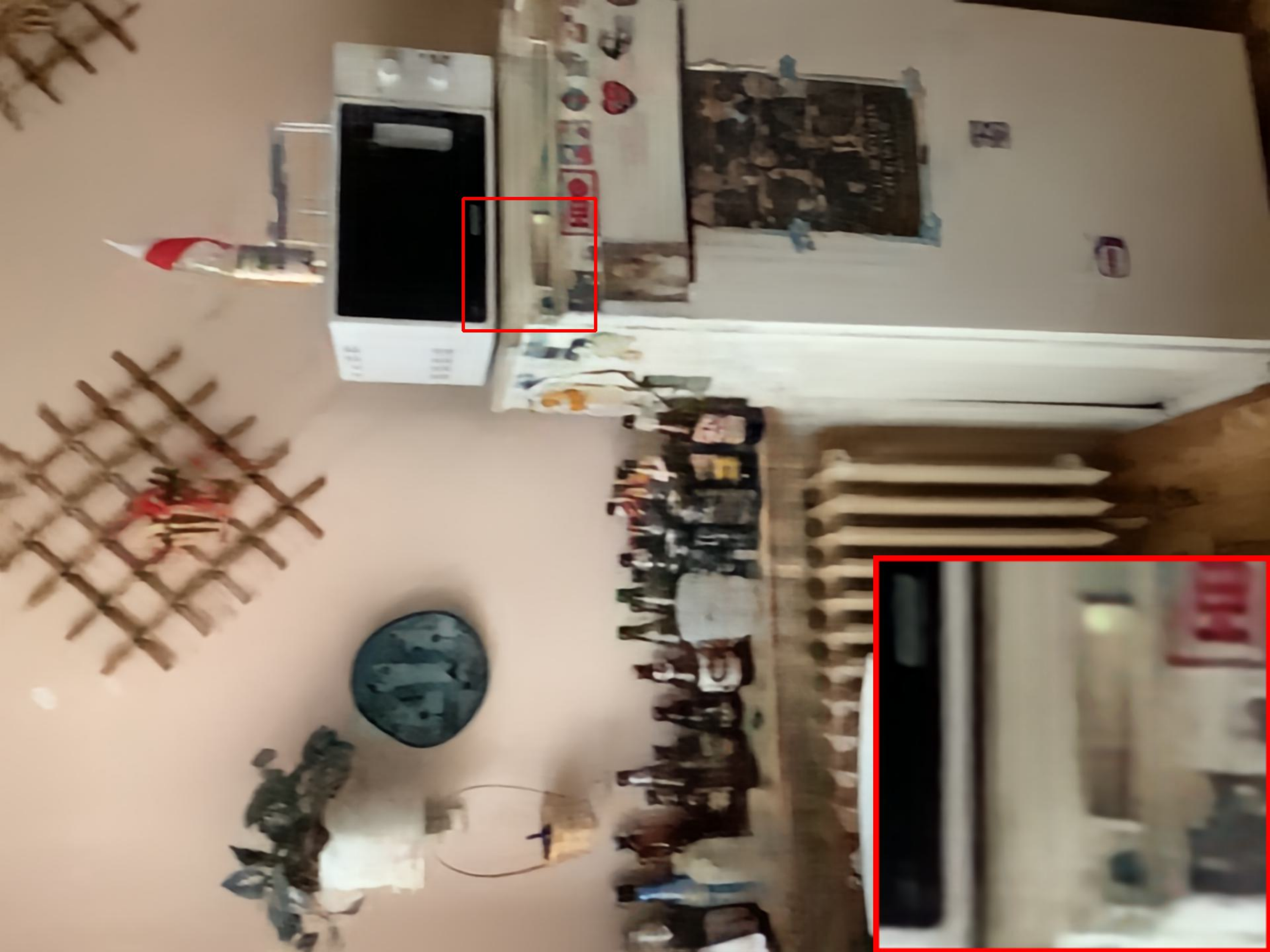}
   \includegraphics[width=0.18\linewidth]{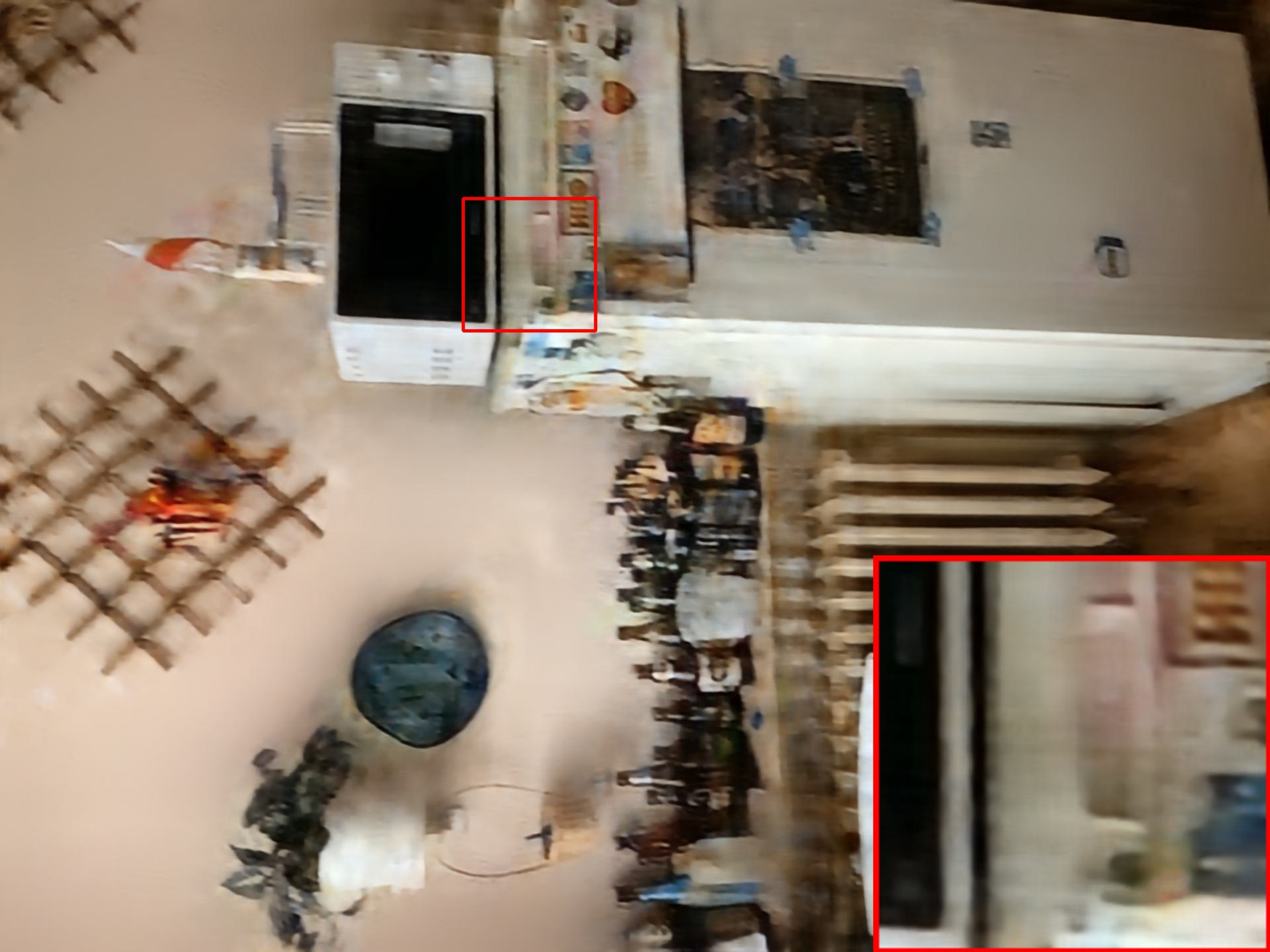}
   \includegraphics[width=0.18\linewidth]{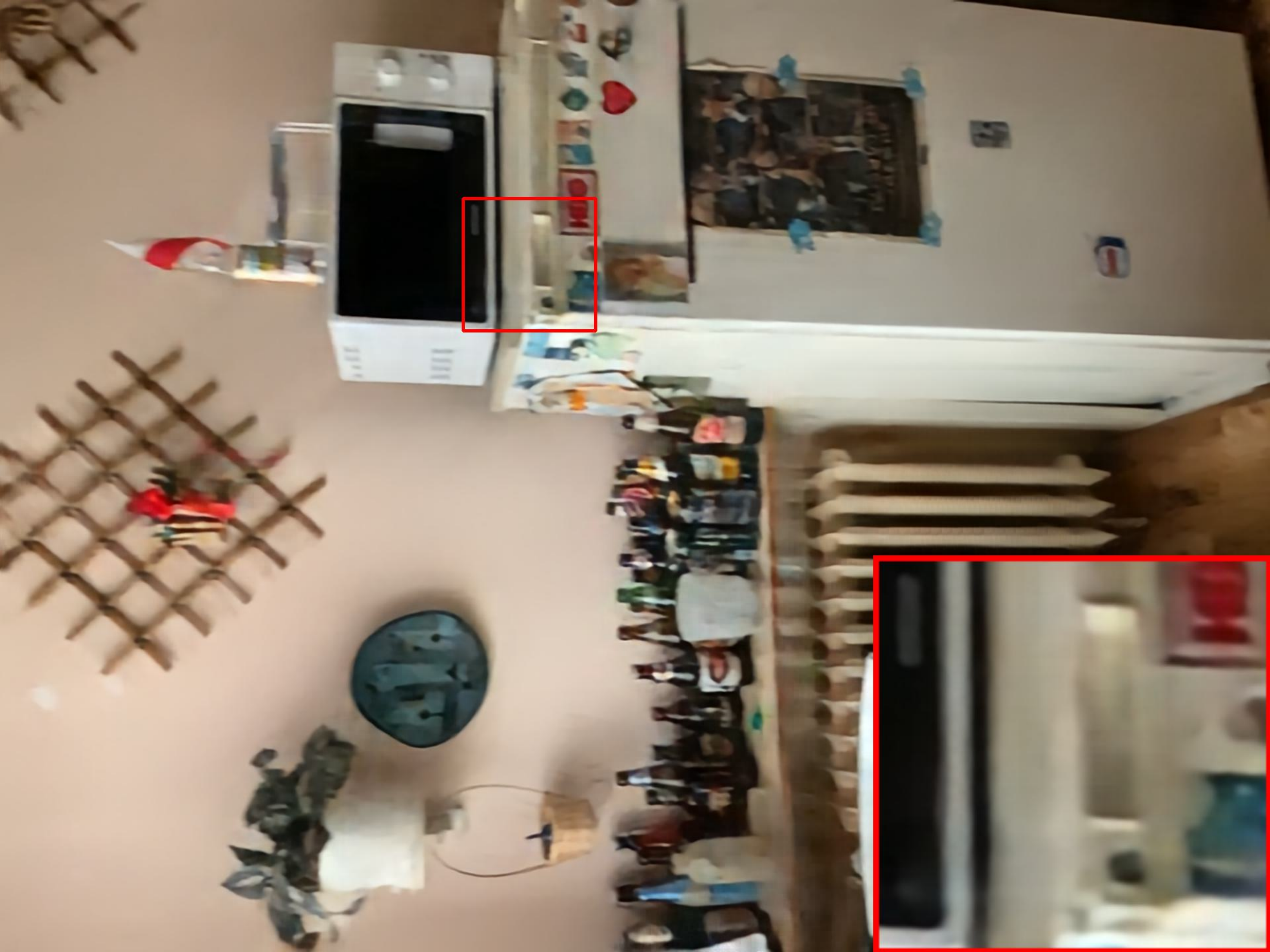} \\[1pt]
   \includegraphics[width=0.18\linewidth]{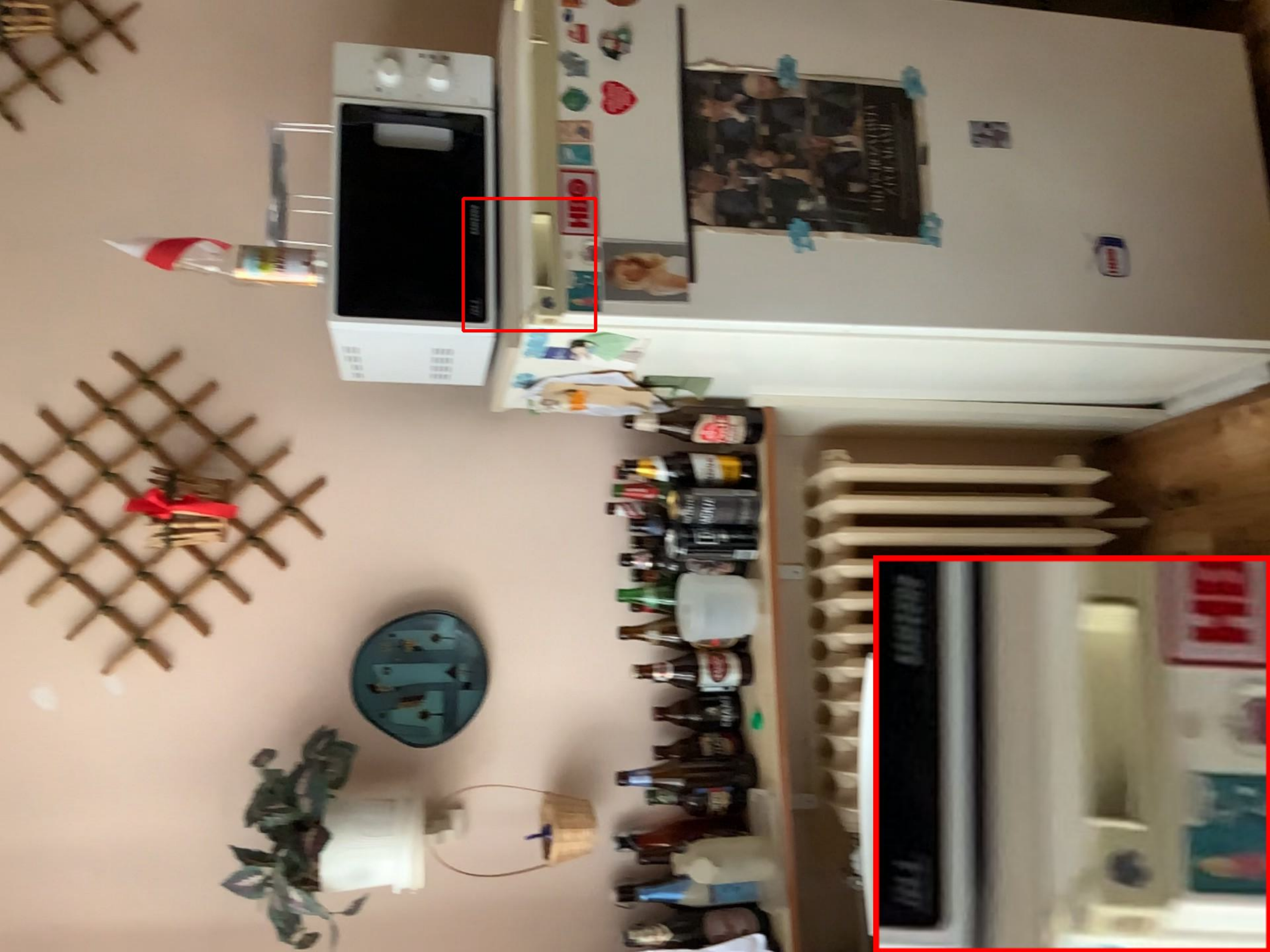}
   \includegraphics[width=0.18\linewidth]{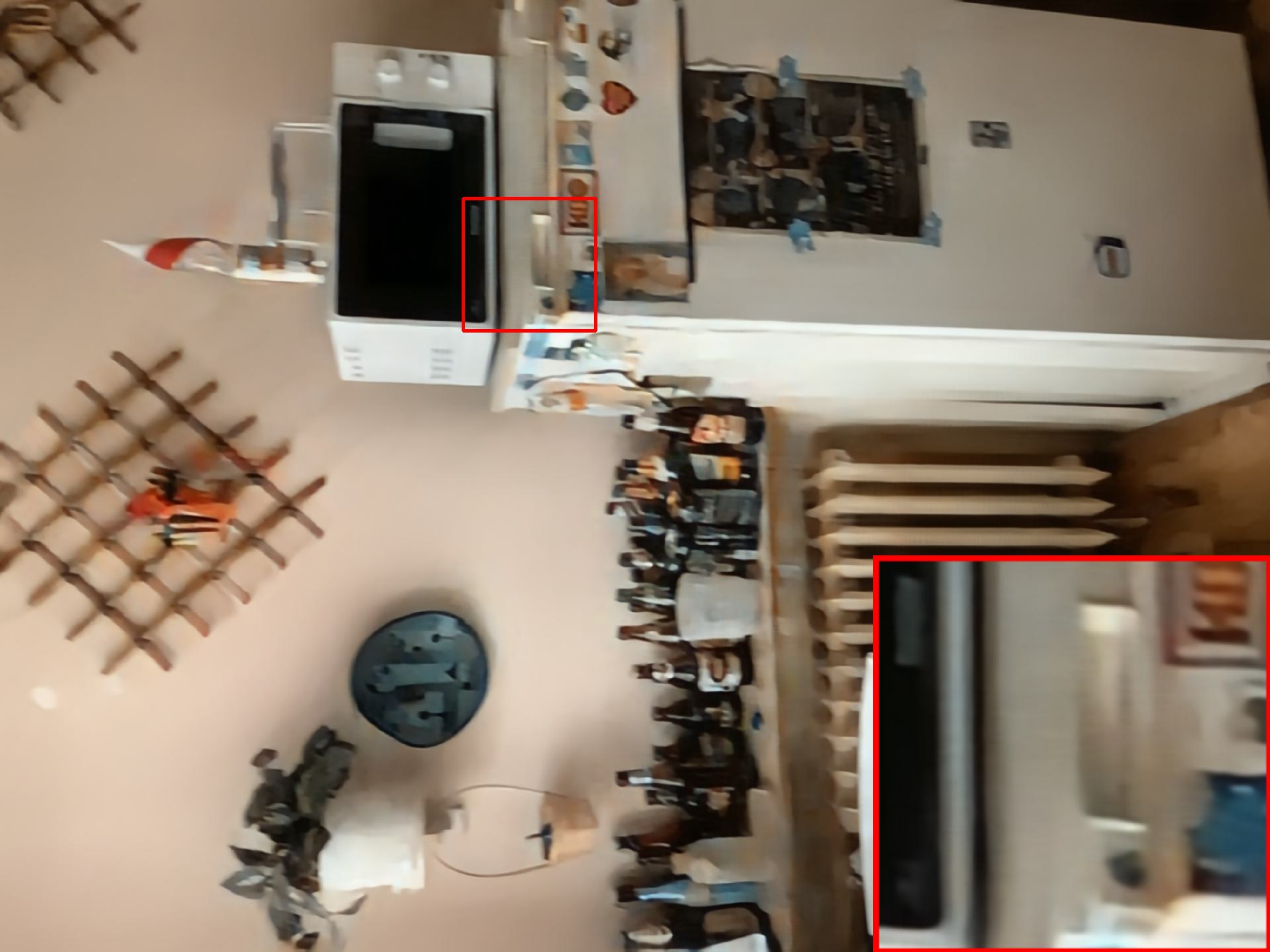}
   \includegraphics[width=0.18\linewidth]{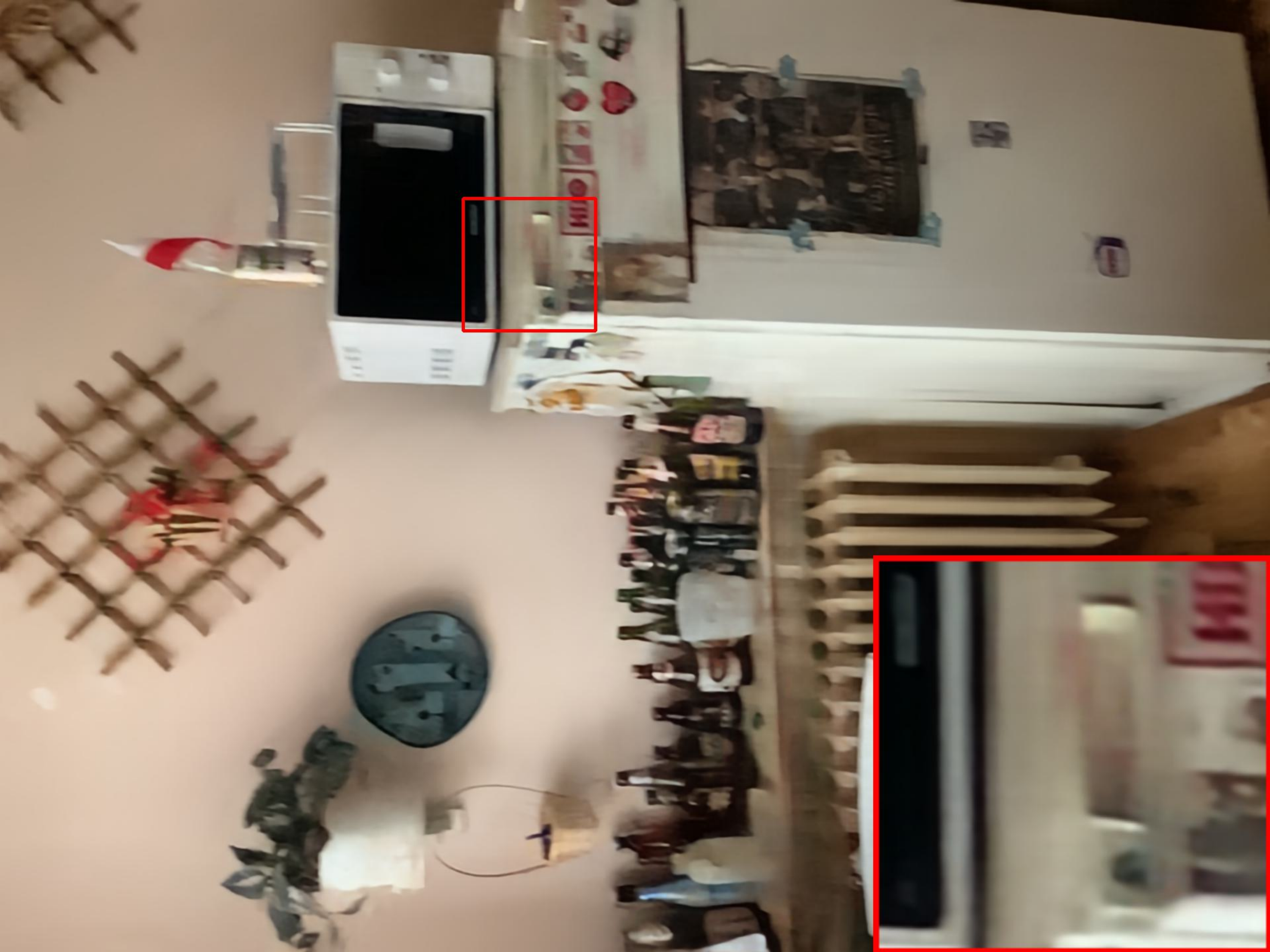}
   \includegraphics[width=0.18\linewidth]{figures/figureresult/47331653_23107.364/depthRestormer.pdf}
   \includegraphics[width=0.18\linewidth]{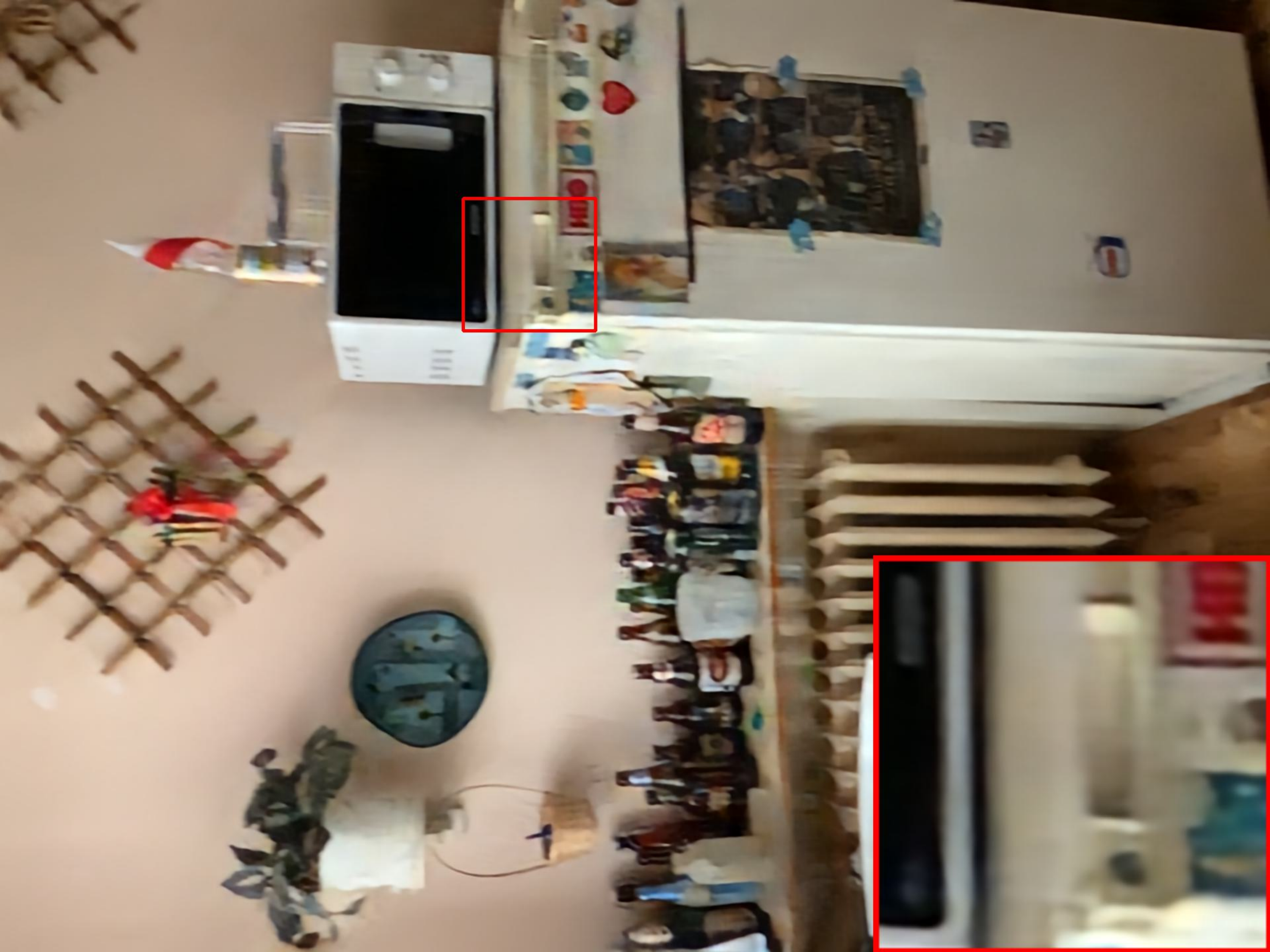} \\[6pt]     
   \includegraphics[width=0.18\linewidth]{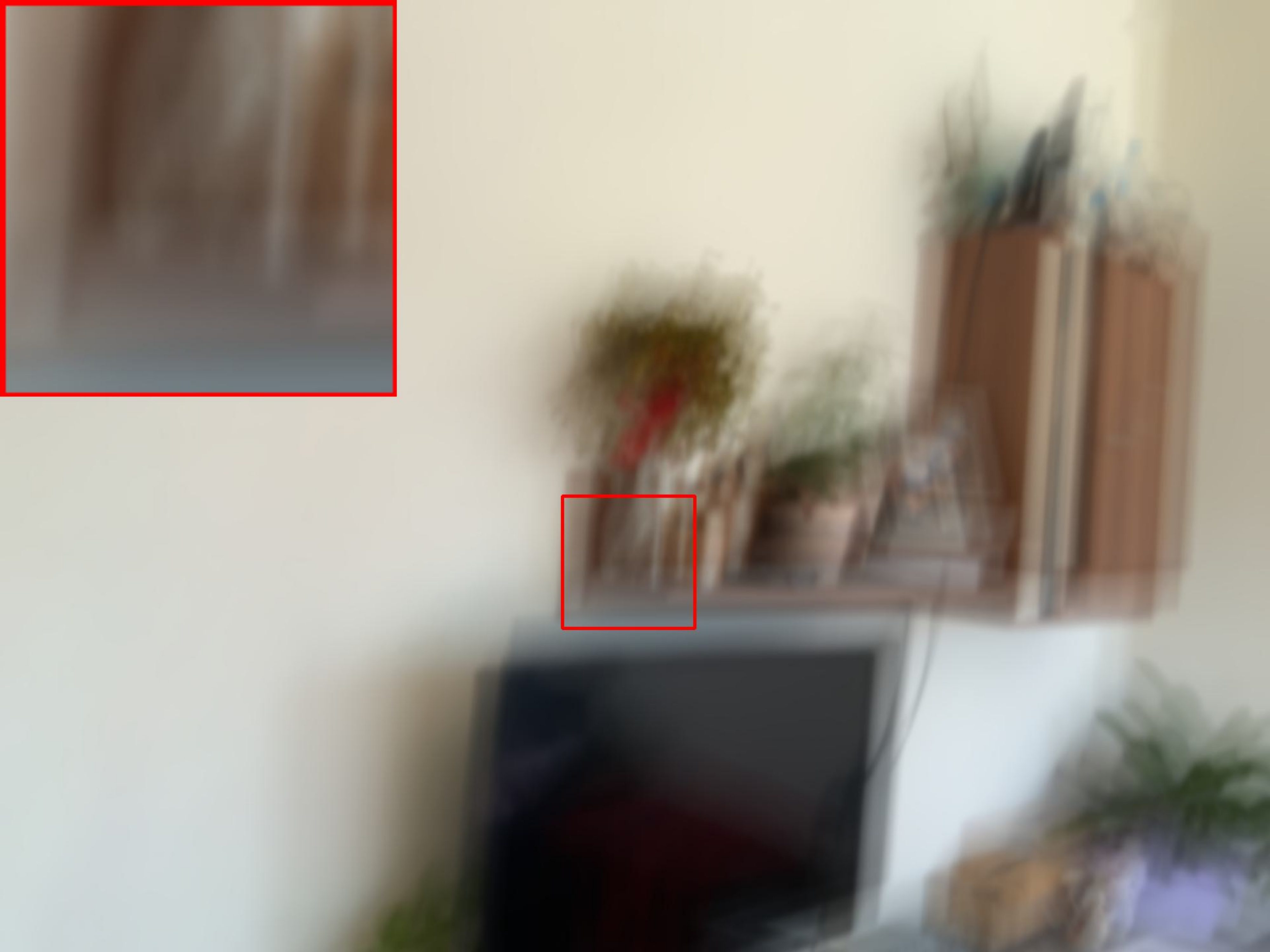}
   \includegraphics[width=0.18\linewidth]{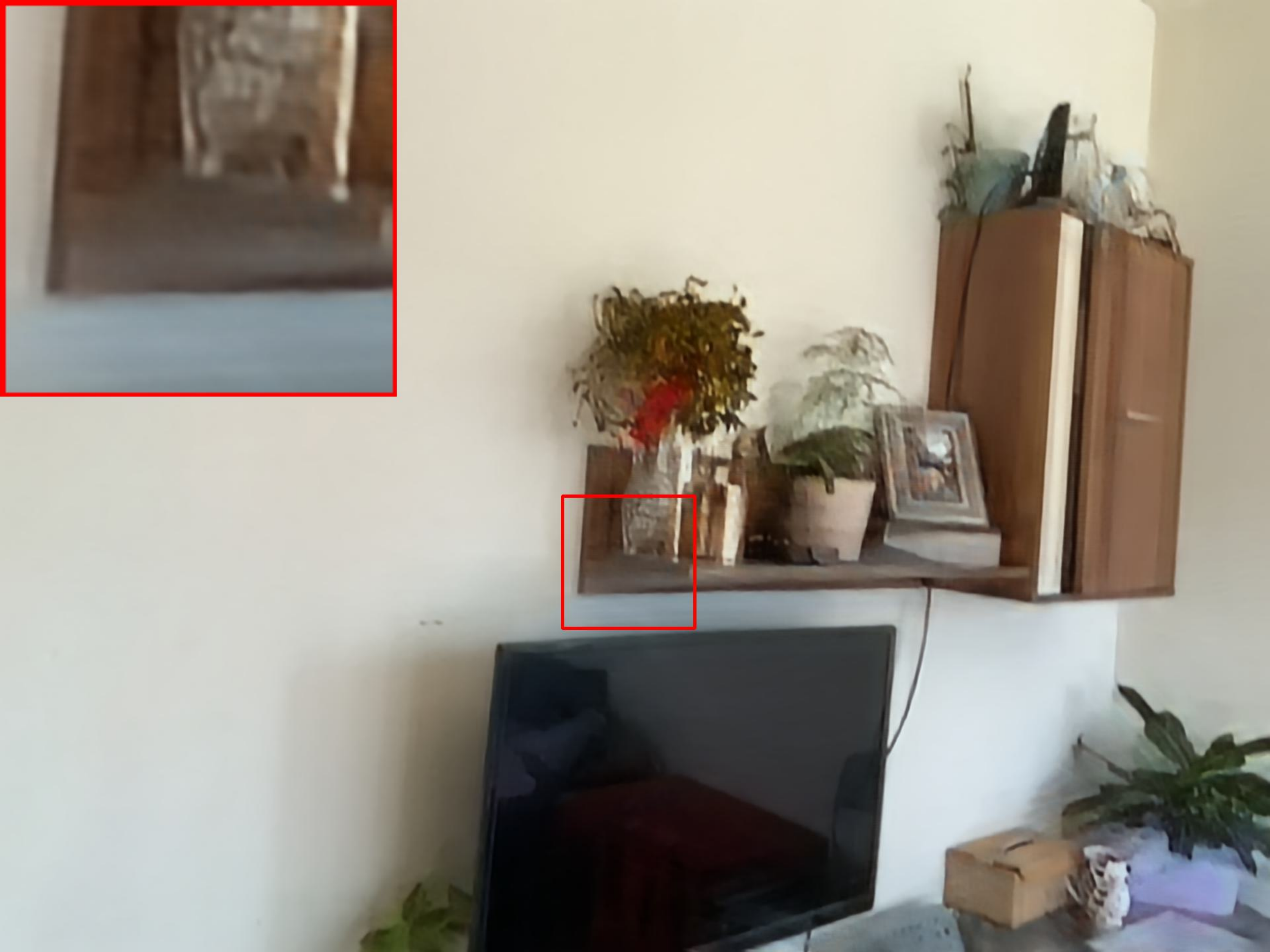}
   \includegraphics[width=0.18\linewidth]{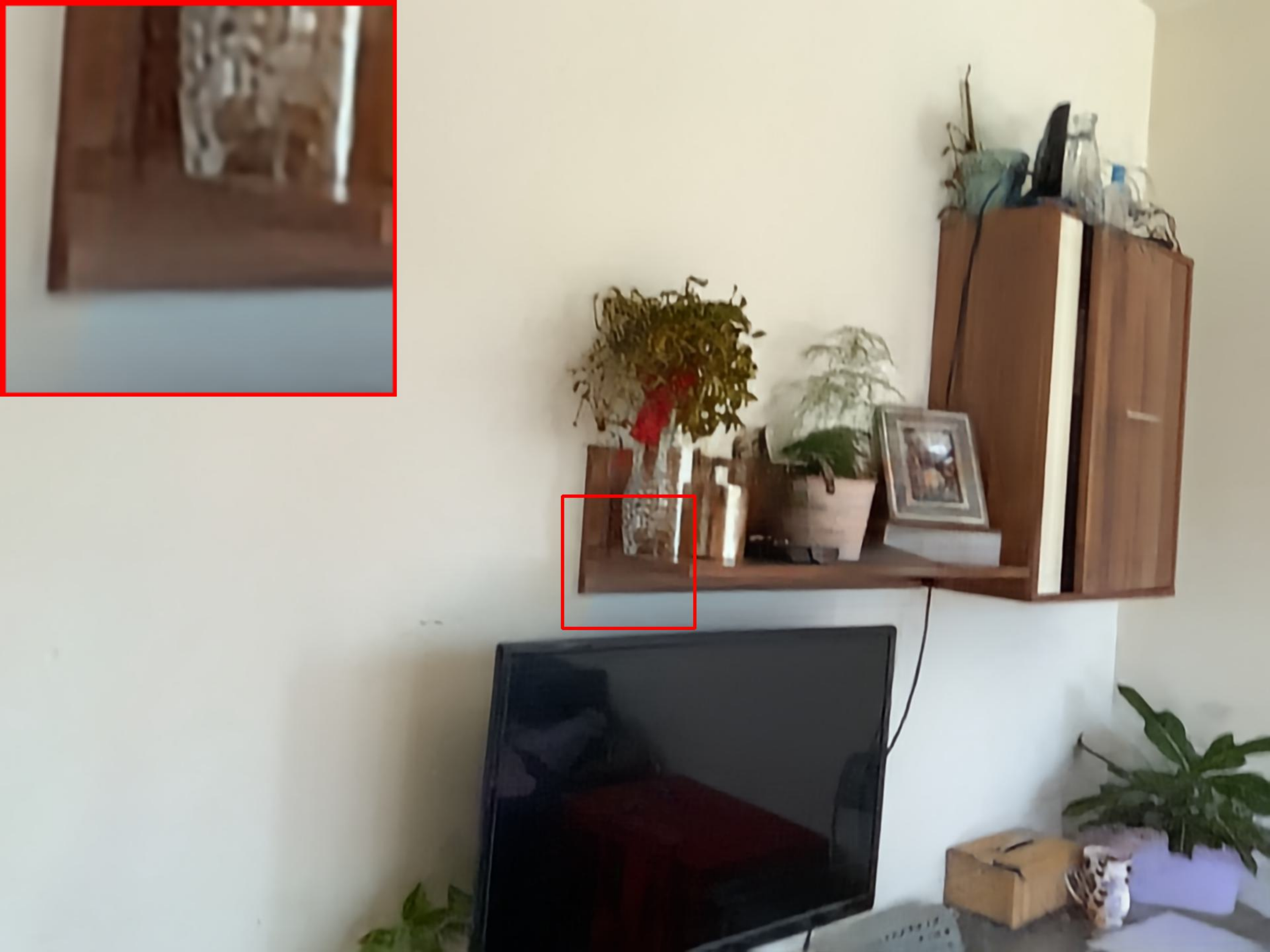}
   \includegraphics[width=0.18\linewidth]{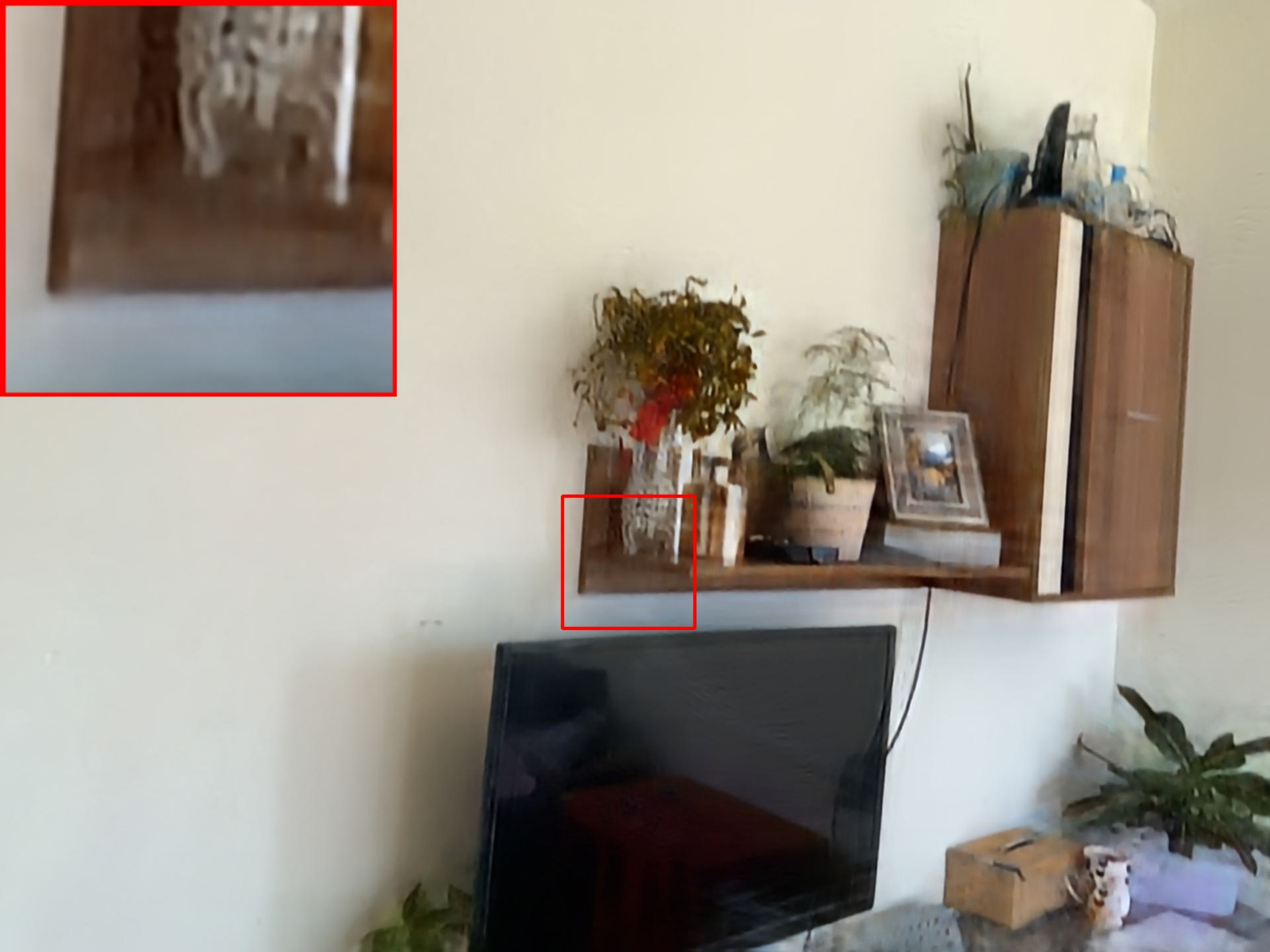}
   \includegraphics[width=0.18\linewidth]{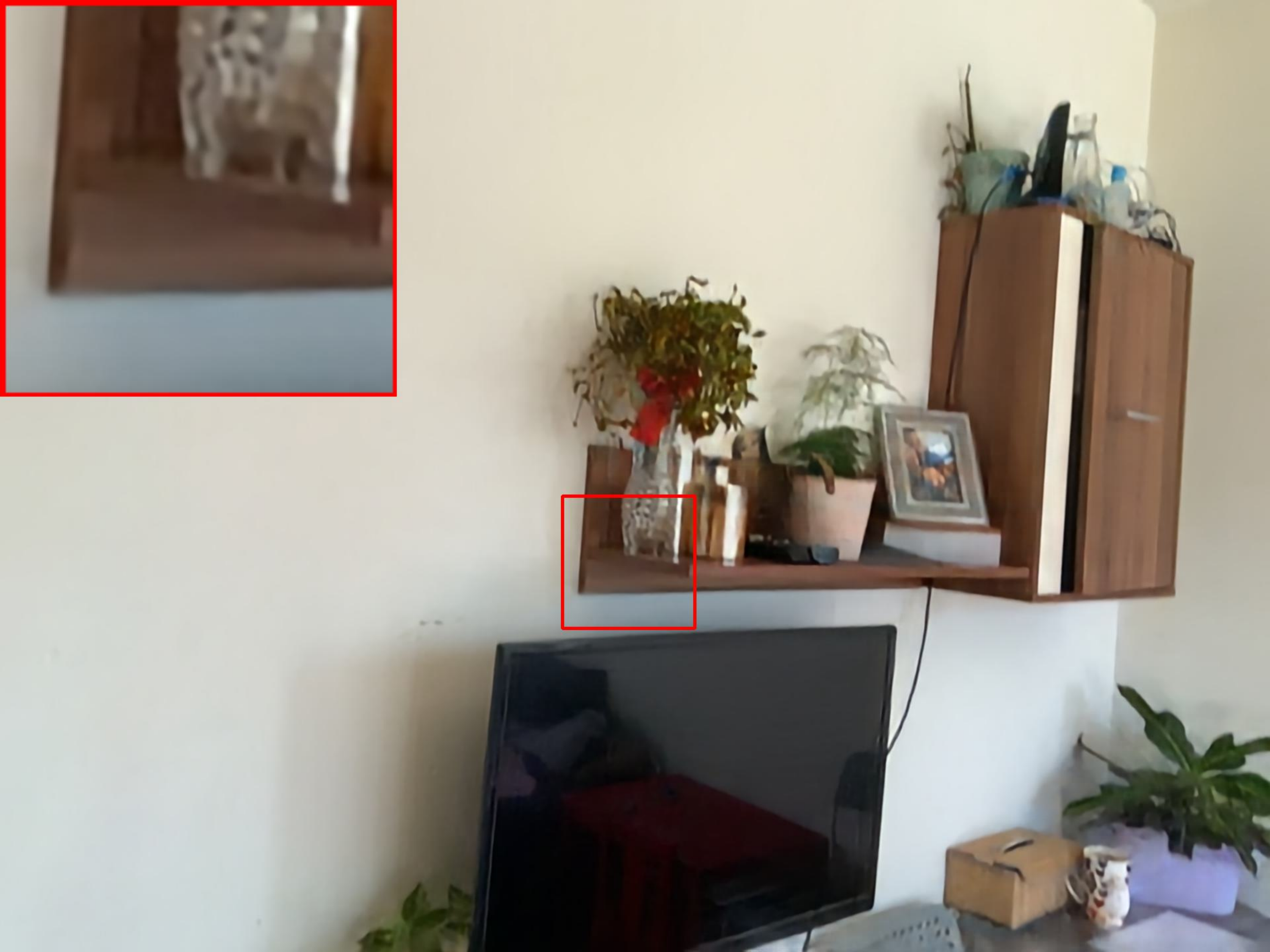} \\[-3pt]
    \subfigure[Blurred/GT]{\includegraphics[width=0.18\linewidth]{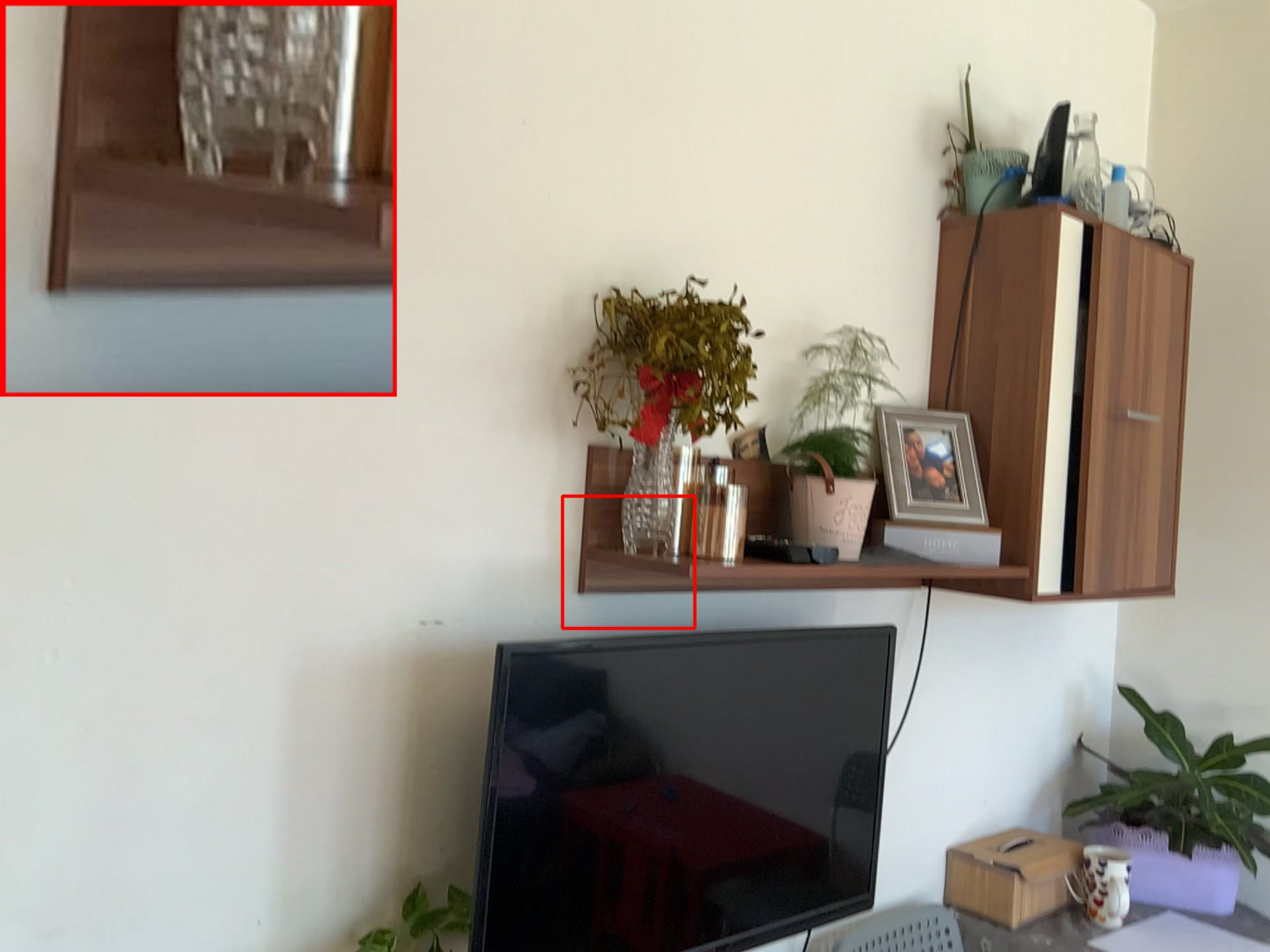}}
    \subfigure[(Depth-)Restormer]{\includegraphics[width=0.18\linewidth]{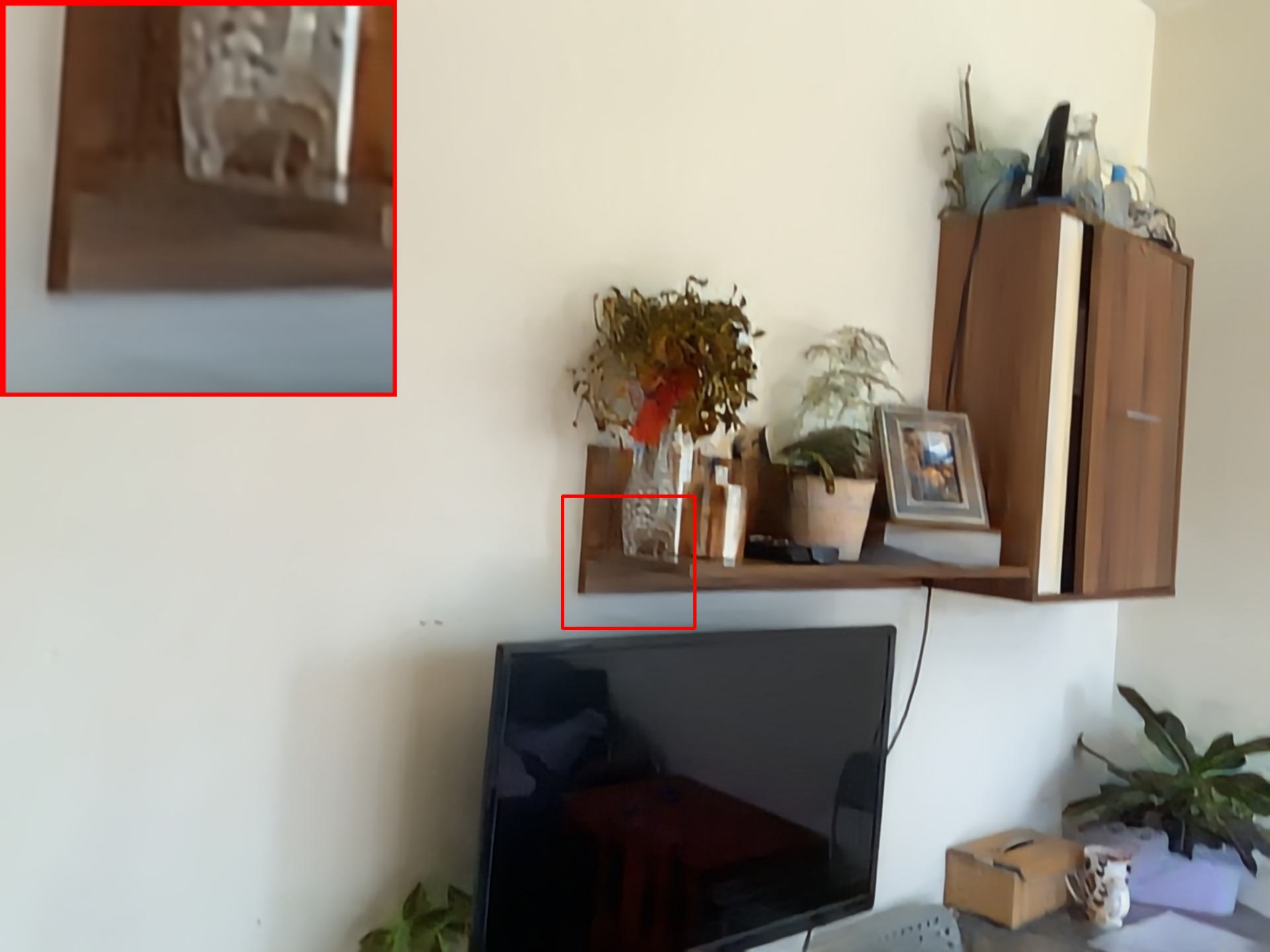}}
    \subfigure[(Depth-)DeblurDiNATL]{\includegraphics[width=0.18\linewidth]{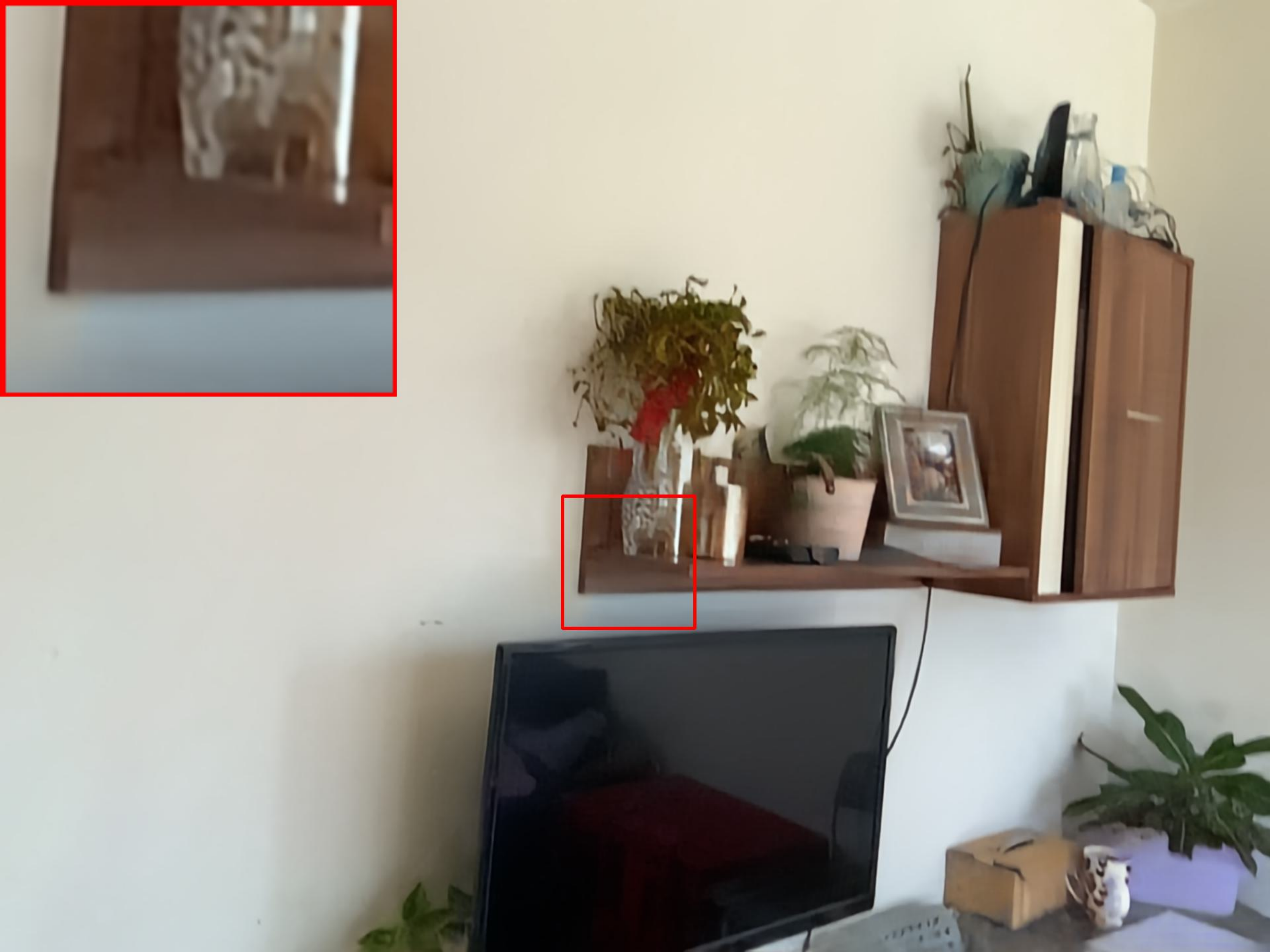}}
    \subfigure[(Depth-)Stripformer]{\includegraphics[width=0.18\linewidth]{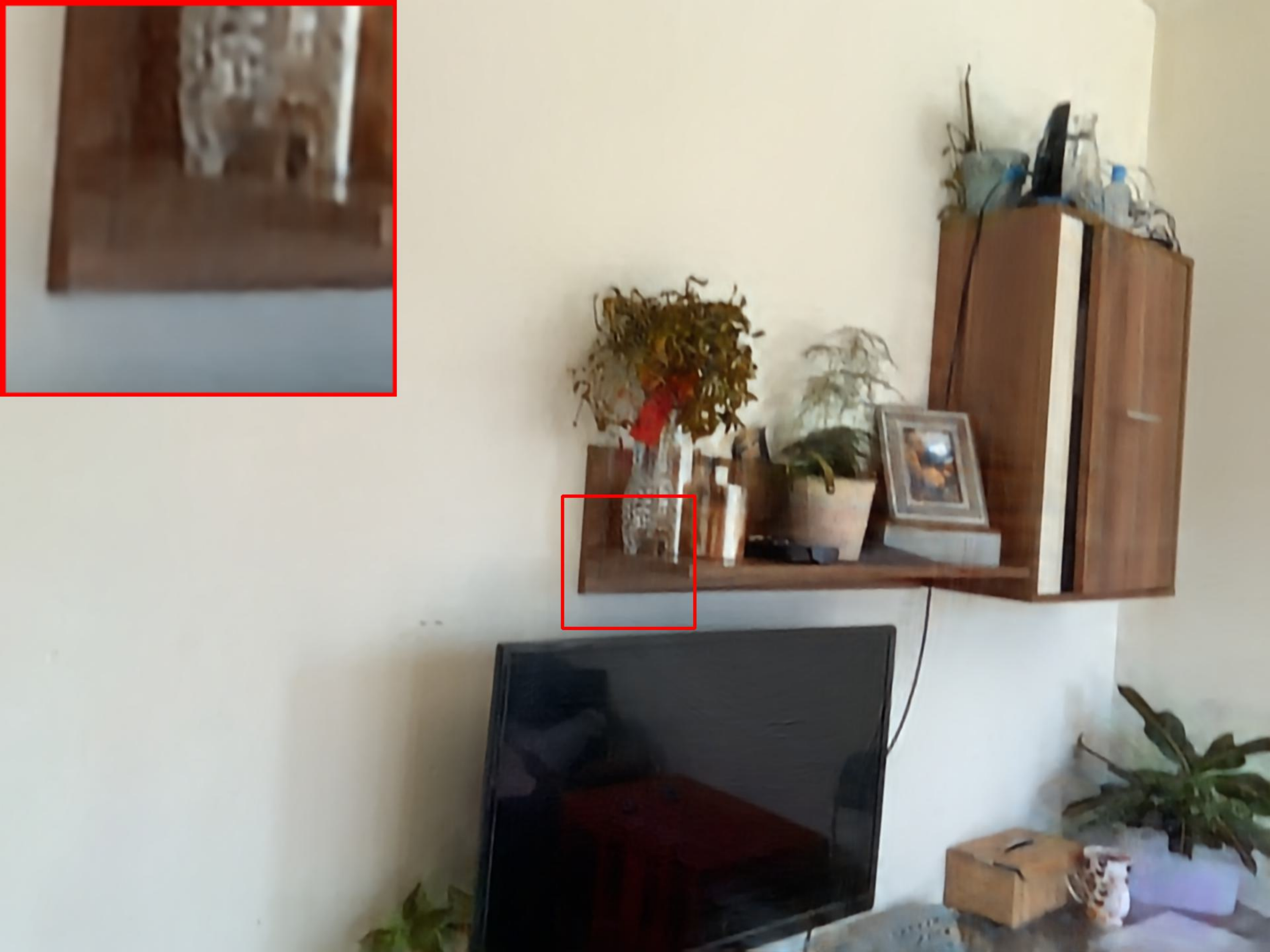}}
    \subfigure[(Depth-)NAFNet]{\includegraphics[width=0.18\linewidth]{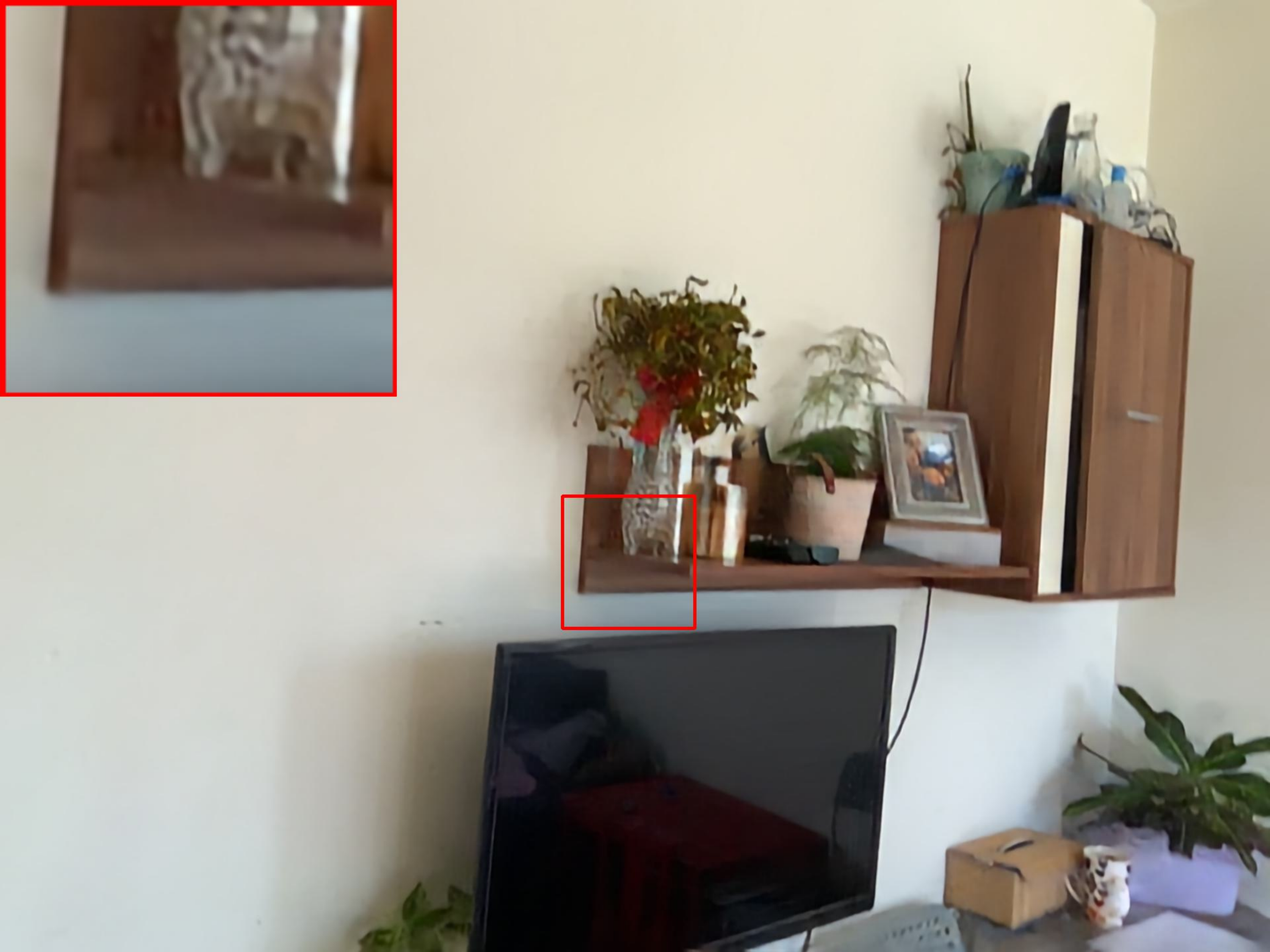}}   
  \caption{Visual comparison of deblurring results for the considered state-of-the-art models. For each scene, the top row presents results of the original model, while the bottom row presents results of the depth-enhanced model.}
  \label{fig:allresults}
\end{figure*}

We first assess whether mobile Lidar data can improve deblurring results on the selected state-of-the-art architectures. Results are shown in Table \ref{table:main}. We can notice that the use of Lidar data generally provides a significant improvement in deblurring performance. The only exception is the NAFNet architecture where we still observe improvement but more modest. This could be explained by both the unusual network design of NAFNet and reaching a saturation point in the ability to deblur by any model (indeed, NAFNet achieves baseline quality significantly better than the other models). We can also observe that the increase in number of parameters is modest with respect to the size of the original models. Qualitative results are reported in Fig. \ref{fig:allresults}. In the figure, for each scene, the top line is the result of the four state-of-the-art conventional deblurring models without depth information, while the bottom line is the results of the depth enhanced models. It can be noticed that in correspondence of object boundaries, the depth-enhanced models significantly reduce ghosting effects. Fig. \ref{fig:result_with_depth} reports the same result for the Restormer architecture while showing the depth information.

Overall, these results demostrate that mobile Lidar depth maps, despite their relatively low resolution, can successfully regularize the deblurring process, when properly used with the proposed scheme.

\subsection{Ablation Study}
\label{Ablation Studies}
In the ablation study, we carefully analyze our design decisions to validate their effectiveness. This study concerns the three main points of contribution of this paper, namely the quality of the depth maps, the scheme to fuse them with the deblurring model, and the continual learning strategy. All the ablation results use the Restormer architecture as baseline.

\subsubsection{Impact of Real Lidar Depth Maps}

\begin{figure*}[t]
    \centering
    \subfigure[GT Image]{\includegraphics[width=0.19\linewidth]{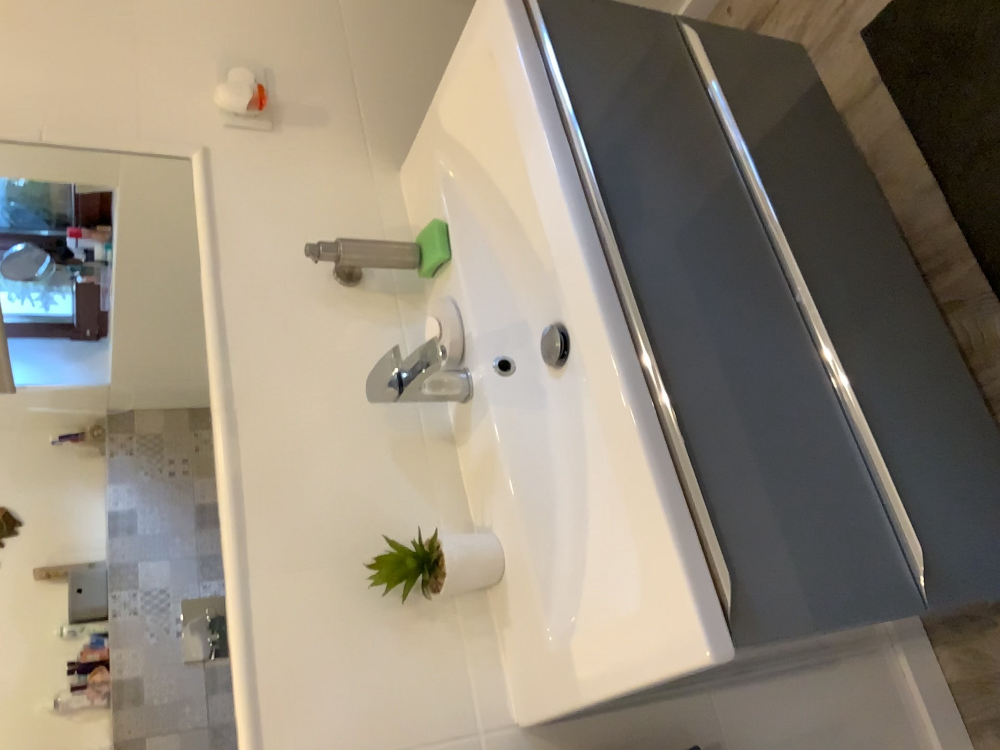}}
    \subfigure[Blurred Image]{\includegraphics[width=0.19\linewidth]{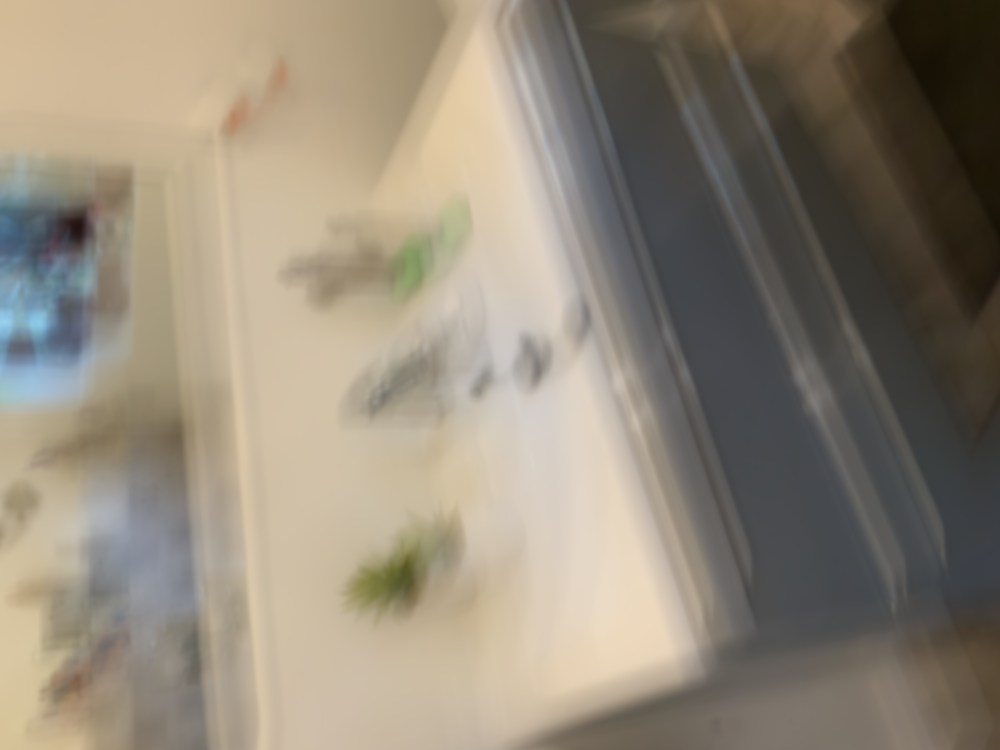}} 
    \subfigure[HR Depth]{\includegraphics[width=0.19\linewidth]{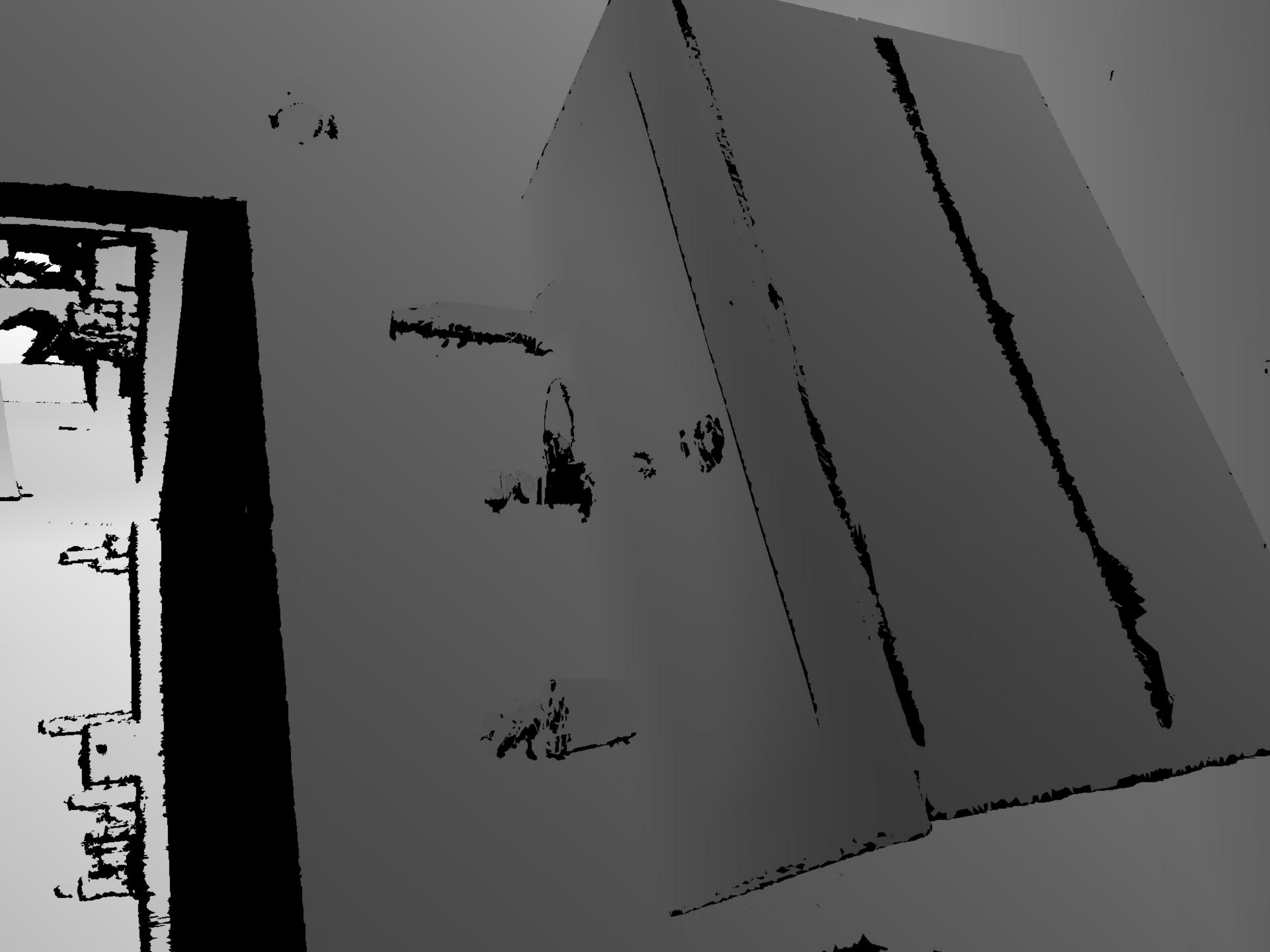}}
    \subfigure[SR Depth]{\includegraphics[width=0.19\linewidth]{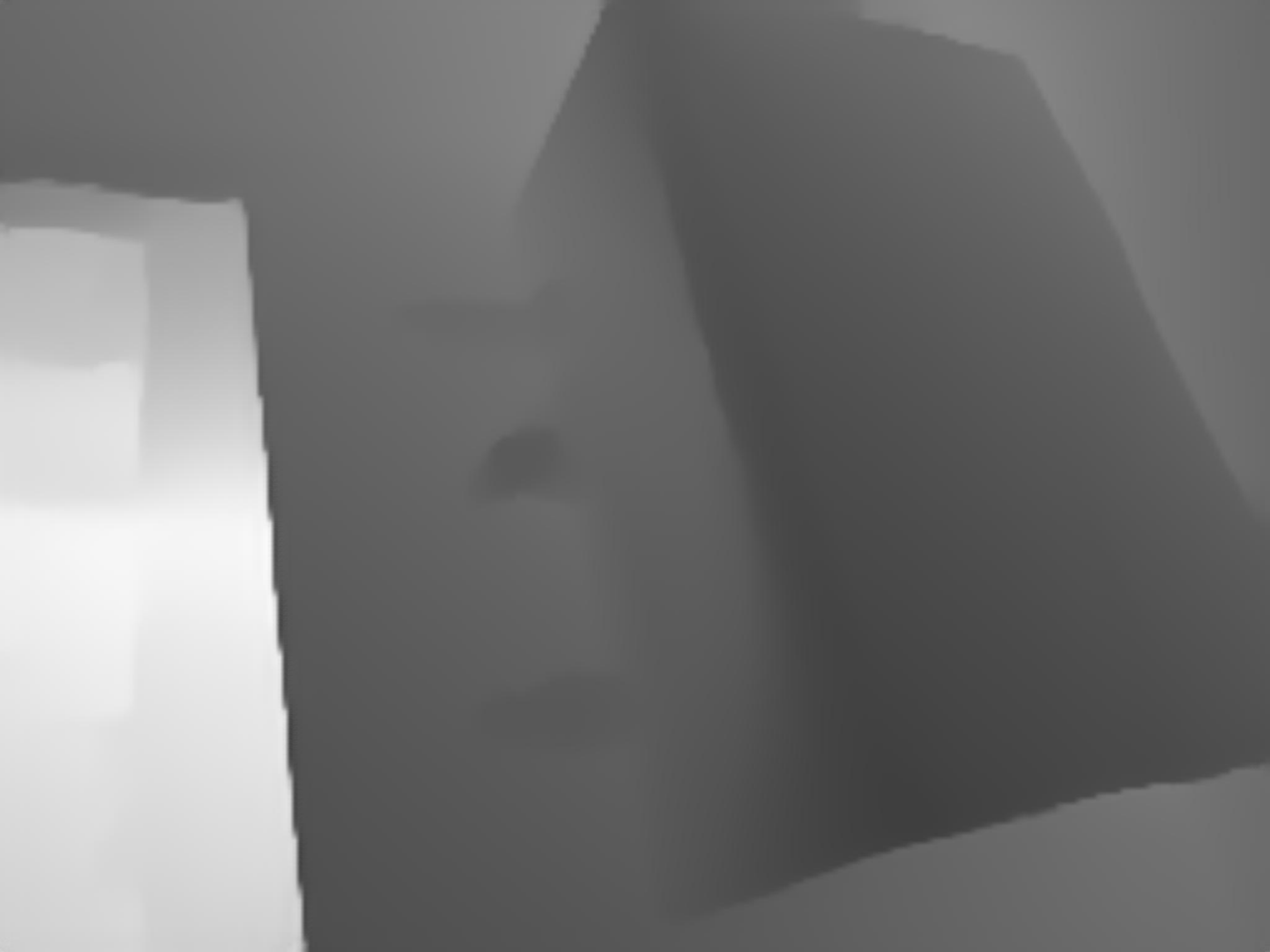}}
    \subfigure[Estimated Depth]{\includegraphics[width=0.19\linewidth]{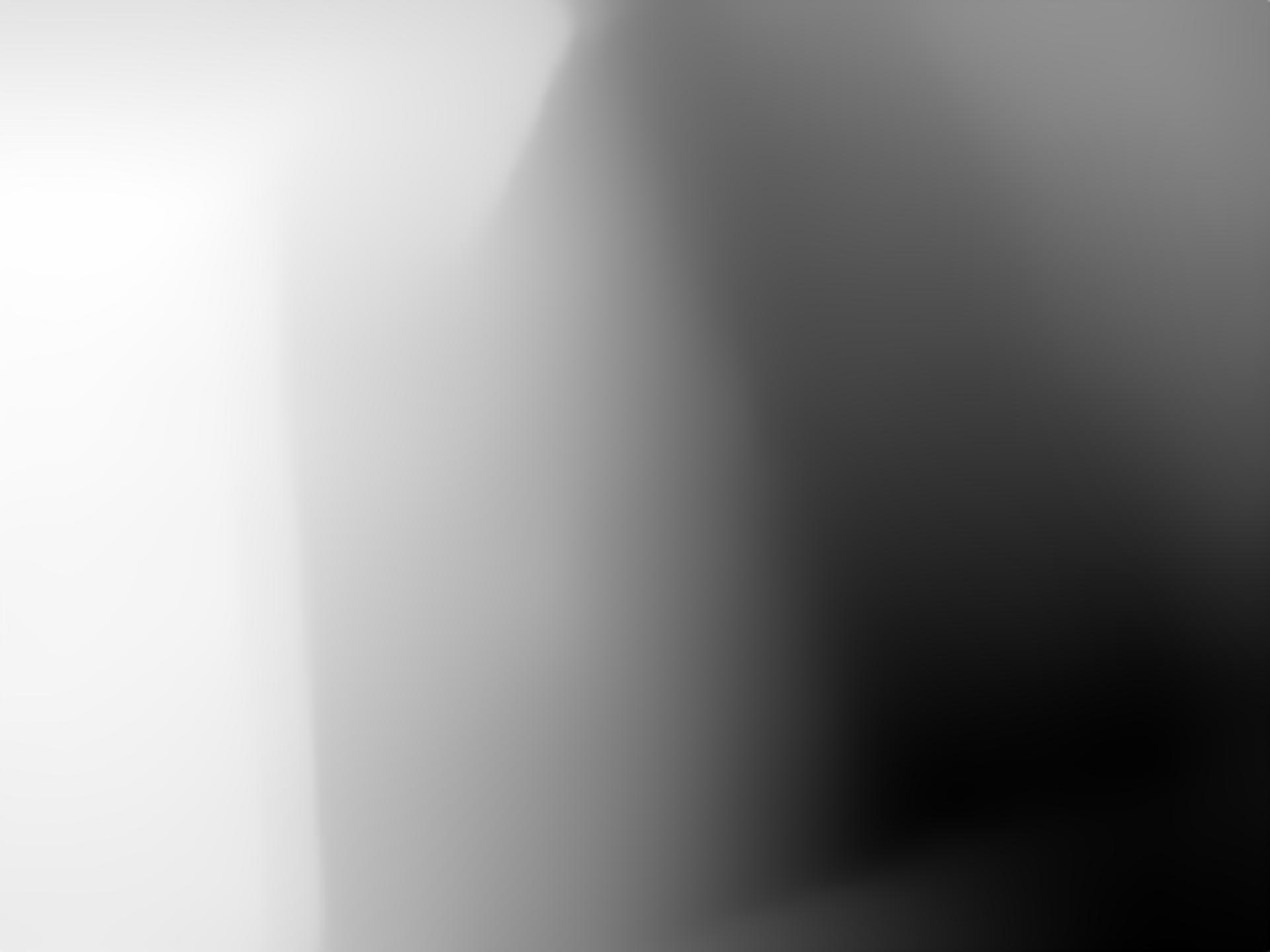}}     
    \caption{Comparison of depth maps. From left to right: ground truth image, blurred image, high-resolution depth map from Faro Focus S70 Lidar, super-resolved depth map from iPad Pro, depth map estimated from blurred image by \cite{depthanything}.}
    \label{fig:depthcompare}
\end{figure*}

\begin{figure*}[h]
  \centering
        \subfigure[RGB]{\includegraphics[width=0.23\linewidth]{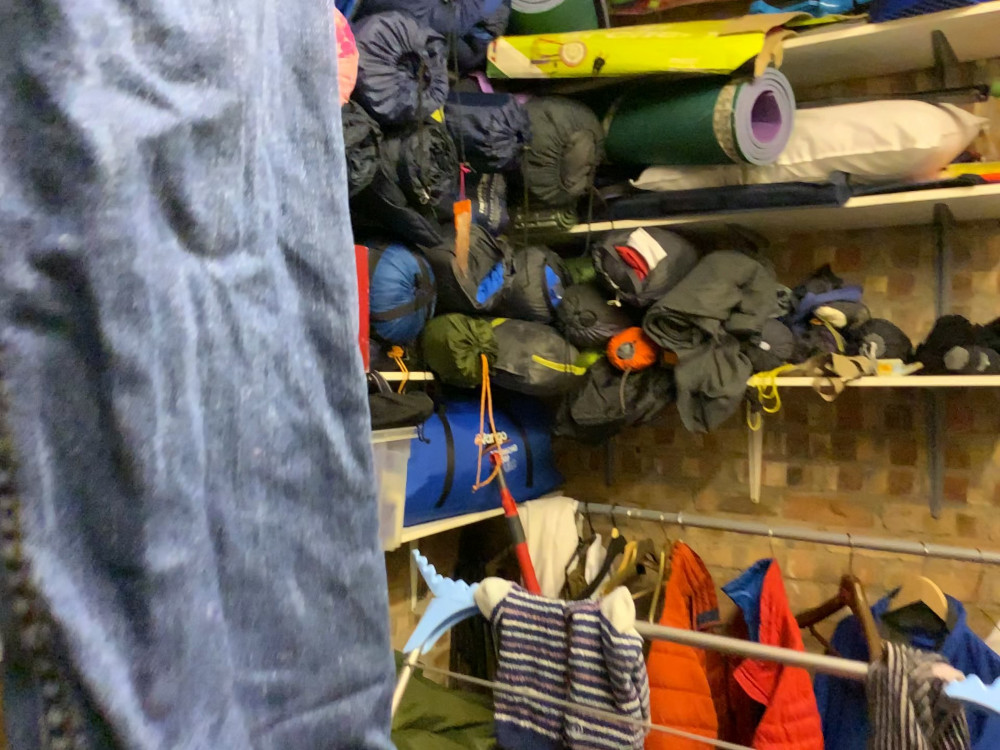}}
        \subfigure[$8\times$ SR Depth]{\includegraphics[width=0.23\linewidth]{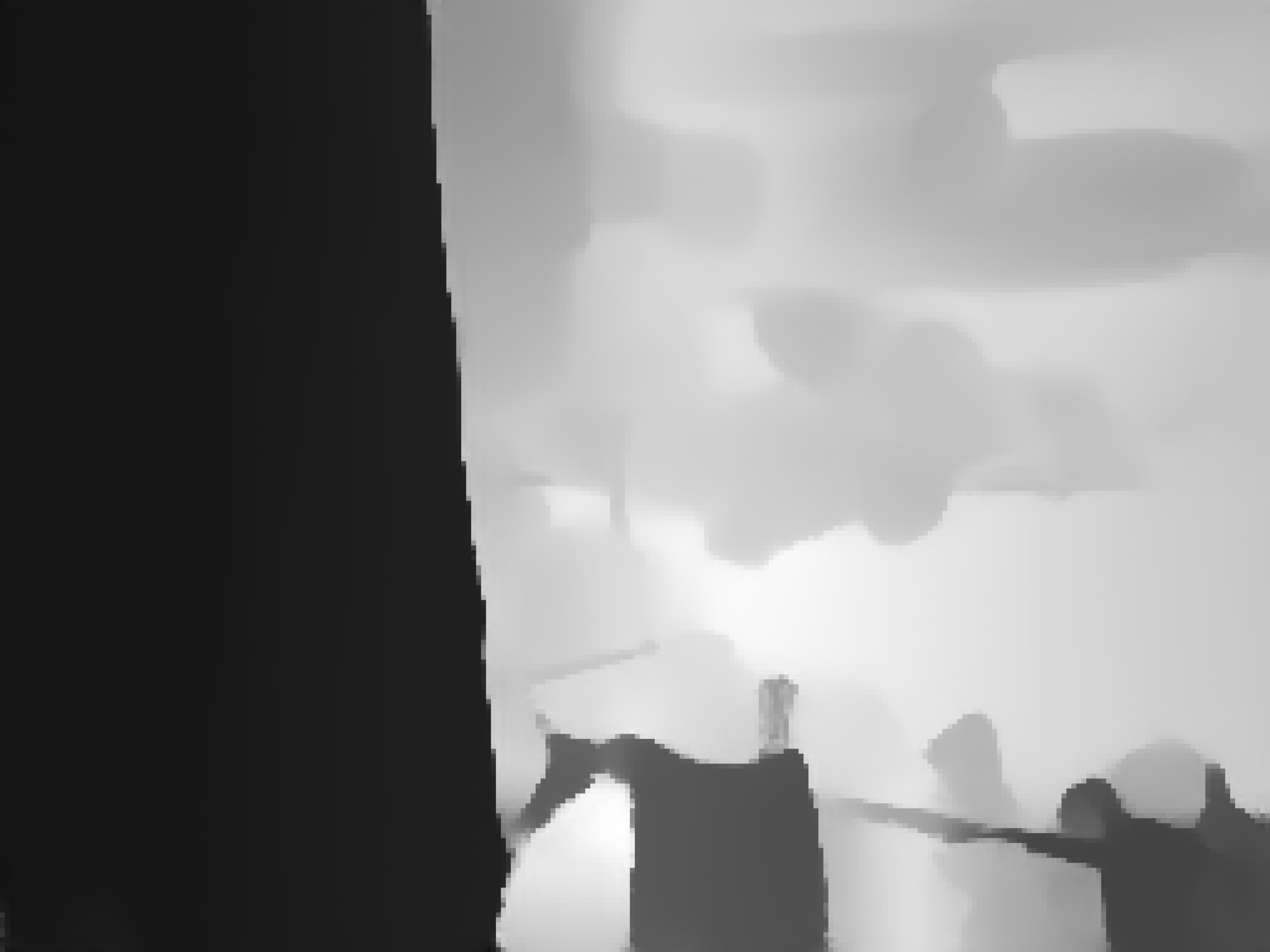}}
        \subfigure[Bicubic]{\includegraphics[width=0.23\linewidth]{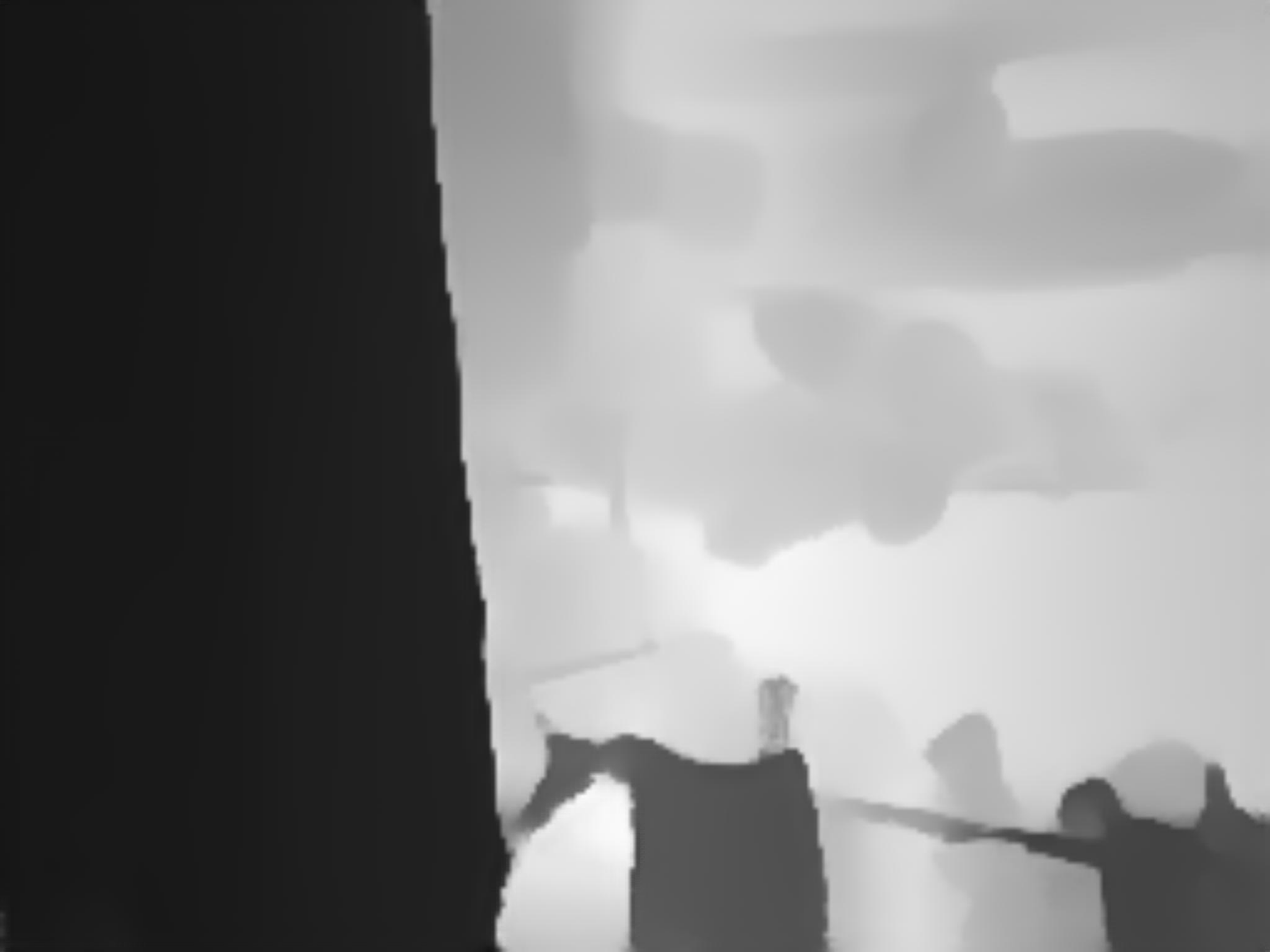}}
        \subfigure[difference]{\includegraphics[width=0.23\linewidth]{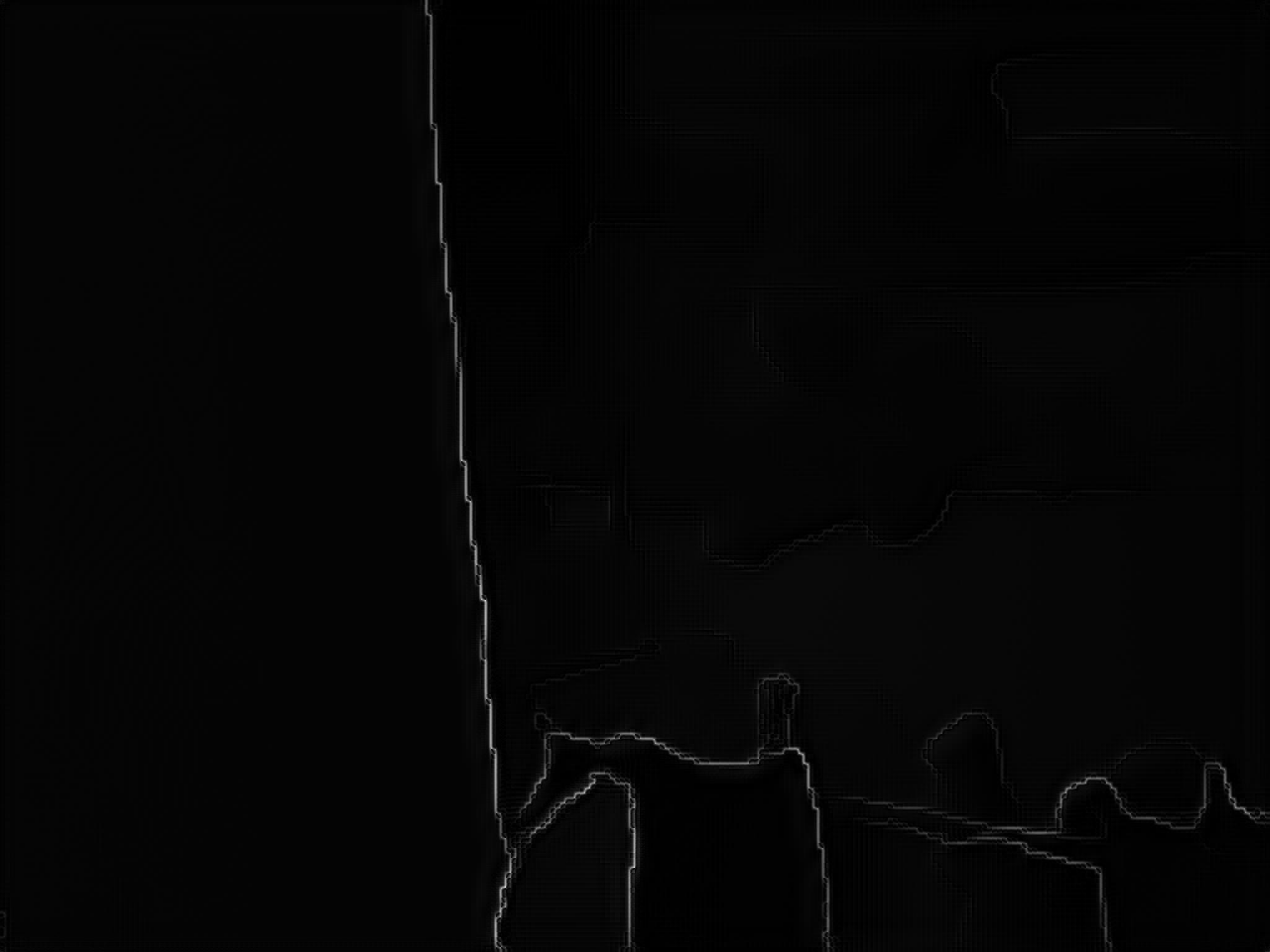}} \\
        \subfigure[RGB]{\includegraphics[width=0.23\linewidth]{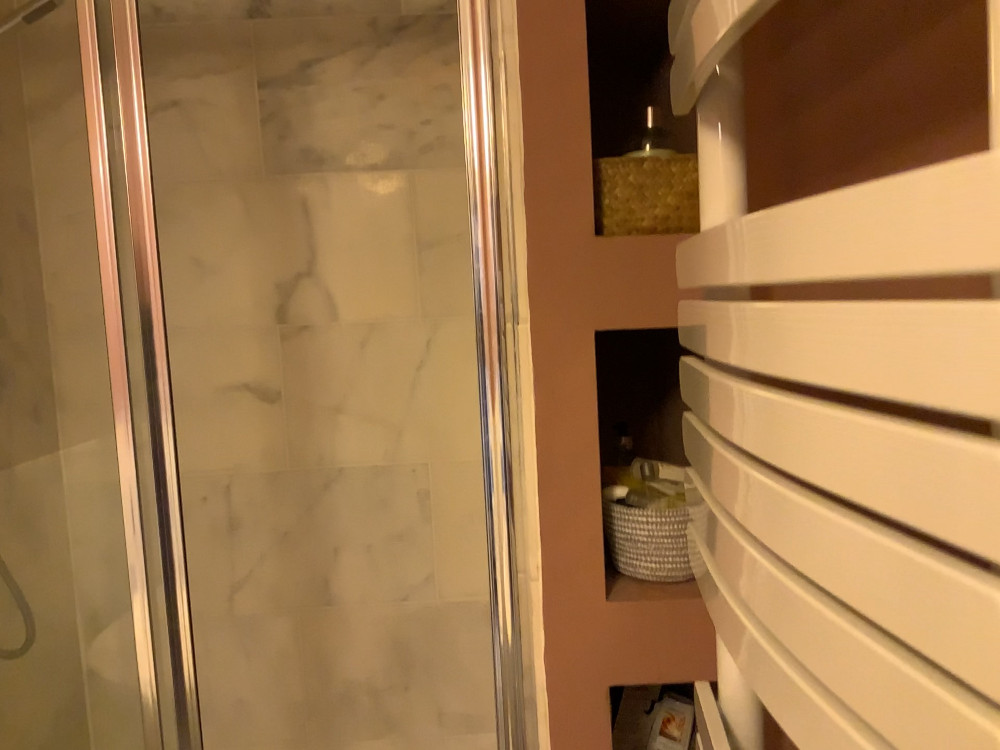}}
        \subfigure[SR Depth]{\includegraphics[width=0.23\linewidth]{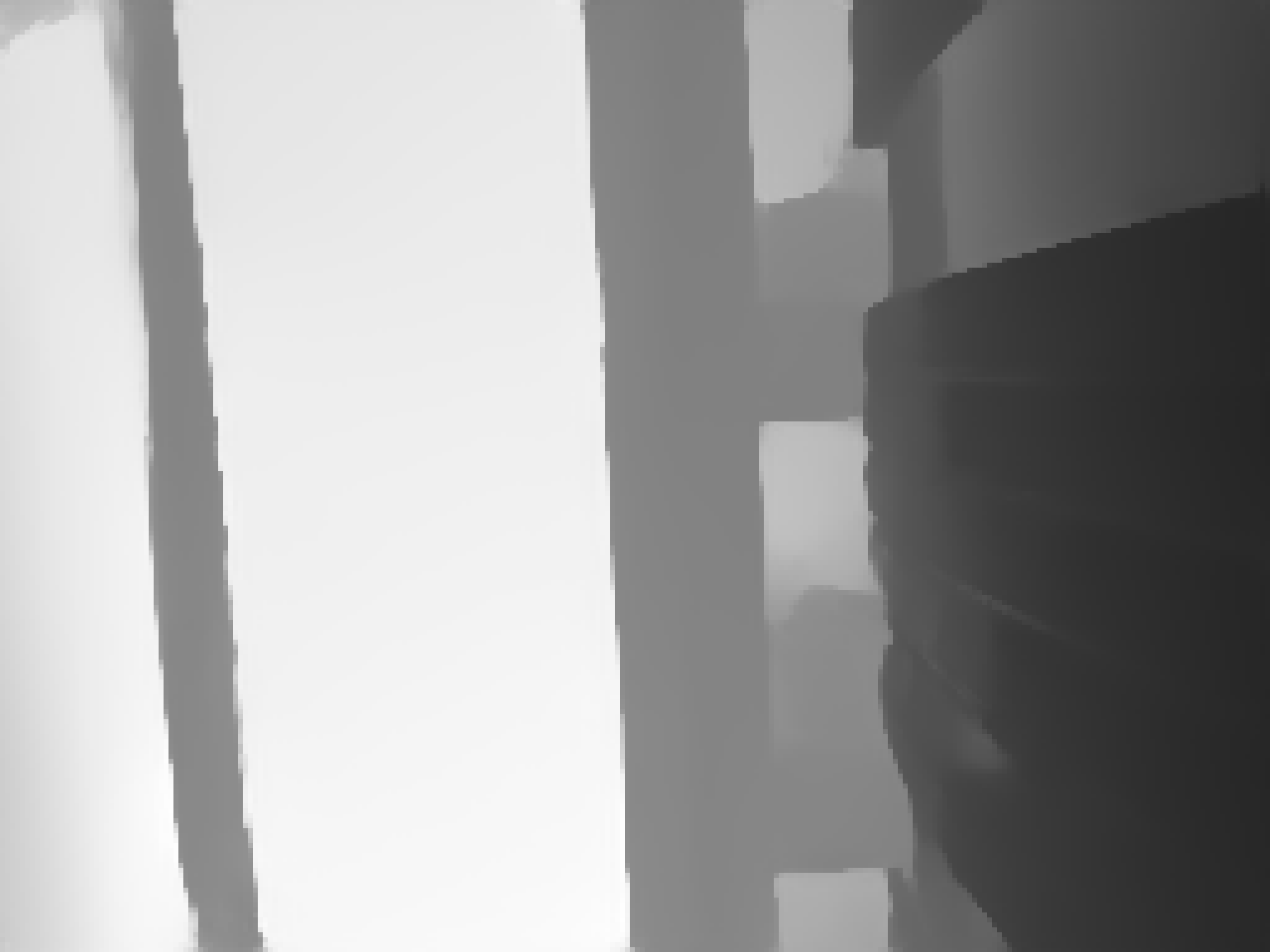}}
        \subfigure[Bicubic Depth]{\includegraphics[width=0.23\linewidth]{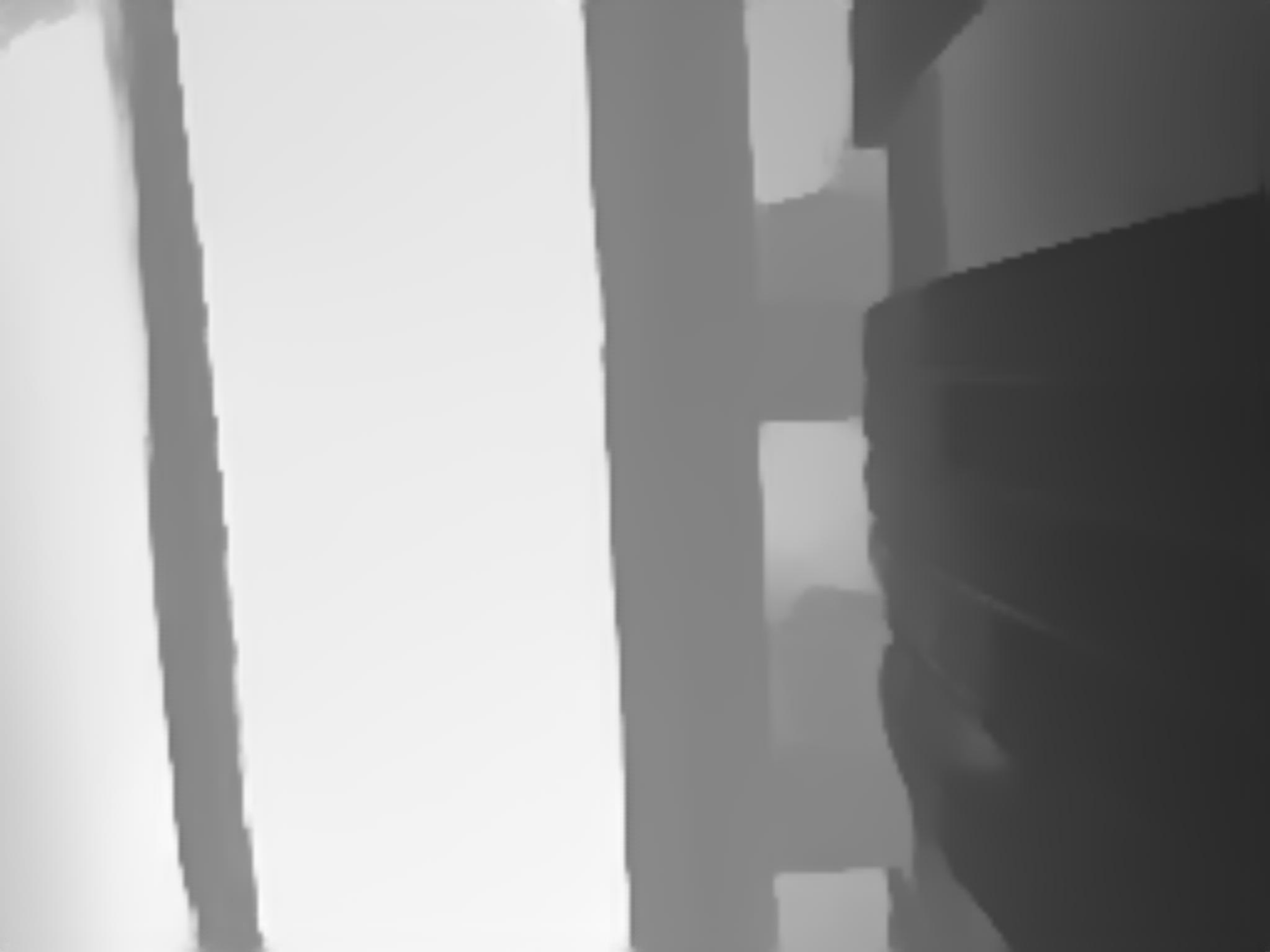}}
        \subfigure[difference]{\includegraphics[width=0.23\linewidth]{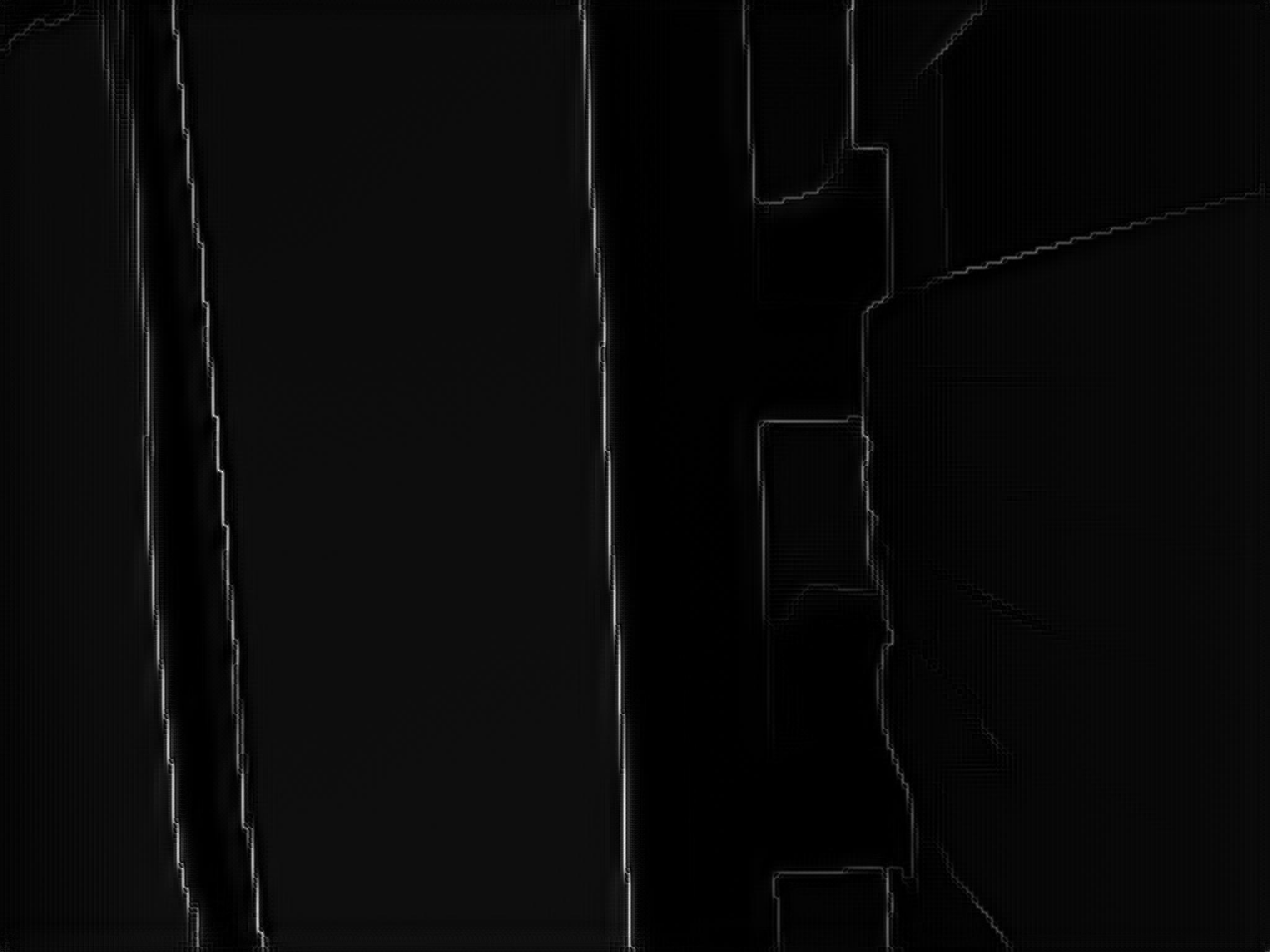}}
  \caption{The neural-network-super-resolved iPad depth map better preserves sharp edges after the upscaling operation, leading to improved deblurring performance.}
  \label{fig:depth_sr}
\end{figure*}

\begin{table}[t]
\caption{Impact of Lidar depth maps vs. depth estimation from blurry image.}
\centering
    \begin{tabular}{cccc}
    \textbf{Depth Type}       &\textbf{PSNR}    & \textbf{$\Delta$PSNR} \\
    \hline\hline
    None & 34.52 dB &  - \\
    Estimated & 35.79 dB & +1.27 dB \\ 
    \textbf{Mobile Lidar}  & \textbf{36.62 dB}  & \textbf{+2.10 dB} \\
    \hline
    \end{tabular}
\label{table:realfake}
\end{table}

Some approaches in the literature \cite{6215220}, \cite{9043904} attempt to use estimated depth maps to aid image deblurring, while \cite{pan2019single} utilized both real and estimated depth maps, achieving good performance. However, literature generally lacks comparisons between the use of real and estimated depth maps, and mobile Lidars have not yet been considered for image restoration problems. We argue that depth map estimation from a blurry image can only provide additional image features that might have some use in the reconstruction process but do not really provide additional side information, as an independent Lidar would. Therefore, in this study, we compared the PSNR of the deblurred image obtained when real Lidar depth maps are used and when, instead, a depth map is estimated from the blurry image. The state-of-the-art Depth Anything \cite{depthanything} model is used to estimate the depth maps. Because the model generates depth maps with the same resolution as the blurred images, the depth map super-resolution block is not used.
From Fig. \ref{fig:depthcompare}, we can see that the depth map generated from the blurred image lacks the explicit geometric information, particularly regarding object edges, that is present both in the high-resolution depth map from the high-end Lidar and the super-resolved depth map of the mobile Lidar. Quantitative results in Table \ref{table:realfake} confirm the results of previous literature \cite{6215220,9043904} in that even estimating the depth map from blurry images provides some degree of regularization to the deblurring process, leading to some improvements. However, Lidar depth maps provide a more significant improvement in performance, proving that the independent side information captured by the Lidar instrument, even if at modest resolution, can boost image quality.

\subsubsection{Impact of Lidar Depth Super-resolution}
\label{sec:exp_superres}

\begin{table}
\caption{Deblurring sensitivity to depth map resolution.}
\centering
\begin{tabular}{lccc}
                \textbf{Depth Type} & \textbf{PSNR}     & \textbf{$\Delta$PSNR} \\
    \hline\hline
    None & 34.52 dB & - \\
    Bicubic $\times 8$ & 35.78 dB & +1.26 dB \\ 
    NN $\times 4$ & 36.08 dB  & +1.56 dB \\
    \textbf{NN $\times 8$} & \textbf{36.62 dB}  & \textbf{+2.10 dB}  \\
    HR & 36.59 dB & +2.07 dB \\
    \hline
\end{tabular}
\label{table:depth_sr}
\end{table}

\begin{table}
\caption{Ablation of fusion design and continual learning strategy.}
\centering
\begin{tabular}{ccccc}
                \textbf{Depth}   & \textbf{Adapter} & \textbf{Encoder Freezing} & \textbf{PSNR}     & \textbf{$\Delta$PSNR} \\
    \hline\hline
                None & $\times$ & $\times$ & 34.52 dB & - \\
    \hline
                NN $\times 8$ & $\checkmark$ & $\times$ & 36.17 dB & +1.65 dB \\
                NN $\times 8$ & concat. & $\checkmark$  & 35.24 dB & +0.72 dB \\
    \hline
                NN $\times 8$ & $\checkmark$ & $\checkmark$  & \textbf{36.62 dB} & \textbf{+2.10 dB} \\
    \hline
\end{tabular}
\label{table:fusion_abl}
\end{table}

While we observed that a super-resolved mobile Lidar depth map can increase deblurring quality more than estimating it from the blurred image, we still need to analyze the sensitivity of the process to the resolution of the depth map. Therefore, we conducted an experiment with different depth map super-resolution scales, specifically 4 times and 8 times, different methods, namely bicubic interpolation and also compared the results with high-resolution depth maps provided by the Faro Focus S70 Lidar. The results are shown in Table \ref{table:depth_sr}. We first notice that the neural network approach to depth super-resolution significantly outperforms bicubic interpolation, as expected. We also notice that the $8\times$ super-resolution factor, which matches the ratio between the RGB images and the iPad depth maps provides the best results. Interestingly, the depth maps processed with $8\times$ super-resolution network are capable of achieving equivalent deblurring performance to the high-resolution depth maps acquired with the Faro Focus S70 Lidar. A visualization of the $8\times$ super-resolved depth maps against bicubic interpolation is shown in Fig. \ref{fig:depth_sr}.

\subsubsection{Impact of depth fusion and continual learning strategy}

In this study, we evaluated the design of the depth map fusion method and the continual learning strategy to create a joint deblurring model. Results are reported in Table \ref{table:fusion_abl}. First, we can notice that the proposed approach of freezing the encoder part of the architecture is indeed more effective than finetuning the entire model, which would result in a lower PSNR by 0.45 dB. The use of adapters in also validated against depth map concatenation to the input of the model. As explained in Sec. \ref{sec:quality_depth}, this is not as effective as the use of deep feature modulation, and, in fact, results in loss of PSNR by 1.33 dB.

\section{Conclusions and Limitations}
We proposed using Lidar depth maps to further enhance the performance of deep deblurring models. In particular, we showed that inexpensive mobile Lidar devices can provide useful side information that improves quality of deblurred images, especially thanks to information about object edges. We also presented a universal way of integrating Lidar depth in any state-of-the-art image deblurring model by means of lightweight neural network adapters, combined with depth super-resolution, and a continual learning strategy. Although the achieved results are very promising, and indeed show that the side information provided by mobile Lidars can significantly boost image quality, there are some fundamental limitations to this sensor fusion approach. In particular, a Lidar cannot provide any information about scene textures but only 3D geometry. In the worst case of a flat textured surface, Lidar would not provide any useful data. Additionally, the range of such instruments can be limited, so scenes with objects far away from the camera may not benefit from the Lidar sensor. Finally, some geometries might be difficult to capture due to light scattering away from the camera.

\bibliographystyle{IEEEtran}
\bibliography{biblio}

\begin{IEEEbiography}[{\includegraphics[width=1in,height=1.25in,clip,keepaspectratio]{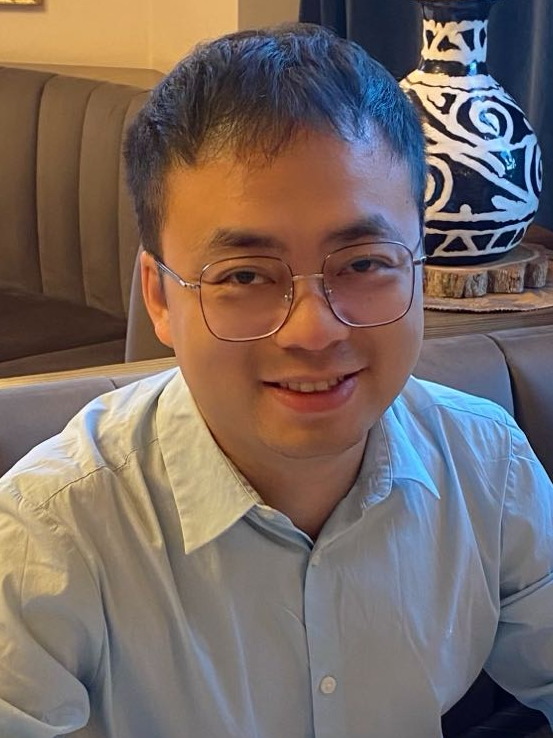}}]{Ziyao Yi} received the M.Sc. degree in Electrical Engineering from Washington University in St. Louis, USA, 2019. Prior to beginning the Ph.D., he worked for four years as a Computer Vision Algorithm Engineer in industry, where he developed and optimized computer vision and deep learning algorithms for various applications. Currently, he is pursuing the Ph.D. degree at Politecnico di Torino, Italy. His primary research interests include learning-based image/video restoration and the development of efficient deep learning models and algorithms.

\end{IEEEbiography}

\begin{IEEEbiography}[{\includegraphics[width=1in,height=1.25in,clip,keepaspectratio]{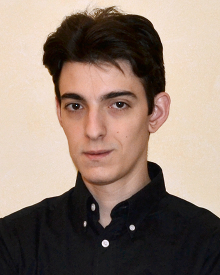}}]{Diego Valsesia} (S'13-M'17) received the Ph.D. degree in electronic and communication engineering from the Politecnico di Torino, in 2016. He is currently an Assistant Professor with the Department of Electronics and Telecommunications (DET), Politecnico di Torino. His main research interests include processing of remote sensing images, and deep learning for inverse problems in imaging. He is a Senior Area Editor for the IEEE Transactions on Image Processing, for which he received the 2023 Outstanding Editorial Board Member Award. He is a member of the EURASIP Technical Area Committee for Signal and Data Analytics for Machine Learning and a member of the ELLIS society. He was the recipient of the IEEE ICIP 2019 Best Paper Award, the IEEE Multimedia 2019 Best Paper Award. \end{IEEEbiography}

\begin{IEEEbiography}[{\includegraphics[width=1in,height=1.25in,clip,keepaspectratio]{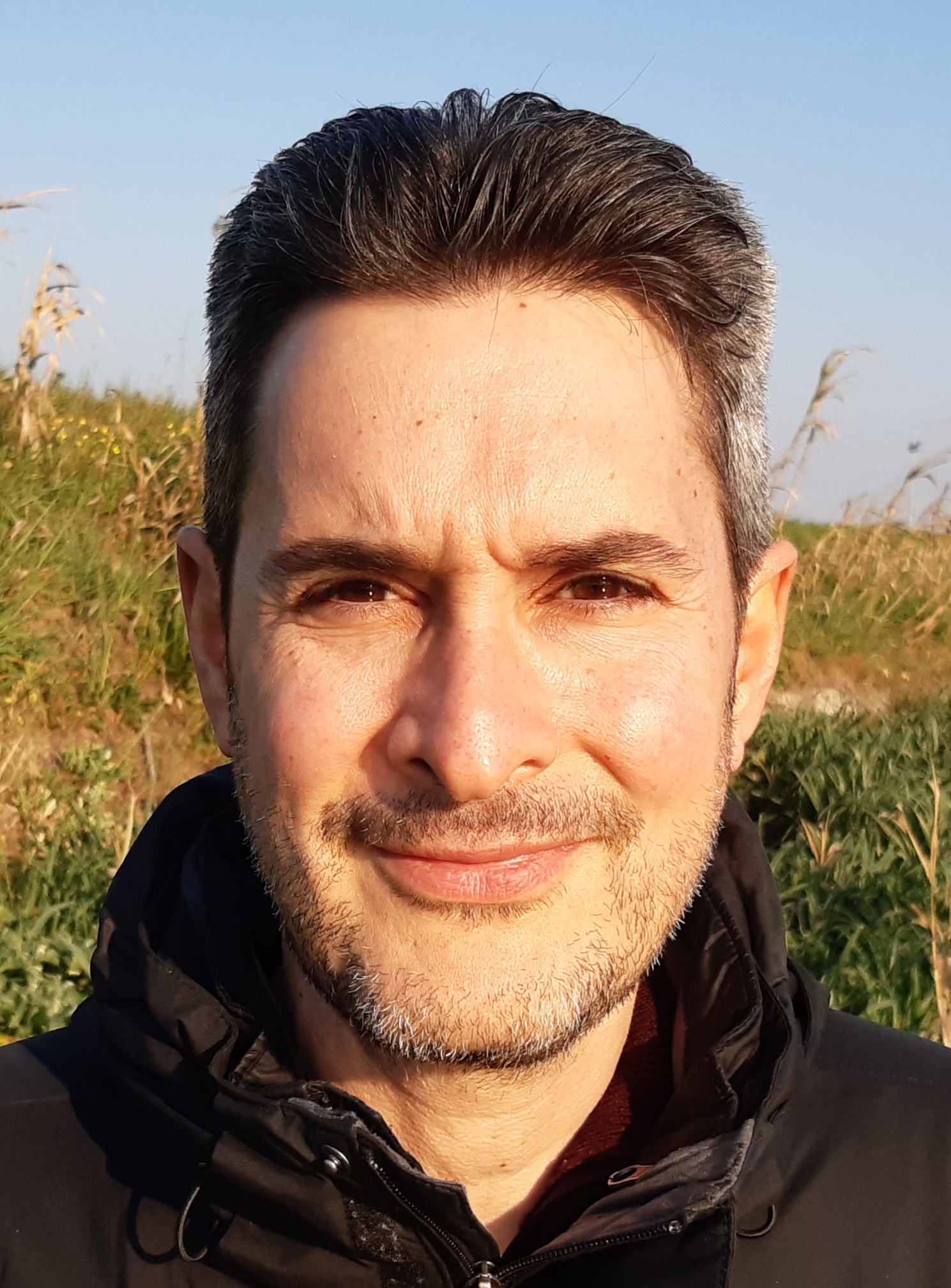}}]{Tiziano Bianchi} received the M.Sc. degree (Laurea) in electronic engineering and the Ph.D. degree in information and telecommunication engineering from the University of Florence, Italy, in 2001 and 2005, respectively. From 2005 to 2012, he was a Research Assistant with the Department of Electronics and Telecommunications, University of Florence. He joined Politecnico di Torino in 2012 as an Assistant Professor. He is currently an Associate Professor with the Politecnico di Torino. He has authored over 100 papers in international journals and conference proceedings. His research interests include multimedia security technologies, multimedia forensics, processing of remote sensing images. He is currently Associate Editor for the IEEE Transactions on Information Forensics and Security and Senior Area Editor for the Journal of Visual Communication and Image Representation. He was the recipient of the IEEE Multimedia 2019 Best Paper Award and the 2021 and 2022 Best Associate Editor Award of the Journal of Visual Communication and Image Representation.
\end{IEEEbiography}

\begin{IEEEbiography}[{\includegraphics[width=1in,height=1.25in,clip,keepaspectratio]{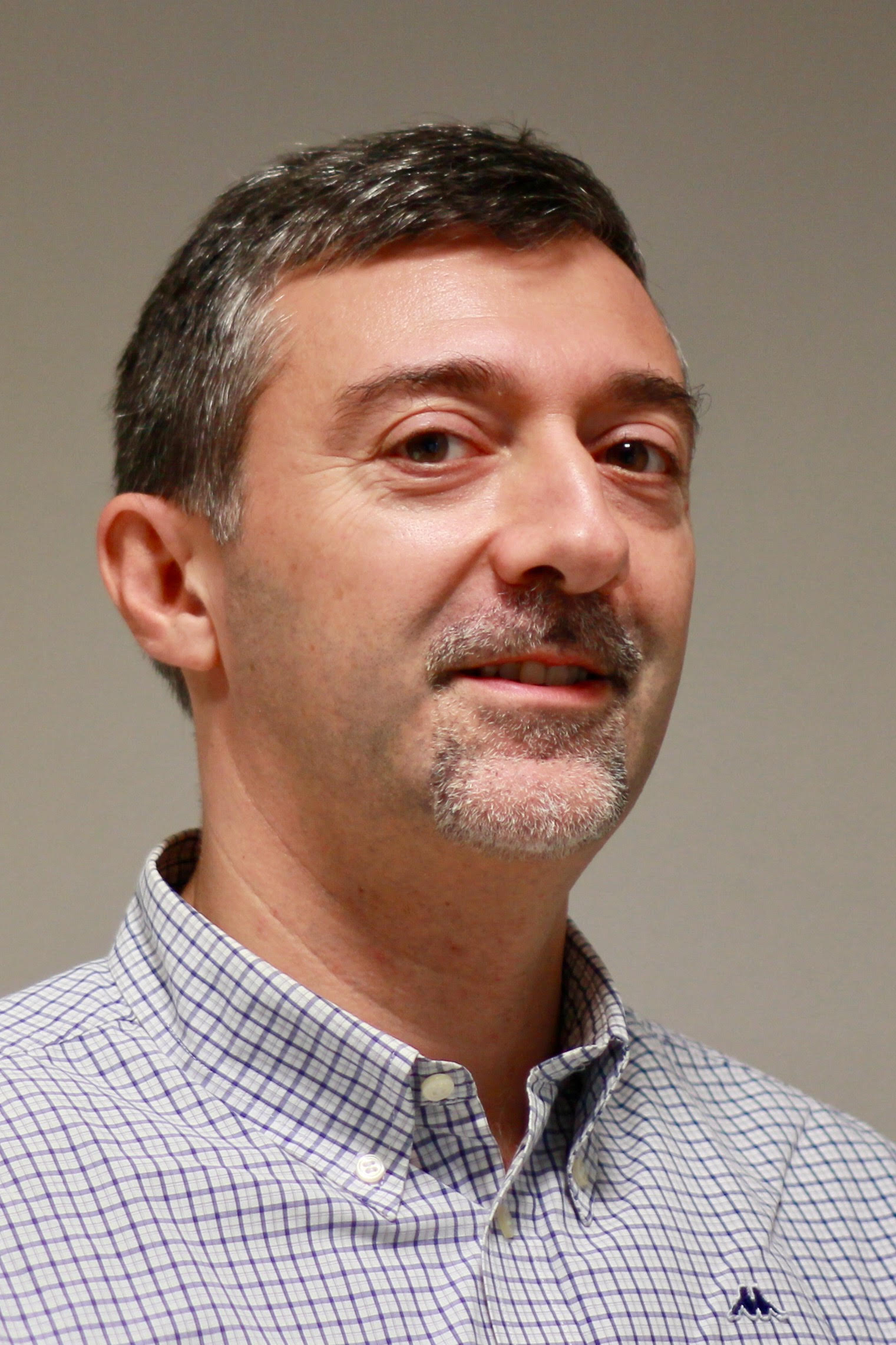}}]{Enrico Magli} (S'97-M'01-SM'07-F'17) received the M.Sc. and Ph.D. degrees from the Politecnico di Torino, Italy, in 1997 and 2001, respectively. He is currently a Full Professor with Politecnico di Torino, Italy, where he leads the Image Processing and Learning group, performing research in the fields of deep learning for image and video processing, image compression and image forensic for multimedia and remote sensing applicaitons. He is an Associate Editor of the IEEE Transactions on Circuits and Systems for Video Technology and the EURASIP Journal on Image and Video Processing. He is a Fellow of the ELLIS Society for the advancement of artificial intelligence in Europe, and has been an IEEE Distinguished Lecturer from 2015 to 2016. He was the recipient of the IEEE Geoscience and Remote Sensing Society 2011 Transactions Prize Paper Award, the IEEE ICIP 2015 Best Student Paper Award (as senior author), the IEEE ICIP 2019 Best Paper Award, the IEEE Multimedia 2019 Best Paper Award, and the 2010 and 2014 Best Associate Editor Award of the IEEE TRANSACTIONS ON CIRCUITS AND SYSTEMS FOR VIDEO TECHNOLOGY.
\end{IEEEbiography}

\end{document}